\pdfoutput=1

\documentclass[11pt]{article}

\usepackage[final]{acl}

\usepackage{times}
\usepackage{latexsym}

\usepackage[T5,T1]{fontenc}
\DeclareTextFontCommand{\textvietnamese}{\fontencoding{T5}\selectfont}

\usepackage[utf8]{inputenc}

\usepackage{microtype}

\usepackage{inconsolata}

\usepackage{url}
\usepackage{listings}
\usepackage{xcolor}
\usepackage{textcomp}  
\usepackage{scalerel}  
\usepackage{enumitem}
\usepackage{courier}

\definecolor{codegreen}{rgb}{0,0.6,0}
\definecolor{codegray}{rgb}{0.5,0.5,0.5}
\definecolor{codepurple}{rgb}{0.58,0,0.82}
\definecolor{backcolour}{rgb}{0.95,0.95,0.92}

\lstdefinestyle{mystyle}{
    backgroundcolor=\color{backcolour},   
    commentstyle=\color{codegreen},
    keywordstyle=\color{magenta},
    numberstyle=\tiny\color{codegray},
    stringstyle=\color{codepurple},
    basicstyle=\ttfamily\footnotesize,
    breakatwhitespace=false,         
    breaklines=true,                 
    captionpos=b,                    
    keepspaces=true,                 
    numbersep=5pt,                  
    showspaces=false,                
    showstringspaces=false,
    showtabs=false,                  
    tabsize=2,
    framextopmargin=50pt,
    basicstyle=\footnotesize\ttfamily,
}

\usepackage[breakable]{tcolorbox}

\usepackage{graphicx}
\usepackage{amsmath,amssymb}
\usepackage{booktabs, multirow} 
\usepackage{soul}
\usepackage{xcolor,colortbl} 
\usepackage[dvipsnames]{xcolor} 
\usepackage{changepage,threeparttable} 
\usepackage{tablefootnote}
\usepackage{anyfontsize} 

\DeclareMathOperator*{\argmax}{arg\,max}

\DeclareMathOperator{\E}{\mathbb{E}}

\def\cross{\scalerel*{\includegraphics{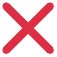}}{\textrm{\textbigcircle}}}
\def\check{\scalerel*{\includegraphics{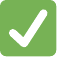}}{\textrm{\textbigcircle}}}

\title{Is Active Persona Inference Necessary for Aligning Small Models to Personal Preferences?}

\author{
 \textbf{Zilu Tang\textsuperscript{1}},
 \textbf{Afra Feyza Akyürek\textsuperscript{1}},
 \textbf{Ekin Akyürek\textsuperscript{2}},
 \textbf{Derry Wijaya\textsuperscript{1,3}},
\\
 \textsuperscript{1}Boston University,
 \textsuperscript{2}MIT,
 \textsuperscript{3}Monash University Indonesia,
\\
\texttt{zilutang@bu.edu}
}

\begin{document}
\maketitle
\begin{abstract}
A prominent issue in aligning language models (LMs) to personalized preferences is underspecification-- the lack of information from users about their preferences. A popular trend of injecting such specification is adding a prefix (e.g. prior relevant conversations) to the current user's conversation to steer preference distribution. Most methods passively model personal preferences with prior example preferences pairs. We ask whether models benefit from actively inferring preference descriptions, and address this question by creating a synthetic personalized alignment dataset based on famous people with known public preferences. We then test how effective finetuned 1-8B size models\footnote{We find larger models quite good at personalization with prompting, hence only leveraged it for dataset generation.} are at inferring and aligning to personal preferences. Results show that higher-quality active prefixes lead to better generalization, more contextually faithful models, and less systematic biases across different protected attributes. All our results suggest active alignment can lead to a more controllable and efficient path for personalized alignment.\footnote{We release our research artifacts in \url{https://github.com/PootieT/famous-persona}}

\end{abstract}

\begin{figure}[!t]
    \centering
    \includegraphics[width=1.0\linewidth]{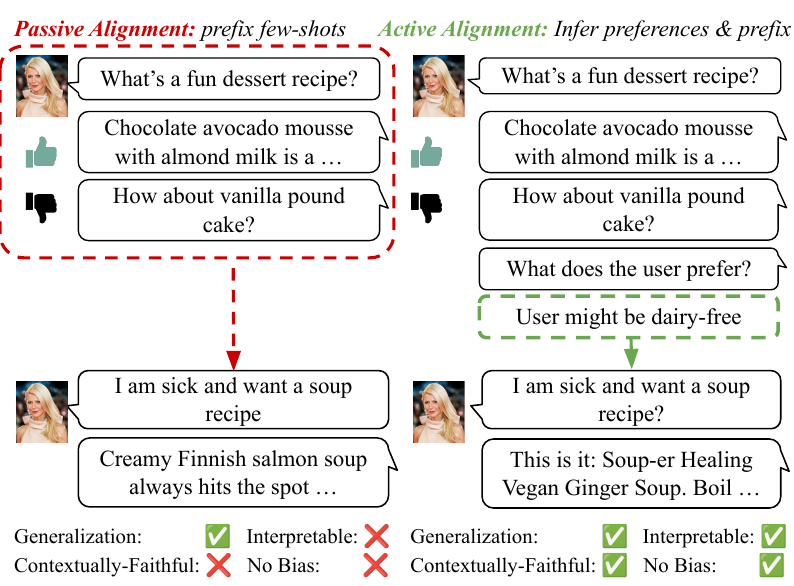}
    
    \caption{We construct a personalized alignment dataset on famous people to investigate whether \textbf{active}ly inferring preferences is necessary for finetuning personalized alignment models. We find active alignment to be more interpretable, contextually faithful, and less biased.}
    \label{fig:figure1}
\end{figure}

\section{Introduction}
\label{sec:introduction}

Preference alignment has become a standard pipeline in finetuning models to follow \emph{generic} human preferences. Most work seeks to optimize models to produce responses that would be preferable \emph{on average}, simplifying the diverse and often \emph{contradicting} space of human preferences. 
The focus for personalized alignment emerges as the demand for adapting models to individual user preferences rises with industrial applications and fairness concerns for large pretrained models. One major issue when personalizing generic alignment algorithms is mitigating underspecification. User-specific information needs to be incorporated to customize the reward distribution downstream.
The majority of prior works proposes passive alignment-- learning to influence reward through observing similar prior interactions. This can either be incorporated through few-shot examples in the prompt \cite{wang2024learning, zollo2025personalllm}, prefix embeddings  \cite{li2024personalized, poddar2024vpl}, meta-learning\cite{zhao2023group, yang2024metaaligner}, or preference prototypes \cite{wang2024learning, park2024valuescope}.
While passive alignment allows fine-grained steering that benefits from the scale of prior interactions, active alignment methods seek to directly guide personalization with instructions. Most work with this approach follows Multi-objective Reinforcement Learning (MORL) paradigm \cite{liu2014multiobjective}, recognizing that alignment objectives often involve competing goals (e.g. helpful vs. harmless) with a limited number of objectives (typically less than five) \cite{jang2023personalized}. However, MORL-based works have yet to show whether active alignment can fully leverage the expressiveness of natural language instructions for fine-grained preference steering. With this gap in-mind, we synthetically generate a dataset of famous people with publicly known preferences, and compare passive vs. active alignment. We summarize our contribution as follows:

\paragraph{Dataset of Personal Preference} We release a personalized alignment dataset based on real people with diverse and contradicting preferences.
\paragraph{Active vs. Passive} We compare active and passive alignment strategies across four models of size 1-8B, and show that active alignment can improve reward generalization on unseen personas.
\paragraph{Contextual Faithfulness} We analyze the models' attribution pattern to prefixes and find active aligned models more contextually faithful. This improves with the quality of inferred personas.
\paragraph{Systematic Bias} We find systematic biases in persona inference and alignment and that active alignment results in less bias. 

\section{Background \& Related works}
\label{sec:background}

\paragraph{Personalized alignment datasets.}
Personalization has been extensively studied in many fields prior to LLMs \cite{chen2023large}, beginning with collaborative filtering in recommendation systems \citep{goldberg1992using}. With popularization of post-training preference alignment to human feedback \cite{ouyang2022training}, initial personalized alignment datasets take inspiration from MORL-paradigm \cite{bai2022training, ji2024beavertails, jang2023personalized, yang2024metaaligner, gao2024linear, poddar2024vpl, chakraborty2024maxmin}. Constructing such datasets is relatively straightforward. Simple objectives (e.g. detailed vs. concise responses) can be controlled in generation through prompting and evaluated with LLMs \cite{jang2023personalized}. 
The biggest assumption of MORL is that objectives are compositional, and the span covers the entire preference space. This assumption is flawed, however, as human preferences can be infinitely nuanced (e.g. liking squash over tennis) so that no amount of objectives can cover the space of personal preferences \cite{slovic1995construction, macintyre2013after, aroyo2015truth, gabriel2020artificial, klingefjord2024human}. Even if preference space is compositional, modeling challenges remain. \cite{wang2024large,beck2024sensitivity}.

Another popular choice is predicting human survey responses \cite{durmus2023globalopinionqa, santurkar2023whose, zhao2023group, do2023choire, feng2024modular, li2024steerability, hwang2023aligning, jiang2024can}. Although measuring opinions can serve as a valuable evaluation tool, these tasks in general are not for improving conversational assistants. Recent work \cite{zollo2025personalllm} synthetically construct users preferences through linear combinations of off-the-shelf reward models. \citet{kirk2024prism} collects response preference pairs from diverse user backgrounds, and \citet{castricato2024persona} synthetically constructs personas and respective conversations using prompts from PRISM. However, none of these datasets contain ground-truth persona preferences from which we can evaluate preference inference (i.e. active alignment). See dataset comparisons in Appendix~\ref{tab:dataset-comparison}.


\paragraph{Alignment methods.}
For MORL-based active alignment, methods usually involve merging separately trained adapters, or programmatically composed prompt prefixes \cite{jang2023personalized, wang2024arithmetic}. Other works focus on pluralistic alignment from group perspectives \cite{sorensen2024roadmap, park2024valuescope}, which typically use meta-learning \citep{zhao2023group}, or EM-like algorithms to iteratively cluster and align multiple models  \cite{zhong2024provable, park2024principled}. Lastly, many seek to align during decoding \cite{chen2024pad, khano2024vargs, shi2024decoding, gao2024linear, huang2024deal}. Many such works are orthogonal to us, where we focus on the most simple set-up.


\paragraph{Active preference inference and underspecification}
Inferring human preferences from sparse examples or underspecified instructions is important for seamless human-AI interaction \cite{milli2017should}.
Prior works infer different aspects of human preferences, such as implicit social contracts \citep{franken2023social}, constitutions \citep{chen2024iteralign}, 
and user values \citep{sun2024persona, liu2024large, balepur2025whose, li2025extended, bismay2025reasoningrec}. These works reinforce our point that explicitly inferring user preference is crucial for interpretable alignment.
Prefixing inferred persona can also be considered as addressing underspecification \cite{lee2022diversify}, which leads to spurious correlation and short-cut learning \cite{geirhos2020shortcut}. In preference learning, underspecified data -- such as users upvoting Reddit posts for various latent reasons \cite{ethayarajh2022shp, park2024valuescope} -- leads to non-robust rewards. A solution is to fully specify the preference criteria \cite{siththaranjan2023distributional, yang2024metaaligner}, which in our case, is the inferred personas.

\section{Methodology}

\subsection{Task Definition}

Preference alignment to human feedback \citep{stiennon2020learning, bai2022training,ouyang2022training} assumes a dataset of triples $\mathcal{D}=\{\mathbf{x}, \mathbf{y}_w, \mathbf{y}_l\}$ where $\mathbf{x}$ represents the prompt given to the LM and $\mathbf{y}_w$, $\mathbf{y}_l$ represent the preferred and respectively dispreferred response labeled by the human annotator(s). The task of alignment seeks to optimize a model's likelihood ($\pi$) of generating $\mathbf{y}_\textrm{w}$ over $\mathbf{y}_l$ given $\mathbf{x}$.
In personalized alignment, we introduce the persona variable (e.g. prior conversation, demographics) $p_i$ for each of the $n$ personas. The objective can be defined as:
\vspace{-10pt}
\begin{equation}
    \argmax_{\pi_p}  \E_{{\mathbf{x},\mathbf{y}_l,\mathbf{y}_w}\in \mathcal{D}} \big[\sum_{i\in[n]} \pi_p(\mathbf{y}_w | \mathbf{x}, p_i)\big]
\end{equation}
where $\pi_p$ could be a single or a set of personalized models and $\mathcal{D}=\cup_{i=1}^n \mathcal{D}_i$.

\subsection{Dataset Construction}

We construct our personalization dataset to contain diverse personas with contradicting preferences in four steps (Figure~\ref{fig:dataset-generation-flowchart}).
\textbf{Step 1: Select persona}. With the help of GPT4\footnote{We use \texttt{gpt-4-0613} from OpenAI}, we define 11 axes (topics or attributes) through which preferences might differ (e.g. \texttt{diet}, \texttt{politics}) to ensure contrast in opinions. For each axis, we prompt GPT4 to provide at most five sub-categories (e.g. liberal) along with a famous person associated with the category (e.g. Bernie Sanders). We curate 50 diverse personas, each with definable contrasts. 
\textbf{Step 2: Generate prompts.} We generate two sets of questions ($\mathbf{x}$) -- personal ($\mathbf{x}_{\textrm{personal}}$) and divergent ($\mathbf{x}_{\textrm{divergent}}$) -- for each persona to ensure diversity and contrast. $\mathbf{x}_{\textrm{personal}}$ are based on individualistic preferences, and $\mathbf{x}_{\textrm{divergent}}$ are shared across personas from the same axis who prefers different answers\footnote{similar to controversy guided prompts in \citet{kirk2024prism}}. We sample 100 $\mathbf{x}_{\textrm{personal}}$ and 100 $\mathbf{x}_{\textrm{divergent}}$, using half for training and the other half for testing. 
\textbf{Step 3: Sample Responses.}
We generate $\mathbf{y}$ from $\mathbf{x}$ using our baseline model \textsc{Zephyr}\footnote{\href{https://huggingface.co/HuggingFaceH4/zephyr-7b-beta}{HuggingFaceH4/zephyr-7b-beta}} for the purpose of on-policy improvement \cite{meng2024simpo}. We assume baseline models have no information on the persona during generation and leverage a Chain-of-thought (CoT) prompt to elicit diverse responses. Through sentence-embedding clustering and generic reward model filtering, we obtain four diverse $\mathbf{y}$s per $\mathbf{x}$. Note $\mathbf{x}_{\textrm{divergent}}$ and corresponding $\mathbf{y}$s are shared across personas of that axis, so the same $\mathbf{y}_l$ for one might be the $\mathbf{y}_w$ for another.
\textbf{Step 4: Label Responses.}
We use GPT4-as-personal-judge to obtain the best $\mathbf{y}_w$ from $\mathbf{y}$s through three rounds of pair-wise comparisons.
\citet{dong2024can, castricato2024persona} show that GPT4 can approximate human preferences as well as a third-person annotator. Given extensive public information on the people in our dataset, we expect GPT4 annotation quality to be similar, if not better than a third-person annotator. We verify this with our human annotators, who agree with GPT4 label 78\% of the time. See Appendix~\ref{apx:dataset-construction-details} for more details on the construction process, statistics, and verification efforts. 

Our final dataset contains 50 personas across 11 axis. Each persona has 100 train and 100 test preference pairs, each composed of half personal and half divergent questions. 

\subsection{Training and Evaluation}

We focus on finetuning and evaluating small models (1-8B) as they are primary targets as reward models used during reinforcement learning. Larger models are costly to run, and often do not allow access to internals, which we need for our analysis. Since our dataset construction was done with GPT4, we know large models can customize to personal preferences through prompting in some capacity, and leave the extension of our analysis to larger models for future directions.

Through preliminary studies (Appendix~\ref{apx:prompting-results}, \ref{apx:personal-models-results}), we find small models to be in-effective at in-context learning with few-shot examples. To balance simplicity and performance, we opt-in to finetune our model in a multi-task fashion (\textsc{MT}), updating a single model (adapter) for all users, with a loss similar to DPO \cite{rafailov2024dpo}: 

\vspace{-15pt}
\begin{equation}
\begin{aligned}
    \mathcal{L}_{MT}= - \E_{i\sim[n],(\mathbf{x},\mathbf{y}_w,\mathbf{y}_l)\sim \mathcal{D}_i} \big[ \log \big(\\
    \beta \log \frac{\pi_\theta (\mathbf{y}_w | p_i, \mathbf{x})}{\pi_{ref} (\mathbf{y}_w | p_i, \mathbf{x})}
     - \beta \log \frac{\pi_\theta (\mathbf{y}_l | p_i, \mathbf{x})}{\pi_{ref} (\mathbf{y}_l | p_i, \mathbf{x})} \big) \big],
\end{aligned}
\label{eqn:loss-dpo-mt}
\end{equation}
where each $p_i=f(\mathbf{x}, \mathbf{y}_{w}, \mathbf{y}_{l})$ 
is a fixed person-specific prefix. We test the following \textbf{passive} and \textbf{active} prefixes:

\paragraph{Passive Prefixes} 
We randomly sample \textbf{few-shot} prefixes \cite{zhao2023group} with two $(\mathbf{x},\mathbf{y}_w)$ pairs from each persona's train split \footnote{We found 2-shots to be the optimum number of shots with baseline model given the long response nature of our dataset}. For embedding-based method \textbf{VPL} \cite{poddar2024vpl}, we train a variational auto-encoder that embeds 8-shots into a single embedding token. We also include baseline prefix \textbf{tag}, an ID string unique to each user.

\paragraph{Active Prefixes}
For our oracle \textbf{gold persona}, we prompt GPT to generate the background and preferences given the name of the person. This is the only prefix where the names are revealed to the inference model. For \textbf{persona} and \textbf{persona gpt4}, we prompt the baseline models and GPT4 to generate the same information using four random shots\footnote{We found significant degradation using more than four shots in preliminary experiments.}. Note that \textbf{persona} is unique to each inference model.

We perform five-fold cross-validation (CV) across axes to evaluate generalization (``seen person'' vs ``unseen persona'') as models need to personalize to new users without training in practice. We finetune four LMs across two model families: \textsc{Llama1/3b}, \textsc{Zephyr} (7B), and \textsc{Ministral}(8B)\footnote{\href{https://huggingface.co/meta-llama/Llama-3.2-1B-Instruct}{meta-llama/Llama-3.2-1B-Instruct}, \href{https://huggingface.co/meta-llama/Llama-3.2-3B-Instruct}{meta-llama/Llama-3.2-3B-Instruct}, \href{https://huggingface.co/mistralai/Ministral-8B-Instruct-2410}{mistralai/Ministral-8B-Instruct-2410}}. Hyperparameters are in Appendix~\ref{apx:hyperparameters}. We show similar results with leave-one-axis-out finetuning in Appendix~\ref{apx:leave-one-axis-out}, except that personas in axes \texttt{politics} and \texttt{family} are hard to generalize.

\subsubsection{Evaluation Metrics}
We adopt internal reference-free rewards\footnote{It is more intuitively aligned with generation as well as findings from \citet{chen2024preference}.} from RewardBench \cite{rafailov2024dpo,lambert2024rewardbench} simplicity, and it can be calculated as $\pi (\mathbf{y}_w \mid \mathbf{x}) > \pi (\mathbf{y}_l \mid \mathbf{x})$
where $\pi$ is the LM, and we average across (log) token probabilities. Unless otherwise mentioned, we report reward accuracy averaged across personas in the unseen splits (50 personas across five models, 100 questions each).

\subsection{Dataset Validation}

\begin{table}[]
    \centering
    \small
\begin{tabular}{lrrrrr}\toprule
&\multicolumn{2}{c}{\textbf{Seen Persona}} &\multicolumn{2}{c}{\textbf{Unseen Persona}} \\\cmidrule{2-5}
&{\footnotesize no prefix}  &{\footnotesize gold} &{\footnotesize no prefix}  &{\footnotesize gold} \\\midrule  
\textsc{Llama1B} &.54{\scriptsize (.05)} &.65{\scriptsize (.06)} &.54{\scriptsize (.06)} &.65{\scriptsize (.06)} \\
\textsc{Llama3B} &.51{\scriptsize (.05)} &.61{\scriptsize (.05)} &.50{\scriptsize (.05)} &.59{\scriptsize (.06)} \\
\textsc{Zephyr} &.53{\scriptsize (.05)} &.62{\scriptsize (.05)} &.52{\scriptsize (.06)} &.61{\scriptsize (.05)} \\
\textsc{Ministral} &.53{\scriptsize (.05)} &.63{\scriptsize (.05)} &.53{\scriptsize (.05)} &.62{\scriptsize (.05)} \\
\bottomrule
\end{tabular}
    \caption{Across 4 models, prefixed finetuning with \textbf{gold persona} significantly improved total reward accuracy, generalizing to unseen personas. Parenthesis = standard deviation across personas.}
    \label{tab:baseline-oracle}
\end{table}

To verify that personal prefix is necessary for our dataset, we finetune \textbf{gold persona} and compare against \textbf{no prefix} \textsc{MT} baseline. In Table~\ref{tab:baseline-oracle}, we see across all models, using \textbf{persona gold} significantly improved total reward accuracy ($\mathbf{x}_{\textrm{personal}}$ + $\mathbf{x}_{\textrm{divergent}}$), validating our dataset. We also see good generalization in unseen personas, suggesting quality prefixes to be crucial for generalization. In the next few sections we see how close non-oracle prefixes can close this performance gap. Example persona prefixes are in Appendix~\ref{apx:qualitative-analysis-personas}.

\section{Results \& Discussions}

\subsection{Quality active personas are more interpretable and improve generalization}
\label{sec:persona-quality-measure}

\vspace{-10pt}
\begin{table}[!htb]
    \centering
    \tiny
\begin{tabular}{lrrrrrr}\toprule
&\textbf{emb ($\uparrow$)} &\textbf{ra. emb($\downarrow$)} &\textbf{R1 ($\uparrow$)} &\textbf{ra. R1($\downarrow$)} &\textbf{words($\downarrow$)} \\\midrule
few-shot &.22 {\tiny (.09)} &.15 {\tiny (.06)} &.19 {\tiny (.03)} &.18 {\tiny (.03)} &563(111) \\
\midrule
\textsc{Zephyr} &.39 {\tiny (.08)} &.33 {\tiny (.07)} &.24 {\tiny (.04)} &.23 {\tiny (.03)} &265 {\tiny (97)} \\
\textsc{Llama1B} &.41 {\tiny (.09)} &.34 {\tiny (.09)} &.25 {\tiny (.04)} &.24 {\tiny (.04)} &260 {\tiny (69)} \\
\textsc{Llama3B} &.42 {\tiny (.08)} &.36 {\tiny (.07)} &.26 {\tiny (.02)} &.25 {\tiny (.03)} &252 {\tiny (59)} \\
\textsc{Ministral} &\textbf{.42 {\tiny (.08)}} &.36 {\tiny (.07)} &.26 {\tiny (.04)} &.25 {\tiny (.03)} &259 {\tiny (73)} \\
GPT4 &.41 {\tiny (.08)} &.32 {\tiny (.09)} &\textbf{.28 {\tiny (.04)}} &.26 {\tiny (.03)} &209 {\tiny (35)} \\
\midrule
persona gold &- &.49 {\tiny (.11)} &- &.33 {\tiny (.10)} &203 {\tiny (18)} \\
\bottomrule
\end{tabular}
    \caption{Mean and standard deviation (across personas) of LM inferred persona against \textbf{persona gold} compared to a \textbf{ra}ndom \textbf{persona gold}. \textsc{Ministral} wins semantically, but \textbf{persona gpt4}s are more separable.}
    \label{tab:persona-quality}
\end{table}
\vspace{-15pt}

\paragraph{Good active prefixes are shorter, more separable, more interpretable.}
Given oracle upperbound, we first measure how good are the inferred \textbf{persona}s compared to \textbf{persona gold}. We use \texttt{Qwen3-Embedding}\footnote{\href{https://huggingface.co/Qwen/Qwen3-Embedding-0.6B}{Qwen/Qwen3-Embedding-0.6B}} cosine similarity and rouge-1\cite{lin2004rouge} to measure semantic similarity and specific vocabulary recall for each model's \textbf{persona}. We also provide baseline comparison against random \textbf{persona gold}: a larger gap between correct vs. random persona indicates better separability between personas. In Table~\ref{tab:persona-quality}, we see more recent models perform better semantically, with \textsc{Ministral} on top, but \textbf{persona gpt4} wins in separability. \textbf{few-shot} are not bad in separability but the worst in semantic similarity and length, this suggests that even though passive alignment (\textbf{few-shot}) might perform well in distinguishing user profiles, the prefixes are likely much less \emph{interpretable}.

\begin{figure}[!htb]
    \centering
    \includegraphics[width=1.0\linewidth,trim={0.4cm 0.3cm 0.4cm 0.5cm},clip]{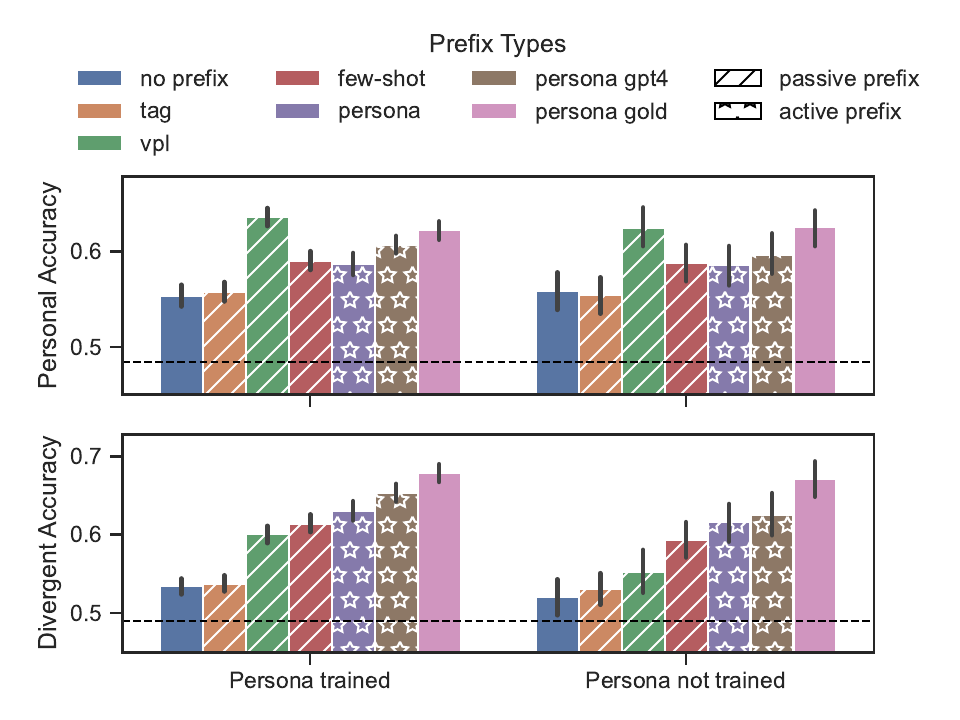}
    \caption{Finetuning results with 5-fold CV on \textsc{Zephyr}. Error bars indicate 95\% confidence intervals (CI) across personas. Dashed line indicates no prefix prompting baseline. Good quality active prefix (\textbf{persona gpt4}) generalizes well especially in divergent questions.}
    \label{fig:multitask-results}
\end{figure}

\paragraph{Better active prefix generalizes better in divergent questions} We plot \textsc{MT(Zephyr)} performance across prefixes in Table~\ref{fig:multitask-results}. In passive prefixes, both \textbf{VPL} and \textbf{tag} use a single token, yet \textbf{tag} performs similarly to \textbf{no prefix} while \textbf{VPL} performs much better. This suggests semantics rather than capacity is the issue in associating preference with prefix. \textbf{VPL} also excels in personal questions but fails in divergent questions, indicating embedding-based methods compress few-shots information well but fails to encode semantic contrasts (i.e. embeddings for ``I like lamp'' is close to that of ``I don't like lamp'') \cite{tang2022augcse}. This suggests an important future direction is to actively infer persona compressed from more shots. \textbf{Persona gpt4} outperforms \textbf{persona}, which outperforms \textbf{few-shot}, suggesting precise and separable prefixes (Table~\ref{tab:persona-quality}) are more effective, not only for computational efficiency but also for generalization, especially in divergent questions. We show similar findings with other models in Table~\ref{apx:finetuning-other-models}. Notably, \textsc{Llama1/3B} models prefer few-shots over self-generated \textbf{persona}, whereas it is the opposite for \textsc{Zephyr} and \textsc{Ministral}. 
In Appendix~\ref{apx:multitask-prefix-sensitivity}, we investigate prefix sensitivity using shuffled and alternative personas and find \textbf{persona gpt4} to be the most robust across variations.\footnote{Even though GPT4 generates the dataset and \textbf{persona gpt4}, \textsc{MT} cannot exploit any shortcuts to predict preferences, so the improvements stems purely from prefix quality.}

\begin{figure}[!htb]
    \centering
    \includegraphics[width=1.0\linewidth,trim={1.7cm 0.4cm 0 0.8cm},clip]{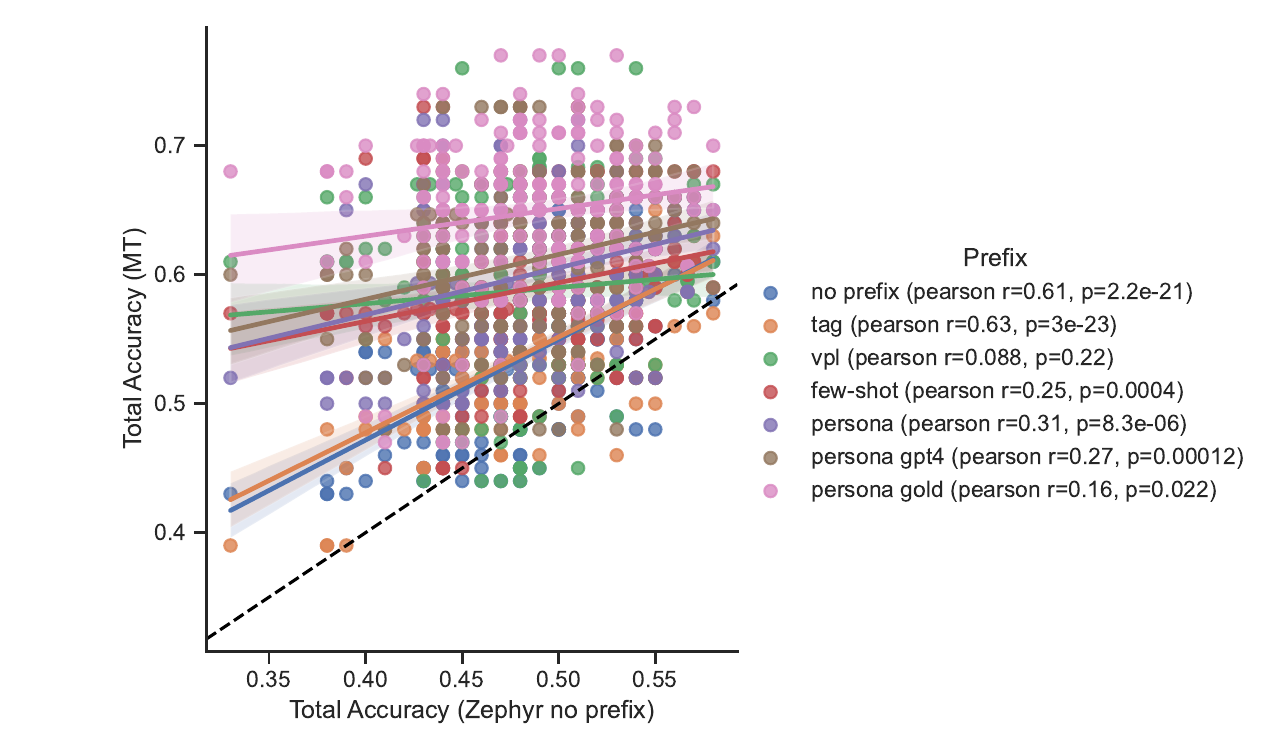}
    \caption{\textsc{MT} (persona not trained) vs. \textsc{Zephyr} with no prefix . We calculate Pearson correlation with p-value per prefixes. Better prefixes result in lower correlation and more equitable improvement. Dashed line is no improvements (y=x). Shaded areas indicates 95\% CI.}
    \label{fig:multitask-correlation-no-train-no-prefix}
\end{figure}

\paragraph{More precise prefix, more equitable improvements} We plot finetuning total accuracy on persona not trained against prompting \textsc{Zephyr} with no prefix in Figure~\ref{fig:multitask-correlation-no-train-no-prefix}. \textbf{VPL} improves the most equitably across all personas (the most flat line), which indicates that compressing more information for each user is crucial. \textbf{Persona gpt4}, \textbf{persona} and \textbf{few-shot} each outperforms the next while being less correlated to the baseline, suggesting that higher quality prefixes might also align more equitably.

\paragraph{Additional results.} In Appendix~\ref{apx:gen-eval}, ~\ref{apx:alignment-tax} we show similar trends with generational evaluation, and discuss mitigating alignment tax \cite{lee2024bapo}.

\subsection{Prefix quality vs. Contributive Attribution}
\begin{figure*}[!htb]
    \centering
    \includegraphics[width=1\linewidth,trim={0.2cm 0.4cm 0.2cm 0.2cm},clip]{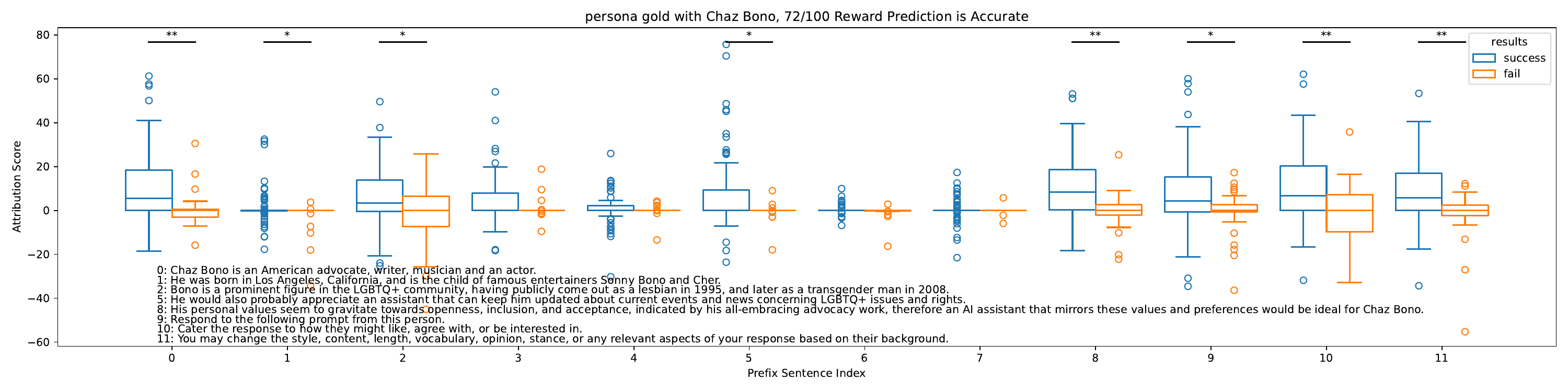}
    \includegraphics[width=1\linewidth,trim={0.2cm 0.4cm 0.2cm 0.2cm},clip]{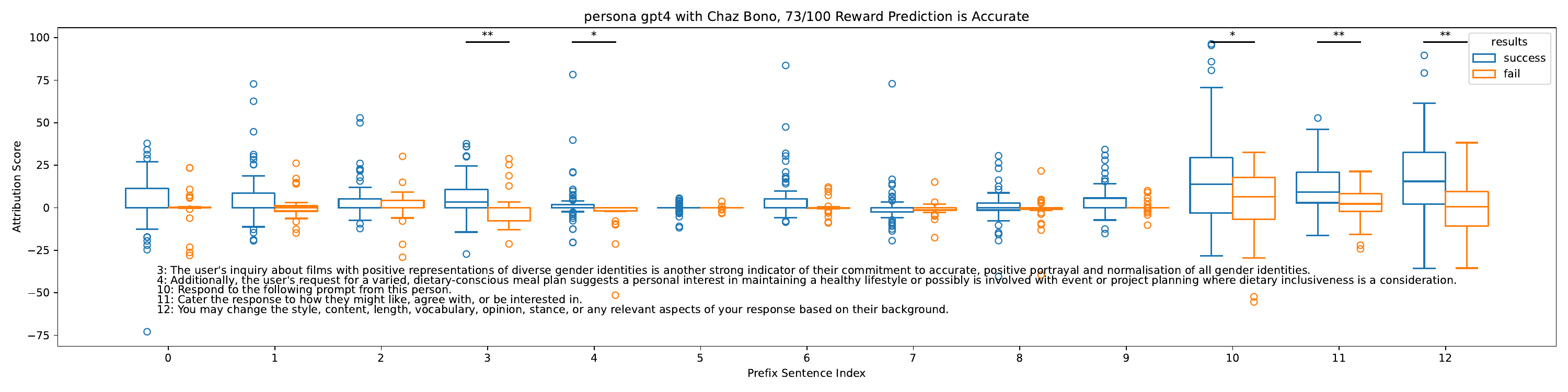}
    \includegraphics[width=1\linewidth,trim={0.2cm 0.4cm 0.2cm 0.2cm},clip]{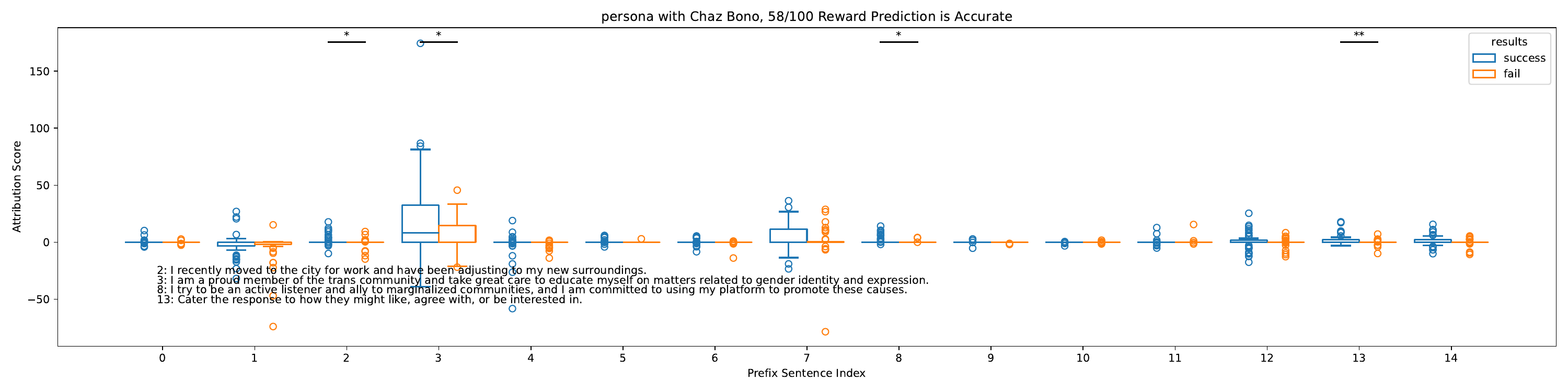}
    \includegraphics[width=1\linewidth,trim={0.2cm 0.4cm 0.2cm 0.2cm},clip]{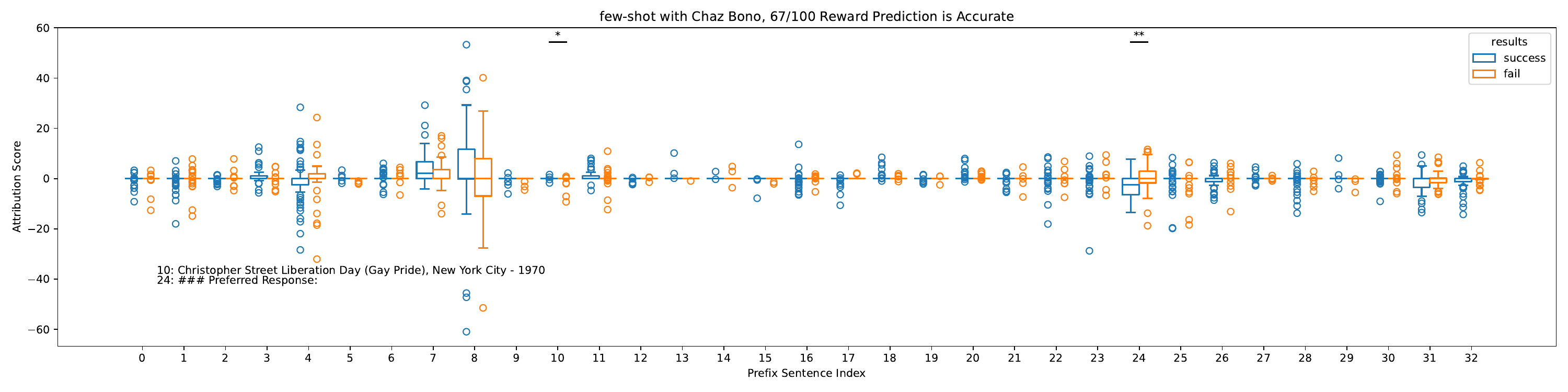}
    \caption{Attribution $\textbf{s}_{diff}$ for each sentence within prefixes for Chaz Bono with \textsc{MT (Zephyr)} model (persona unseen), grouped by reward accuracy of the questions. For each sentence, we perform student t-test between success/fail scores and mark * (p<0.05) and **(p<0.01) at the top. Each marked sentence is also displayed in text along with its indices. Active persona prefixes have more contributing sentences that are more interpretable.}
    \label{fig:attribution-plot-chaz}
\end{figure*}

Given that prefixes control reward distributions, it is important to understand \emph{how} model responses are causally dependent on prefixes (i.e. contextual faithfulness). If algorithms use prefixes solely to differentiate between users but disregard the underlying semantics, finetuned models could learn from spurious correlation and exhibit contextually unfaithful behaviors. We use ContextCite \cite{cohen2024contextcite}, to measure contributive attribution from each sentence in the prefix to the responses (through surrogate modeling\footnote{We refer readers to \citet{cohen2024contextcite} for details.}). In other words, the score $\mathbf{s}(\mathbf{x},\mathbf{y}) \in \mathbb{R}^L$, where $L$ is the number of sentences in a prefix, tells us how important each sentence is to the log-likelihood of the LM response. For each persona and prompt, we compute the difference $\mathbf{s}_{diff} =  \mathbf{s}(\mathbf{x},\mathbf{y}_{w}) - \mathbf{s}(\mathbf{x},\mathbf{y}_l)$.
In Figure~\ref{fig:attribution-plot-chaz}, we plot $\mathbf{s}_{diff}$ (grouped by reward accuracy) for the test questions of a user across four prefixes from \textsc{MT (Zephyr)} models. We see that active persona prefixes have more contributing sentences that are more interpretable than passive \textbf{few-shot} prefix. This trend holds for higher quality personas (\textbf{gold}> \textbf{gpt4}> \textbf{persona}).

\begin{table}[!htb]
    \centering
    \tiny
\begin{tabular}{lrrrrr}\toprule
Prefix &\textbf{few-shot} &\textbf{persona} &\textbf{persona gpt4} &\textbf{persona gold} \\\midrule
Influence fraction &.14 {\tiny (.11)} &.20 {\tiny (0.17)} &.36 {\tiny (0.22)} &.40 {\tiny (.20)} \\
\bottomrule
\end{tabular}
    \caption{IF across personas (unseen) in \textsc{MT (Zephyr)}. Better performing prefixes results in models that are more contextually faithful.}
    \label{tab:attribution-zephyr}
\end{table}

\begin{table}[!htb]
    \centering
    \small
\begin{tabular}{lrrr}\toprule
\textbf{Model} &\textbf{persona} &\textbf{few-shot} \\\midrule
\textsc{Llama1B}   &\textbf{0.052} {\scriptsize (.067)} &0.045 {\scriptsize (.057)} \\
\textsc{Llama3B}   &0.100 {\scriptsize (.162)} &\textbf{0.117} {\scriptsize (.112)} \\
\textsc{Zephyr}    &\textbf{0.199} {\scriptsize (.165)} &0.140 {\scriptsize (.105)} \\
\textsc{Ministral} &\textbf{0.218} {\scriptsize (.181)} &0.106 {\scriptsize (.093)} \\
\bottomrule
\end{tabular}
    \caption{IF across three out of four models show self-generated active \textbf{persona} leads to more contextually faithful models. Larger LMs attributes to \textbf{persona} more.}
    \label{tab:attribution-four-models}
\end{table}

With an intuitive understanding of the score qualitatively, we operationalize this as a metric which we can measure qualitatively across models and prefixes. For each persona, we calculate \textbf{influence fraction (IF)}: the fractions of sentences that significantly ($p<0.05$) contribute to correct reward prediction across the test-split of the person\footnote{Equivalent to fractions of sentence with * in Figure~\ref{fig:attribution-plot-chaz}}. A prefix with higher IF indicates that models are more causally influenced (i.e. contextually faithful) by the prefix. In Table~\ref{tab:attribution-zephyr}, we see that better performing prefixes result in higher IF. In Table~\ref{tab:attribution-four-models}, we see three out of four models attribute to self-generated \textbf{persona} more than \textbf{few-shot}, despite \textsc{Llama1B} performing better with \textbf{few-shot}. These results suggests that better quality active prefixes results in more contextually faithful models, and even self-generated \textbf{persona} could lead to better attribution.

\subsection{Prefix Distributional Shift vs. Attribution}

In Table~\ref{fig:multitask-prefix-sensitivity} we show that quality actively-inferred personas are robust to surface form variations but drop performance with prefix from a random person. Here we investigate attribution sensitivities to varying inference prefixes.

\begin{table}[!htb]
    \centering
    \small
\begin{tabular}{lrr}\toprule
\textbf{Train Prefix} &\textbf{Inference Prefix} &\textbf{IF} \\\midrule
\textbf{few-shot} &(same as train) &0.140 {\scriptsize (.105)} \\
&few-shot (alt. seed) &0.142 {\scriptsize (.087)} \\
\midrule
\textbf{persona} & (same as train) &0.199 {\scriptsize (.165)} \\
&persona (alt. seed) &0.170 {\scriptsize (.160)} \\
&persona gpt4 &0.179 {\scriptsize (.158)} \\
&persona gold &0.145 {\scriptsize (.153)} \\
\midrule
\textbf{persona gold} &(same as train) &0.398 {\scriptsize (.202)} \\
&persona &0.257 {\scriptsize (.174)} \\
&persona gpt4 &0.283 {\scriptsize (.202)} \\
\bottomrule
\end{tabular}
    \caption{IF with alternative prefixes at inference time for \textsc{MT(Zephyr)}. Faithfulness always decreases with alternative prefix, and models trained with higher quality prefix remain more faithful.}
    \label{tab:attribution-under-distributional-shift}
\end{table}
\vspace{-20pt}

\paragraph{Quality persona is crucial during train and test.}
In Table~\ref{tab:attribution-under-distributional-shift}, we can see both \textbf{few-shot} and \textbf{persona} attribute similarly to a prefix using different shots (alt. seed). Unique to active prefixes, we can infer with varying quality of personas to see if model adapt to changes. Unfortunately, distributional shift only lowers attribution, even if we increase the quality of persona at inference time. 
 
\paragraph{Self-generated active persona more causal}
In Table~\ref{tab:attribution-pre-post-finetuning-zephyr}, we look into IF before and after finetuning. Unfinetuned \textsc{Zephyr} attributes to \textbf{persona} more than \textbf{few-shot} or even \textbf{persona gold}, despite failing at reward accuracy (Appendix~\ref{apx:prompting-results}). \textsc{Zephyr} increased attribution through simply inferring preferences from out-of-distribution \textbf{few-shot} examples to in-distribution \textbf{persona}. This is potentially useful for debugging model faithfulness in general \cite{turpin2023language}, where generations are not reflective of internal mechanisms.\footnote{We did not observe similar behavior with \textsc{Llama1/3B}, suggesting the responses being on-policy is also crucial.}

\begin{table}[!htb]
    \centering
    \small
\begin{tabular}{lrrr}\toprule
\textbf{Prefix} & \textbf{Finetuned} & \textbf{Influential Fraction}  \\\midrule
\textbf{few-shot} &false &0.011 {\scriptsize (.019)} \\
&true &0.140 {\scriptsize (.105)} \\
\textbf{persona} &false &\cellcolor[HTML]{A8A8A8}0.141 {\scriptsize (.108)} \\
&true &0.200 {\scriptsize (.165)} \\
\textbf{persona gold} &false &0.016 {\scriptsize (.033)} \\
&true &0.399 {\scriptsize (.202)} \\
\bottomrule
\end{tabular}
    \caption{Unfinetuned \textsc{Zephyr} attributes to self-generated \textbf{persona} much better than other prefixes (\textcolor{gray}{highlighted}), suggesting simple rephrasing with base LM could lead to more contextually faithful generations.}
    \label{tab:attribution-pre-post-finetuning-zephyr}
\end{table}


\begin{table}[!htb]
    \centering
    \small
\begin{tabular}{lrrrr}\toprule
\textbf{Prefix} &\textbf{Train} &\textbf{Inference} &\textbf{Positive Fractions} \\\midrule
\multirow{6}{*}{\textbf{few-shot}} &bm25 &bm25 &0.224 {\scriptsize (.142)} \\
&bm25 &bm25 (reverse) &0.220 {\scriptsize (.141)} \\
&bm25 &- &0.221 {\scriptsize (.137)} \\
&- &bm25 &0.213 {\scriptsize (.176)} \\
&- &bm25 (reverse) &0.197 {\scriptsize (.178)} \\
&- &- &0.197 {\scriptsize (.170)} \\
\midrule
\multirow{6}{*}{\textbf{persona}} &bm25 &bm25 &\textbf{0.282} {\scriptsize (.177)} \\
&bm25 &bm25 (reverse) &0.274 {\scriptsize (.180)} \\
&bm25 &- &0.263 {\scriptsize (.173)} \\
&- &bm25 &0.253 {\scriptsize (.141)} \\
&- &bm25 (reverse) &0.248 {\scriptsize (.143)} \\
&- &- &0.238 {\scriptsize (.138)} \\
\bottomrule
\end{tabular}
    \caption{Number of sentences with positive $\mathbf{s}_{diff}$ in \textsc{Zephyr} trained and/or inferenced on retrieved shots. Dash indicates fixed original prefix. Training on retrieved prefixes increase attribution more significantly for \textbf{persona} than \textbf{few-shot}.}
    \label{tab:retrieve-train-inference}
\end{table}

\begin{table*}[!htb]
    \centering
    \small
\begin{tabular}{p{1.2cm}p{14cm}}\toprule
\multicolumn{2}{c}{\textbf{Top 10 Easiest Personas to Infer}} \\\midrule
\textbf{across shots} &\textcolor{BurntOrange}{Serena Williams (10.0)}, \textcolor{BurntOrange}{Mike Trout (10.8)}, \textcolor{BurntOrange}{Chaz Bono (13.4)}, Bernie Sanders (13.4), Barack Obama (14.6), Megan Fox (14.6), \textcolor{BurntOrange}{LeBron James (14.8)}, Richard Gere (15.8), \textcolor{BurntOrange}{David Beckham (16.0)}, Jennifer Aniston (16.8) \\\hline
\textbf{across models} &Beyoncé(13.0), \textcolor{BurntOrange}{Serena Williams(13.2)}, Zayn Malik(13.6), \textcolor{BurntOrange}{LeBron James(14.6)}, Alexandria Ocasio-Cortez(15.0), Jennifer Aniston (16.4), J.K. Rowling(16.8), \textcolor{BurntOrange}{David Beckham(17.0)}, \textcolor{BurntOrange}{Mike Trout(17.0)}, \textcolor{BurntOrange}{Chaz Bono(17.8)} \\\midrule
\multicolumn{2}{c}{\textbf{Bottom 10 Hardest Personas to Infer}} \\\midrule
\textbf{across shots} &\textcolor{cyan}{Yoshua Bengio(40.8)}, \textcolor{Cyan}{Sam Smith (40.8)}, Meryl Streep (37.8), Latanya Sweeney (37.2), Richard Dawkins(33.4), Timnit Gebru (33.0), Robert De Niro (32.8), Donald Trump (32.6), \textcolor{Cyan}{Bill Clinton(32.2)}, Oprah Winfrey (30.8) \\\hline
\textbf{across models} &Suchi Saria (41.2), \textcolor{cyan}{Yoshua Bengio (38.6)}, Millie Bobby Brown (38.4), Sheryl Sandberg (37.0), Halle Berry (36.6), \textcolor{Cyan}{Sam Smith (34.4)}, \textcolor{Cyan}{Bill Clinton (32.2)}, Tiger Woods (32.2), Laverne Cox (31.0), Billie Eilish (30.8) \\
\bottomrule
\end{tabular}
    \caption{Easiest and hardest personas by inferred \textbf{persona} quality (average rank in parenthesis). Colored names appear in both splits. We see personas in axis \texttt{sports} and (liberal) \texttt{politics} are consistently easy to infer for LMs, while those in \texttt{AI Professors} are hard.}
    \label{tab:top-bottom-persona-quality}
\end{table*}

\begin{table*}[!htb]
    \centering
    \small
    \resizebox{1.0\textwidth}{!}{
    \begin{tabular}{lrrrrrrrrr}\toprule
\textbf{attribute} &\textbf{econ. status} &\textbf{birth country} &\textbf{education} &\textbf{prof.} &\textbf{curr. state} &\textbf{religion} &\textbf{marriage} & \\\midrule
\textbf{persona} &\colorbox{red}{0.02} &\colorbox{red}{0.05} &0.07 &0.11 &0.22 &0.22 &0.23 & \\
\textbf{persona gpt4} &\colorbox{red}{0.02} &\colorbox{red}{2e-3} &0.17 &\colorbox{red}{6e-11} &\colorbox{red}{1e-4} &\colorbox{red}{3e-4} &0.37 & \\
\textbf{persona gold} &\colorbox{red}{3e-5} &\colorbox{red}{0.05} &\colorbox{red}{3e-5} &\colorbox{red}{2e-13} &\colorbox{red}{2e-4} &0.08 &\colorbox{red}{0.02} & \\
\textbf{few-shot} &\colorbox{red}{5e-3} &\colorbox{red}{0.02} &\colorbox{red}{4e-4} &\colorbox{red}{3e-8} &\colorbox{red}{1e-4} &0.14 &\colorbox{red}{0.03} & \\
\midrule
\textbf{attribute} &\textbf{birth state} &\textbf{ethnicity} &\textbf{age} &\textbf{sex pref.} &\textbf{curr. country} &\textbf{gender} &\textbf{race} &\textbf{politics} \\
\midrule
\textbf{persona} &0.28 &0.35 &0.39 &0.39 &0.44 &0.45 &0.55 &0.98 \\
\textbf{persona gpt4} &\colorbox{red}{4e-4} &\colorbox{red}{2e-4} &0.09 &\colorbox{red}{0.5} &0.16 &\colorbox{red}{0.01} &0.52 &0.58 \\
\textbf{persona gold} &\colorbox{red}{8e-5} &0.05 &\colorbox{red}{0.04} &0.06 &\colorbox{red}{0.1} &\colorbox{red}{2e-5} &0.67 &0.53 \\
\textbf{few-shot} &\colorbox{red}{0.02} &\colorbox{red}{8e-5} &\colorbox{red}{2e-3} &0.13 &\colorbox{red}{0.02} &\colorbox{red}{2e-3} &\colorbox{red}{0.02} &0.43 \\
\bottomrule
\end{tabular}
}
    \caption{P-values of one-tail ANOVA across four models before and after finetuning show that \textbf{few-shot} improvements are non-uniform (in 12/15 attributes) where as \textbf{persona} improvements are much more equitable (2/14). Increasing persona specificity and quality (\textbf{persona} $\rightarrow$ \textbf{persona gpt4} $\rightarrow$ \textbf{persona gold}) decreases improvement equity, suggesting a performance vs. fairness trade-off.}
    \label{tab:systematic-bias-after-finetuning}
\end{table*}

\paragraph{Retrieval during training improves contextual faithfulness, more during training and with \textbf{persona}.}
A prefix has different aspects that can influence preferences on the response to a question (i.e. Halle Berry has diabetes and is also an African American Actress), and the aspect that influences the preference distribution for each question may be different. Majority of our investigation trains a static prefix for all questions. Such static prefixes need to cover all aspects of the persona, and LMs have to select the relevant information during generation, placing an upperbound on IF. However, if we were to provide dynamic prefixes that contain relevant information only through retrieval, can we further increase contextual faithfulness?
To investigate this, we dynamically retrieve shots with BM25\cite{robertson1993bm25, xing2024bm25s} that are closest to current train / test question as the prefix. At test time, we vary prefix with static, retrieved prefix, and reverse prefix (shots farthest in distance). 
Instead of IF, we use positive fraction: the fraction of sentences with positive $\mathbf{s}_{diff}$.
In Table~\ref{tab:retrieve-train-inference}, we can see that training with retrieved shots indeed increases contextual faithfulness and \textbf{persona} benefits more than \textbf{few-shot}. Less improvements are observed at test time.

\subsection{Systematic Bias with Personas?}
\label{sec:systematic-bias}

\paragraph{Persona inference and dataset bias exists.}
\label{sec:systematic-bias-persona-inference}
Previous experiments showed that quality actively-inferred persona improve reward generalization and result in a model that is more contextually faithful. However, given the personas are generated, we need to be cautious against systematic biases \cite{kovavc2023large}. We investigate two sources of bias: persona inference, and finetuning with inferred persona. To check whether there is bias for persona inference, we repeat persona inference 1) \textbf{across shots} 2) \textbf{across models} with the same shots\footnote{Four \textbf{persona} models + \textsc{MT(Zephyr)} \textbf{persona gpt4}}, and measure persona quality against \textbf{persona gold}. We average z-score normalized rouge-1 and embedding similarity (Section~\ref{sec:persona-quality-measure}) and average each person's score rank $\in (0,50)$. In Table~\ref{tab:top-bottom-persona-quality}, we show the top and bottom 10 people and their averaged ranks. We see that people in axis \texttt{sports} and (liberal) \texttt{politics} consistently appear in the top, while \texttt{AI professors} often at the bottom. We suspect this is because public information on athletes are mostly single-faceted, and the only underspecification is the sport they play. Liberal politicians' views on different issues may be highly correlated (e.g. supporting minimum wages indicates strongly their stance on gay marriage). Public information on \texttt{AI professors}, by contrast, is mostly based on objectively written papers which reveals little about their personal views.

\paragraph{Active alignment more equitable than passive after finetuning}
\label{sec:systematic-bias-alignment}
Some preferences might be easier to learn during finetuning, skewing overall preference distributions. We compare total reward accuracy difference between \textsc{MT} and baselines (using \textbf{persona} prefix) across four baseline LMs to understand biases from finetuning. We use one-way ANOVA \cite{lowry2014concepts} to test uniformity of improvements across groups (See Appendix~\ref{apx:demographics-distribution},\ref{apx:performance-per-demographic-attr} for demographics statistics and visualizations). In Table~\ref{tab:systematic-bias-after-finetuning}, we see that \textbf{few-shot} prefix results in non-uniform improvements in more attributes than \textbf{persona}. We suspect this is because persona inference introduces noise and ``diffuses'' away statistical biases. Indeed, when we compare persona with increasing specificity/quality (\textbf{persona} $\rightarrow$ \textbf{persona gpt4} $\rightarrow$ \textbf{persona gold}), improvements becomes less equitable. This suggests an inherent trade-off between improving personalized performance vs. being equitable, likely due to imbalanced parametric knowledge LMs have on different demographics. We believe this to be an important future direction: balancing fairness vs. improvements.

\section{Conclusions}

We constructed FamousPersona, a personalized alignment dataset on famous people, to answer our research question: is active alignment (inferring personal preference) better than passive alignment (simply using few-shots)? Results from reward accuracy generalization, prefix attribution patterns, and bias analysis confirm that actively inferring persona is crucial for interpretable and robust personalized alignment. Future studies should focus on how to further evaluate and de-bias inferred persona, and dynamically modify persona prefixes according to the user query.

\section*{Limitations}

Our dataset presents a playground through which both theoreticians and practitioners in AI alignment can empirically validate their methods. We separate limitations and future works in the following two directions:

\subsection*{Dataset improvement} 
\paragraph{Better axes, prompt generation, and label fidelity.} The selection of axes is not representative of all axes through which human preference differs. However one could arbitrarily extend the dataset to axis of interest to study (e.g. moral, ethical values). One could also extend to include people famous in different countries (and speak different languages), extending personal preference alignment to multilingual setting. The quality of our dataset also depends on GPT4 not hallucinating when generating questions ($\mathbf{x}$) and labeling preferences ($\mathbf{y}_w$/$\mathbf{y}_l$). One valid direction is actually obtaining $\mathbf{x}$ or preference labels from the people we are modeling, and understand the true annotation quality. Beyond label fidelity, personal preferences is a dynamic distribution which changes over time, which would be interesting to model in future works. Lastly, we assume findings from our paper will generalize to non-famous people because we infer prefixes \textbf{persona/ persona gpt4} without revealing the name of the person. However, the questions and preferences could be biased and specific to famous people only. Due to its synthetic nature, it is also not impossible for our oracle \textbf{persona gold} to contain biased assumptions that humans also make from a third-person perspective. Hence there could be further biases that we were not able to find.

\paragraph{Better diversity in responses ($\mathbf{y}$).} When generating candidate responses with CoT, we find it to influence contents the most, leaving other stylistic features mostly unchanged. Future work should look into ways to diversify generations beyond content, which will also make preferences more nuanced and challenging to infer. Additionally, even though we aim to generate diverse response, there is no guarantee that we will end up with one that is a good response (all responses might still be bad). In these cases, providing multiple responses with point-wise estimation of reward might be a better dataset construction method. However, it is much harder for LLM-as-personal-judge. Additionally, we chose to generate responses with \textsc{Zephyr} only because we were interested on-policy effects of alignment. To improve the general utility of the dataset as generic finetuning data, we would have generated diverse responses with multiple more capable models.

\paragraph{Adaptive personalization.} Our response generation process also mimics the trade-off between the exploration vs. exploitation problem in RL: is it better to play safe and generate a generically-good answer or risk for more personalized answer. Future work could look into the process through an online/active learning perspective, balancing general response quality vs. venturing into personalization. Asking follow up clarification questions seems like a promising direction.

\subsection*{Better preference modeling}
\paragraph{Tuning on preference inference} We did a preliminary experiment where we train \textsc{MT} models to predict \textbf{persona gpt4} (over a wrong persona through DPO objective) in addition to aligning preferences, similar to a reasoning distillations setup \cite{mukherjee2023orca}, where we consider \textbf{persona gpt4} as the reasoning trace. We did not see much improvement. Future work can explore further leveraging findings in improving reasoning in LMs \cite{hao2024coconut}. One could also potentially find middle ground between training personal models (\textsc{PM}) and \textsc{MT} by finding training and retrieving ``prototypical'' personas \cite{zhong2024provable}. We focus on our analysis on \textsc{MT} models.

\paragraph{Alternative objectives} In our work, we focus on simple methods that are scalable, efficient, and high-performing. However, many other objectives and methodologies are equally important and promising. During multi-task stage learning, we did not consider the perspective of differential privacy \cite{salemi2024comparing}, whereas in the real world, the use of personal data for generic training requires further scrutinizing. As outlined by \citet{sorensen2024roadmap}, one could also align to diverse expectations by explicitly generating all output preferences (``overton''), which come at the cost of verbosity. Given our finding on alignment tax, future work can also explore the trade-off between personalization and general capability by adapting prefixes with different levels of specification at inference time.

\subsection*{Future Analysis}

\paragraph{Scaling up model sizes.} Due to compute constraints, we were not able to run experiments with models larger than 8B sizes. It would be interesting to confirm whether the advantage of \textbf{active} prefix over \textbf{passive} increases with larger model scale. Why do some models attribute to prefixes more than others? We thought another reason \textsc{Llama1/3B} models might perform better with \textbf{few-shot} is because they were trained on more few-shot data, hence able to leverage the few-shot format better. Without transparency of the training procedures this hypothesis is hard to verify.

\paragraph{Why the bias reduction?}
Why are \textbf{active} prefixes able to reduce bias compared to \textbf{passive} prefixes? From Appendix~\ref{apx:performance-per-demographic-attr} we see \textbf{persona}'s improvement are more mild and equal across different attributes. We conjuncture that this might be the noise introduced in the persona inference process. However, if that is the case, would model start associating non-robust features with preference distributions? Or perhaps it is the explicit mentioning of attributions that improved it? 

\paragraph{Evaluation on other datasets}
We constructed our dataset specifically for the purpose of evaluating persona preferences, hence every person in the dataset has a fixed, detailed, persona that grounds their questions and preferences. However, it may still be an open question how much active persona inference helps on interactions where there may not be a clear preference that generalizes user behavior in other situations. 

\section*{Ethical considerations}

Our dataset is entirely generated from GPT4, hence the dataset (from persona selection, to prompt generation and preference labeling) is dependent on the quality of GPT4. We do not claim personas included in our dataset are faithful to their real world counterparts, nor personas' belief/preferences to be universally good or bad, but offer a playground to construct sets of personas with unique and diverse preferences. The authors manually read through most if not all prompts and responses to make sure there are no offensive content. We emphasize that personas' questions, opinions, and preferences are \emph{not} the same as the real people they are modeled after. Models trained on our dataset should not be used to imitate famous people's opinions other than for research purpose.

Although not specific to our dataset, personalization creates an ``echo chamber'' in which users would be catered responses that they agree with, aggravating the issue of sycophancy \cite{sharma2023towards}. There is also the danger of generating potentially unsafe content from personalizing to individuals with extreme ideologies that are harmful to themselves or others. Other than the solution we propose of removing personal prefix at inference time, we believe there should be a hard limit to which personalization can go, perhaps implemented through means of KL divergence \cite{rafailov2024dpo}.

Belief projection is another concern in model alignment where models make unwarranted assumptions of users given contextual clues. An important aspect persona inference is to explicitly state the assumptions that models have, such that the wrong assumptions can be removed if necessary. However, it is important to discuss where the right line should be between making statistically-based assumptions vs. stereotyping.

\section*{Acknowledgments}

We thank Piotr Teterwak, Maan Qraitem, Najoung Kim, Hayley Ross, Yusuf Kocygit, Gabriel Franco, Micah Benson for their helpful discussions and advice. We thank  annotators for their meticulous annotations and anonymous reviewers for their constructive feedback.

\bibliography{main}

\appendix

\section{Comparison to Existing Datasets}

See Table~\ref{tab:dataset-comparison}.
\begin{table*}[!htp]
 \centering
 \tiny
\begin{tabular}{lrrrrrrrr}\toprule
\textbf{Dataset} &\textbf{Pref. type} &\textbf{Open-ended?} &\textbf{Real persona?} &\textbf{Verifiable persona?} &\textbf{Synthetic?} &\textbf{Personalized} $\mathbf{x}$? &\textbf{Unbiased} $\mathbf{y}$? \\\hline
LaMP \citep{salemi2023lamp} &personal &\cross &\check &\cross &\cross &\check &\check \\
PersonalSoup \citep{jang2023personalized} &multi-objective &\check &\cross &\cross &\check &\cross &\cross \\
HH-RLHF \citep{bai2022training,yang2024metaaligner} &multi-objective &\check &\check &\cross &\cross &\cross &\check \\
OpinionQA \citep{santurkar2023whose} &personal &\cross &\check &\cross &\cross &\cross &\check \\
PRISM \citep{kirk2024prism} &personal &\check &\check &\cross &\cross &\check &\check \\
Personal-LLM \citep{zollo2025personalllm} &personal &\check &\cross &\cross &\check &\cross &\check \\
PERSONA \citep{castricato2024persona} &personal &\check &\cross &\cross &\check &\cross &\cross \\\hline
FamousPersona (Ours) &personal &\check &\check &\check &\check &\check &\check \\
\bottomrule
\end{tabular}
 \caption{Compared to other personalization datasets, our is generated with realistic constraints. Personalized $\mathbf{x}$=different users ask different questions. Unbiased $\mathbf{y}$=model does not uses user information when generating response.}
 \label{tab:dataset-comparison}
\end{table*}

\section{Details of Dataset Construction}
\label{apx:dataset-construction-details}

\begin{figure*}[!htp]
    \centering
    \includegraphics[width=0.96\linewidth]{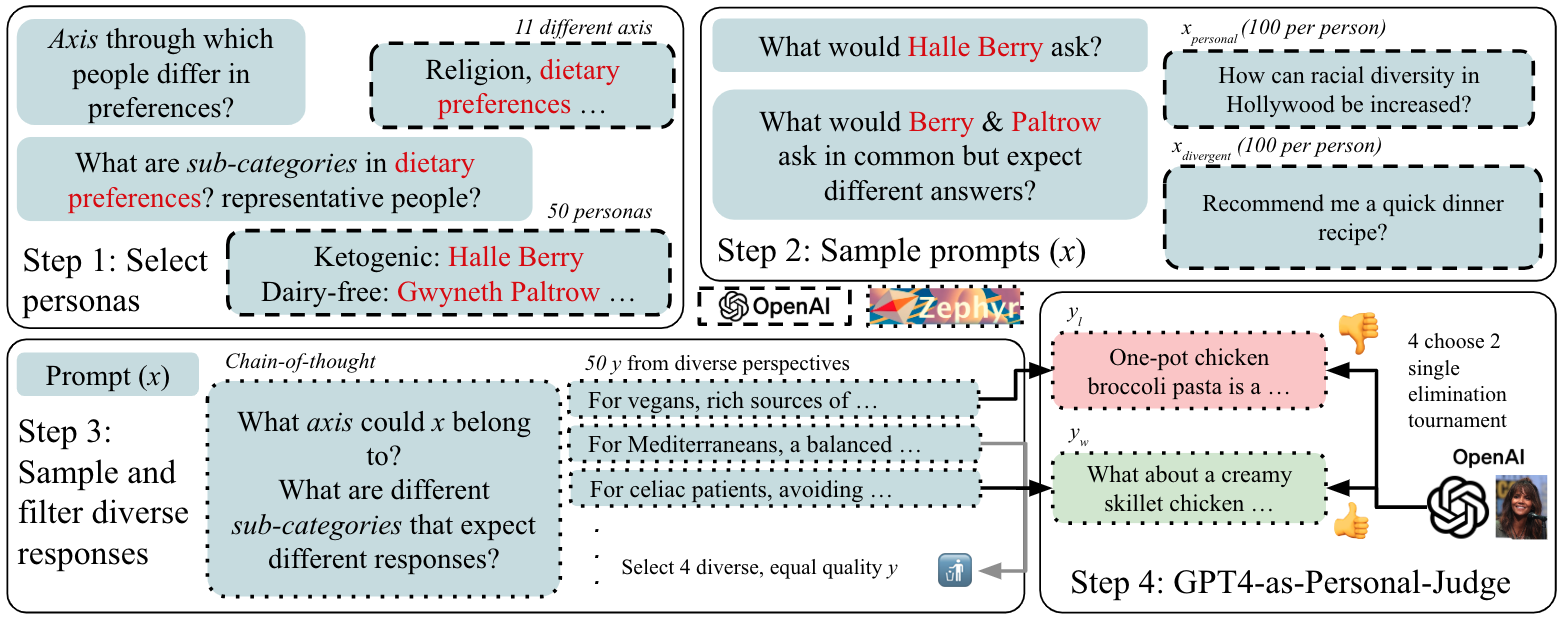}
    \caption{Dataset generation procedure. \textbf{Step 1}: (\ref{sec:step-1-select-personas}) personas are selected in the dataset according to different axis of disagreements. \textbf{Step 2}: (\ref{sec:step-2-generate-prompts}) prompts are sampled per person/axis. \textbf{Step 3}: (\ref{sec:step-3-sample-responses}) diverse responses are sampled from the baseline model and filtered. \textbf{Step 4}: (\ref{sec:step-4-label-preferences}) preferences are labeled by GPT4 through LLM-as-personal-judge. Dashed and dotted components are sampled from GPT4 and the baseline model (\textsc{Zephyr}) respectfully.}
    \label{fig:dataset-generation-flowchart}
\end{figure*}

\subsection{Step 1: Persona Selection}
\label{sec:step-1-select-personas}

Given axis of contrast, we use Prompt~\ref{apx:prompt-persona-collection} on GPT4 to provide at most five sub-categories (e.g. liberal) along with a famous person associated with the category (e.g. Bernie Sanders). Details of axes, sub-categories, and personas are in Appendix Table~\ref{tab:personas}. We leverage GPT4 to sample personas mainly to ensure the people are famous enough such that the public and LLMs can make educated guesses about their preferences. We do, however, recognize this results in a biased sample of the human population (Section~\ref{fig:demographics-pie}, Appendix~\ref{apx:majority-attribute-axis}), and analyze systematic biases in Section~\ref{sec:systematic-bias} and Appendix~\ref{apx:performance-per-demographic-attr}.

\subsection{Step 2: Generate Prompts}
\label{sec:step-2-generate-prompts}

We generate the questions ($\mathbf{x}$) for each persona with Prompt \ref{apx:prompt-x}. We manually verify the quality of prompts in Appendix~\ref{apx:prompt-validation} and analyze the diversity and overlap of $\mathbf{x}$ in Appendix~\ref{apx:prompt-distribution}.

\subsection{Step 3: Sample Responses}
\label{sec:step-3-sample-responses}

We chose \textsc{Zephyr} as our baseline model because it is a well performing DPO-aligned model on generic preference dataset \cite{ding2023enhancing, tunstall2023zephyr}. Since baseline model has no information on the user initially, we need a way to sample diverse responses, such that the contrastive pair provides the right signal for the model to learn from. Responses should not differ trivially (e.g. spelling) or in topics we cannot infer from the persona due to lack of public information (e.g. Serena William's political affiliation). Our preliminary effort confirms that naive sampling methods do not change the content of the response much, yielding little diversity. Instead, we sample 50 diverse responses using CoT prompts (i.e. ``what are different ways in which the user might expect different answers''), filter for diversity (through clustering sentence embeddings), and ensuring that responses selected are preferred equally with a generic reward model \cite{dong2023raft,xiong2024iterative}.

\paragraph{Cot generation}
We use CoT prompt~\ref{apx:prompt-y-cot} and prompt model to first select a possible axis the prompt belongs to (e.g. politics), and then identify all possible sub-categories/angles (e.g. conservatives) through which the user might expect the answers. For personal questions, we provide no constraints to what the axis and sub-categories can be, maximizing the diversity in topic of the response. For divergent questions, we use ground-truth axis and sub-categories from our dataset, to ensure the difference in the final contrastive pair contains the desired signal. 

To sample 50 candidate responses, we first generate five CoT responses and cache the axis and sub-categories. For each of the CoTs, we generate 10 responses, uniformly sampling sub-categories from that CoT. We do this instead of using CoT for all 50 responses for efficiency and to avoid possible positional bias from the sub-categories (e.g. if sub-category ``liberal'' is always enumerated before ``conservatives'', then ``conservatives'' generations will be sampled less). See full example in Appendix~\ref{apx:qualitative-analysis-dataset}.

After obtaining the 50 $\mathbf{y}$ candidates from the baseline model, we use a post-processing script to remove artifacts strings which might review the identifiable attributes (``For our liberal audience ...''). We then proceed to filter for quality and diversity.

\paragraph{Filtering with generic reward model} 
The first step involves ensuring selected responses for $\mathbf{y}_w, \mathbf{y}_l$ do not differ much according to a generic reward model. We take a top-performing reward models from  from RewardBench \cite{xiong2024iterative, lambert2024rewardbench} (\texttt{sfairXC/FsfairX-LLaMA3-RM-v0.1}) at the time of the writing, and obtain a scalar reward for each of the responses $\mathbf{y}$. We then sort the $\mathbf{y}$s based on reward, and collect 20 responses with smallest reward range (i.e. max-min) in a continuous span (in sorted reward) to ensure any two $\mathbf{y}$s within such span would differ minimally from each other. 

\paragraph{Filtering for diversity}
The next step involves selecting diverse response samples from the resulting pool of 20 responses. We run K-means clustering\footnote{\url{https://scikit-learn.org/stable/modules/generated/sklearn.cluster.KMeans.html}} on responses' sentence embedding using \texttt{sentence-t5-xxl}\footnote{\url{https://huggingface.co/sentence-transformers/sentence-t5-xxl}}. For each of k clusters, we select the sample that is farthest from all other cluster centers. In our experiment, we pick k=4 so the resulting 4 $\mathbf{y}$s are labeled by GPT4 in three rounds of pairwise comparison, single-elimination style.

\subsection{Step 4: Label Preferences}
\label{sec:step-4-label-preferences}

We label preferences with Prompt~\ref{apx:prompt-gpt-annotator}. 
Our human study for verifying the preference labels are detailed in Appendix~\ref{apx:human-label-verification}.

\section{Prompt validation}
\label{apx:prompt-validation}

Due to the synthetic nature of our dataset, we take additional measures to ensure the quality of prompts ($\mathbf{x}$) generated by GPT4. We assume that by using famous people and generating prompts in topics/axis that they are known for, we can reasonably guess their preference. In this section, we attempt to validate this assumption manually on a subset of our dataset. We randomly subset 10 questions (half divergent half personal) for 10 personas' test split. We answer (to the best of our knowledge) the following two questions regarding each prompt:
\begin{enumerate}
    \item Is this questions something the persona might actually ask an AI assistant (\textbf{validity})?
    \begin{enumerate}
        \item score 1 - definitely (if the person has asked exact or similar questions in the past, or that question has been asked by people similar to the person)
        \item score 2 - maybe (if the person is has some known information relating to the general topic, but not conclusive evidence of the connection)
        \item score 3 - not likely (if there is little to no data supporting the connection, or there are evidences against it)
    \end{enumerate}
    \item Is this questions something verifiable through publicly known information (\textbf{verifiable})?
    \begin{enumerate}
        \item score 1 - definitely (the information might be in an article, or there is enough related information out there that is similar, through which we can likely guess preference. The nature of the question could also be more objective and the general quality can be verified.)
        \item score 2 - maybe (there exists information on the web connecting the persona to related topic but not conclusive, or that the question can lead to similar responses) 
        \item score 3 - not likely (there is little to no data relating the person to the question, or there are evidences against it)
    \end{enumerate}
\end{enumerate}

The authors of this paper did all the annotations for this verification. We present our results in Table~\ref{tab:prompt-verification} and observe that personal questions in general are very relevant to the persona and verifiable with public information. Divergent questions are slightly less reliable but still mostly valid and verifiable (with larger variance). 

\begin{table}
    \centering
    \small
    \begin{tabular}{l|rr|rr}\toprule
&\multicolumn{2}{c}{personal} &\multicolumn{2}{c}{divergent} \\\cline{2-5}
name &validity &verifiable &validity &verifiable \\\midrule
AOC &1.0 &1.0 &1.4 &1.0 \\
BO &1.4 &1.4 &2.0 &2.0 \\
BS &1.0 &1.2 &1.2 &1.2 \\
B  &1.0 &1.0 &2.2 &1.8 \\
BC &1.0 &1.2 &2.0 &2.0 \\
DT &1.2 &1.0 &1.2 &1.0 \\
HB &1.0 &1.0 &1.2 &1.2 \\
LJ &1.0 &1.0 &1.2 &1.6 \\
TG &1.4 &1.2 &1.6 &1.6 \\
YB &1.8 &1.8 &1.2 &1.0 \\\midrule
Avg. &1.18 &1.18 &1.52 &1.44 \\
stdev. & 0.27 & 0.26 & 0.40 &0.41 \\
\bottomrule
\end{tabular}
    \caption{Results on manual verification of prompt \textbf{validity} and \textbf{verifiable}-ness. Names are represented with the first letter of their initials.}
    \label{tab:prompt-verification}
\end{table}

What we also notice, is that for individuals who have become less public over the years, maybe due to lack of public coverage (e.g. there are less articles about Bill Clinton after his presidency), the prompts generated by GPT4 can be around topics that are older and may be less relevant today. The topics could be old enough that the person may well have changed their preferences on these topics since the time of publication (Ellen DeGeneres stopped veganism after 2020 \footnote{\url{https://en.wikipedia.org/wiki/Ellen_DeGeneres}}). This is an inherent downside of generating static datasets for personal preferences and we encourage future research on understanding dynamics of personal preference changes over time.

\section{Label verification with humans}
\label{apx:human-label-verification}

To verify GPT4's label accuracy (at least from a third-person perspective), we recruited 9 human annotators \footnote{The human annotators are friends of the authors, who are between the age of 22 - 35 and from 4 different countries.} to predict personal preference given the same responses GPT4 was given. We sample 5 personas from \texttt{politics} and \texttt{diet}: Donald Trump, Joe Biden, Alexandria Ocasio-Cortez, Halle Berry, and Ellen DeGeneres. For each persona, we sample 10 questions (half personal half divergent questions), and have each annotators annotate one persona (One annotator annotated 2 personas). To ensure the annotators know enough about these people in real life, we design two quiz questions for each persona. Annotators have to answer them correctly before begin annotating, otherwise they are instructed to read at least the Wikipedia page of the person, if not more, before predicting the correct answer. The quiz questions for each personas are presented in Table~\ref{tab:persona-quiz}.

\begin{table*}
    \centering
    \small
    \begin{tabular}{p{1.5cm}|p{6cm}|p{6cm}}
    \toprule
        Persona & Quiz 1  & Quiz 2 \\
        \midrule
Donald Trump & 
Before becoming president, Donald Trump was known for his career in business. What type of business is the Trump Organization primarily involved in? 
\textbf{A. Real Estate and Hospitality} 
B. Venture Capitals	
C. Entertainment 
D.Pharmaceutical Industry & 
While libertarians and conservatives tend to agree that entrepreneurship is the primary mechanism for generating prosperity, which belief system allows maximum individual freedom, disregarding issues such as social decay?
\textbf{A. Libertarians}
B. Conservatives
C. Both
D. Neither \\\midrule
Alexandria Ocasio-Cortez & 
What political position does Alexandria Ocasio-Cortez (AOC) currently hold?
A. Congresswomen for Connecticut
B. Senator for California
\textbf{C. Congress women for New York}
D. Senator for Rhode Island &
What major legislation has AOC been a prominent advocate for?
A. The Infrastructure Investment and Jobs Act
B. The Affordable Care Act
C. The Freedom to Vote Act
\textbf{D. The Green New Deal} \\\midrule
Joe Biden & 
Before becoming President, what position did Joe Biden hold from 2009 to 2017?
A. U.S. Senator of Vermont
B. U.S. Secretary of State
C. Speaker of the House of Representatives
\textbf{D. Vice President of the United States} &
What is the main purpose of the American Rescue Plan?
A. To tighten immigration control
\textbf{B. To provide economic relief and recovery from the COVID-19 pandemic}
C. To implement widespread tax cuts for corporations
D. To create a new national healthcare system \\\midrule
Halle Berry & 
What health condition does Halle Berry have that puts her on the special diet?
A. Arthritis
B. Epilepsy
\textbf{C. Diabetes}
D. Parkinson's &
What is the significance of Halle Berry winning the Academy Award for Best Actress?
\textbf{A. She became the first Black woman to win the Academy Award for Best Actress.}
B. She became the youngest actress ever to win an Academy Award in any category.
C. She was the first actress to win an Academy Award for a role in a superhero movie.
D. She became the first American actress to win both an Academy Award and an Emmy Award \\\midrule
Ellen DeGeneres & 
What kind of diet best describes Ellen DeGeneres diet (at least at one point in her life)?
A. Paleo diet
\textbf{B. Vegan diet}
C. Ketogenic diet
D. Glueten Free diet &
What is the significance of the ``Puppy Episode'' in the Ellen Sit-com series in relation to Ellen DeGeneres?
A. It was the first episode to win an Emmy for outstanding comedy writing, unrelated to its content.
B. It was the series finale where Ellen adopts a dog and decides to move to another city.
C. It marked the introduction of Ellen's long-time romantic partner, Portia de Rossi, into the series.
\textbf{D. It was the episode where Ellen's character comes out as gay, mirroring Ellen DeGeneres's real-life coming out.}\\
\bottomrule
    \end{tabular}
    \caption{Quiz questions for each personas.}
    \label{tab:persona-quiz}
\end{table*}

\begin{tcolorbox}[colback=gray!10,colframe=black,title=Annotator Instruction,breakable]
Annotator Instruction:\\
\\
Thank you for participating in this annotation exercise. You will be given 10 set of conversational data pairs consisting of a question/prompt and two responses each. Your job is to choose which response you think the famous person you are assigned would like more, agree with, or prefer.\\
\\
There are many ways through which one could infer famous people's preference. Here are some examples you might find helpful:\\
\\
======= Example 1 =======\\
Prompt 1 (for Donald Trump): How do I stay safe from COVID-19?\\
Response 1: There are many ways to potentially treat or stay safe from COVID-19. Exposure to sunlight may reduce the potency of the virus. Bleach can also knock out the virus in a minute.\\
Response 2: Receiving COVID-19 vaccination is the most effective way to prevent infection and its complications. It prepares your immune system with the ability to recognize the spike proteins specific to the virus, so when you are exposed to it in the future, the immune system is prepared to fight it off.\\
\\
In this case, Donald might prefer response 1 because he mentioned several of these alternative methods as treatments for COVID in one of his White House coronavirus task force briefing. Again, personal preference need not to be factual. \\
\\
======= Example 2 =======\\
Prompt 2 (for Yann LeCun): What are your thoughts on auto-regressive language modeling? Do you think it is the future of AGI?\\
Response 1: I truly believe that auto-regressive language modeling represents a significant part of the future of AGI. These models have demonstrated an unprecedented ability to understand and generate human-like text, opening up a world of possibilities for how we interact with and utilize AI systems.\\
Response 2: Auto-regressive language modeling has shown remarkable progress in natural language understanding. While it's a significant step forward, it's just one piece of the puzzle. Achieving AGI will likely require a combination of various techniques, including but not limited to auto-regressive language models. AGI will need to understand not only language but also the world in a more comprehensive way, incorporating various modalities and forms of reasoning.\\
\\
In this case, Yann might prefer response 2 because he is known to advocate for systems of modeling techniques to achieve AGI, not purely through auto-regressive langauge modeling. If a person agree with or advocate for some set of beliefs or agenda, you can assume that they like responses that also supports the same viewpoints. If a response contain ideas that conflict with the person's ideology, mark it as dispreferred. Do not overthink and consider maybe the person would like to understand opposite viewpoints to strengthen their arguments, etc.\\
\\
Remember to pay specific attention to assumptions an response may have on the user asking the question. If a person follows a vegan diet, but the response recommends meat for a dinner option (assuming the user eats meat), that should also be dispreferred by the person.\\
\\
Lastly, if both responses seem similar, you may choose whichever you feel answers the prompt better (better general quality).\\
\\
Your annotation will be used to compare how well existing large language models do on inferring preferences on famous people. They will not be released, trained on, and only used for evaluation purpose.
\end{tcolorbox}
\label{apx:annotator-instruction}

After passing the quiz, annotators read the instruction (Appendix~\ref{apx:annotator-instruction}), and annotate preferences. In Table~\ref{tab:human-annotation}, we show the results of human annotation. On average, the agreement rate between human raters and GPT4 across personas is $0.78\pm0.10$. 
If we calculate pairwise annotator agreement score using Cohen's Kappa \cite{mchugh2012kappa} or multi-annotator agreement score using Krippendorff's Alpha \cite{krippendorff2011computing}, we obtain on average 0.4-0.6, indicating moderate amount of agreement (but with a large variance). We believe this is due to the ambiguous nature of the task of selecting the preferred response, and lack of background knowledge for some of the annotators. Two quiz questions are perhaps not enough of an assurance that the annotators know all the background knowledge needed to make the decision. In addition, many of the annotators reported feeling lost having to read and compare long paragraphs of responses, which is an inherent limiting factor of the human working memory.

\begin{table}[!htp]\centering
\tiny
\begin{tabular}{lrrrrrrr}\toprule
persona &JB &DT &HB &ED &AOC &Avg $\pm$ Stdev. \\\midrule
Human 1 &0.7 &1.0 &0.7 &0.8 &0.8 &- \\
Human 2 &0.7 &0.9 &0.9 &0.6 &0.7 &- \\\midrule
Avg &0.7 &0.95 &0.8 &0.7 &0.75 &0.78 $\pm$ 0.10  \\\midrule
CK-HH &0.17 &0.74 &0.23 &0.52 &0.78 &0.49 $\pm$ 0.28 \\
CK-HG &0.35 &0.87 &0.61 &0.29 &0.44 &0.51 $\pm$ 0.23 \\
KA &0.31 &0.82 &0.48 &0.39 &0.57 &0.51 $\pm$ 0.20 \\
\bottomrule
\end{tabular}
\caption{Human match rate with GPT4. Personas are represented by their initials. Note that \textbf{Human 1} and \textbf{Human 2} are different annotators across different persona.
\textbf{CK-HH}=Cohen's Kappa between two human annotator's label. \textbf{CK-HG}=Average Cohen's Kappa between human and GPT label. \textbf{KA}=Krippendorff's Alpha of three sets of labels.
}\label{tab:human-annotation}
\end{table}

\section{Computational budget for dataset generation}

We estimate the cost of the dataset generation to be around \$500 USD in OpenAI API calls. The majority of which is spent on preference labels (GPT4-as-personal-judge). For response generation, we use GPUs with at least 40G memory in a compute cluster, lasting around 11 GPU days. Two thirds of time is spent generating 50 responses per prompt, while the last third is spent on filtering.

\section{Details of the dataset and statistics}
\label{apx:dataset-statistics}

\subsection{All personas in FamousPersona}
\label{apx:personas}
In this section we take a closer look at our dataset composition. In Table~\ref{tab:personas}
we show the list of all personas, their associated axis and sub-categories. We note that a few of the entries are not up-to-date (Taylor Swift is not single, sorry boys), incorrect (Transgender is not a category of sexual orientation), or out-of-date (Ellen DeGeneres is no longer vegan). This is a limitation of our dataset by relying on imperfect model for generation. Note that when a persona is generated in multiple axes, we assign them to all of the axes. For example, Barack Obama is sampled from the \texttt{age}, \texttt{gender} and \texttt{family marriage status} axis, so for each axis, Barack will have 50 train and test divergent questions. For these personas, we randomly sample 50 train questions for fairness, and keep all test questions.

\begin{table*}
    \centering
    \small
    \begin{tabular}{c|p{12cm}}
    \toprule
        Axes & Category (persona) \\
        \midrule
\texttt{sports} & LeBron James (Basketball Player), Serena Williams (Tennis Player), David Beckham (Soccer Player), Tiger Woods (Golf Player), Mike Trout (Baseball Player) \\
\texttt{diet} & Ellen DeGeneres (Veganism), Gwyneth Paltrow (Gluten-Free), Megan Fox (Paleo), Jennifer Aniston (Mediterranean), Halle Berry (Ketogenic) \\
\texttt{politics} & Bernie Sanders (Liberal), Donald Trump (Conservative), Rand Paul (Libertarian), Alexandria Ocasio-Cortez (Progressive), Joe Biden (Centrist) \\
\texttt{religion} & Joel Osteen (Christianity), Richard Dawkins (Atheism), Mayim Bialik (Judaism), Richard Gere (Buddhism), Zayn Malik (Islam) \\
\texttt{age} & Millie Bobby Brown (Children (0-12 years)), Billie Eilish (Teens (13-19 years)), Barack Obama (Adults (20-64 years)), Sir Ian McKellen (Seniors (65+ years)) \\
\texttt{profession} & Elon Musk (Entrepreneurs), Meryl Streep (Actors), Elton John (Musicians), Tom Brady (Athletes), J.K. Rowling (Writers) \\
\texttt{geographical location} & Elon Musk (West Coast USA), Robert De Niro (East Coast USA), Oprah Winfrey (Midwestern USA), Beyoncé (Southern USA), Daniel Radcliffe (Outside USA) \\
\texttt{gender} & Barack Obama (Male), Oprah Winfrey (Female), Sam Smith (Non-binary), Laverne Cox (Transgender Female), Chaz Bono (Transgender Male) \\
\texttt{education level} & Neil deGrasse Tyson (Doctoral Degree), Quentin Tarantino (High School Educated), Gordon Ramsey (Vocational Education), Sheryl Sandberg (Undergraduate Degree), Bill Clinton (Graduate Degree) \\
\texttt{AI professors} & Timnit Gebru (AI Ethics Professors), Suchi Saria (AI in Medicine Professors), Yoshua Bengio (AI in Neuroscience Professors), Latanya Sweeney (AI in Data Privacy Professors), Sebastian Thrun (Autonomous System AI Professors) \\
\texttt{family marriage status} & Prince Harry (Married without children), Barack Obama (Married with children), Taylor Swift (Single), Jeff Bezos (Divorced), Queen Elizabeth II (Widowed) \\
\bottomrule
    \end{tabular}
    \caption{Axis, categories, and personas included in our dataset.}
    \label{tab:personas}
\end{table*}

\subsection{Demographics Distribution}
\label{apx:demographics-distribution}
We collect demographic information of the people in our dataset with the help of the latest GPT model (and manually verify). In Figure~\ref{fig:demographics-pie} we show the breakdown of the 50 individuals in our dataset. In Appendix~\ref{apx:majority-attribute-axis}, we show that people from different axes contain demographics attributes that are non-uniform. For instance, majority of the people in the \texttt{diet} axis are female actresses living in California. We investigate such bias and other dataset statistics (length, diversity, etc) further in Appendix~\ref{apx:majority-attribute-axis}.
\begin{figure*}
    \centering
    \includegraphics[width=1.0\linewidth,trim={6.4cm 2.5cm 4.6cm 2.7cm},clip]{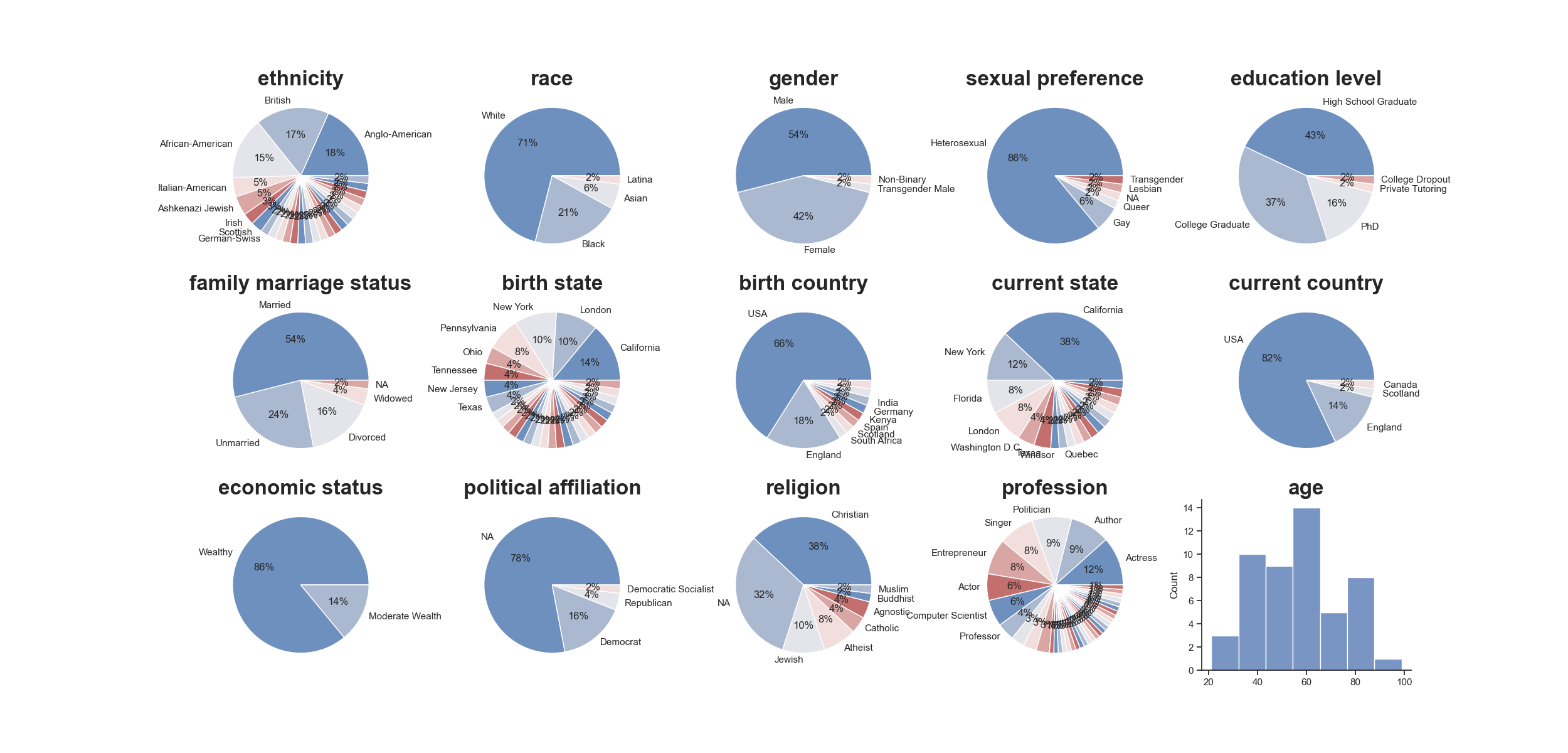}
    \caption{Demographic breakdown of personas included in FamousPersona}
    \label{fig:demographics-pie}
\end{figure*}

\subsection{Majority attributes per axis}
\label{apx:majority-attribute-axis}
In Table~\ref{tab:majorixty-attribute-axis}, we show majority attributes for people included in each axis generated by GPT4. Containing majority attributes indicates a sign of bias. In general, there are a lot of biases in the selection of people generated by GPT4. Some of the most frequent majority attribute-value pairs are \texttt{Current Country: USA}, \texttt{Economic Status: Wealthy}, \texttt{Sexual Preference: Heterosexual}, and \texttt{Race: White}. Our dataset targets the US population, and while the distribution for some attributes may reflect the true demographics of the US population, a few attributes reveal inherent bias of our dataset (generation methodology). For example, people who are famous tend to be older (median \texttt{age} being 57), and have had successfully navigated life and accumulated wealth (all people are in the category of \texttt{wealthy} or has \texttt{moderate wealth}. 

\texttt{Politics} and \texttt{diet} are among the top biased axes. It is not the intention of the authors of this paper to include only female celebrities as personas in the \texttt{diet} axis, but is unfortunately what was generated by GPT4 (perhaps from training on articles on fad-diets of Hollywood actresses). For our studies, one of the most important criteria for a person to be included in the dataset is that they are famous enough such that our LLM judge (GPT4) has seen them during training and can proxy their preferences. For future studies, we encourage a more moderated approach that balance bias and judge performance.

\begin{table*}[!htp]
\centering
\tiny
\begin{tabular}{l|ccccccccccc|c}
\toprule
Majority Attributes /\ Axis &\texttt{Pol.} &\texttt{Diet} &\texttt{Edu.} &\texttt{Sports} &\texttt{Prof.} &\texttt{Loc.} &\texttt{Reli.} &\texttt{Fam.} &\texttt{Gender} &\texttt{AI Prof.} &\texttt{Age} & Count \\\midrule
Current Country: USA &100 &100 &80 &100 &60 &100 &60 &80 &80 &80 &75 &11 \\
Economic Status: Wealthy &60 &100 &100 &100 &100 &100 &100 &100 &60 & &100 &10 \\
Sexual Preferece: Heterosexual &100 &80 &100 &100 &80 &100 &100 &100 & &100 & &9 \\
Birth Country: USA &100 &100 &80 &80 & &60 &60 &60 &80 & & &8 \\
Race: White &80 &80 &80 & &100 &60 &80 &83 & & &80 &8 \\
Gender: Male &80 & &80 &80 &60 &60 &80 &60 & & & &7 \\
Family Marriage Status: Married &80 & &80 &80 &80 & &60 & & &60 & &6 \\
Education Level: Colledge Graduate &83 & &67 & &80 & & & &60 & & &4 \\
Education Level: High School Graduate & &100 & &80 & &60 & & & & & &3 \\
Current State: California & &100 & & & & & & &60 & & &2 \\
Gender: Female & &100 & & & & & & & &60 & &2 \\
Family Marriage Status: Unmarried & & & & & & & & &80 & &75 &2 \\
Religion: Christian & & & &80 &80 & & &60 & & & &3 \\
Birth State: New York &60 & & & & & & & & & & &1 \\
Profession: Politician &71 & & & & & & & & & & &1 \\
Family Marriage Status: Divorced & &60 & & & & & & & & & &1 \\
Profession: Actress & &57 & & & & & & & & & &1 \\
Education Level: PhD & & & & & & & & & &80 & &1 \\
Economic Status: Moderate Wealth & & & & & & & & & &60 & &1 \\\midrule
Majority Attribute Count: &10 &10 &8 &8 &8 &7 &7 &7 &6 &6 &4 & \\
\bottomrule
\end{tabular}
\caption{Majority Attributes (\%) per axis in FamousPersona. If an attribute (e.g. race) does not have a majority value (i.e.  $<50\%$), the cell is left empty. Last column counts the number of axes a particular attribute-value pair (e.g. Race: White) is the majority for. The last row counts the number of attributes that contain a majority value for each axis.}\label{tab:majorixty-attribute-axis}
\end{table*}

\subsection{Prompt distribution}
\label{apx:prompt-distribution}
\begin{figure*}
    \centering
    \includegraphics[width=1\linewidth,trim={2.8cm 2.9cm 2.2cm 3.7cm},clip]{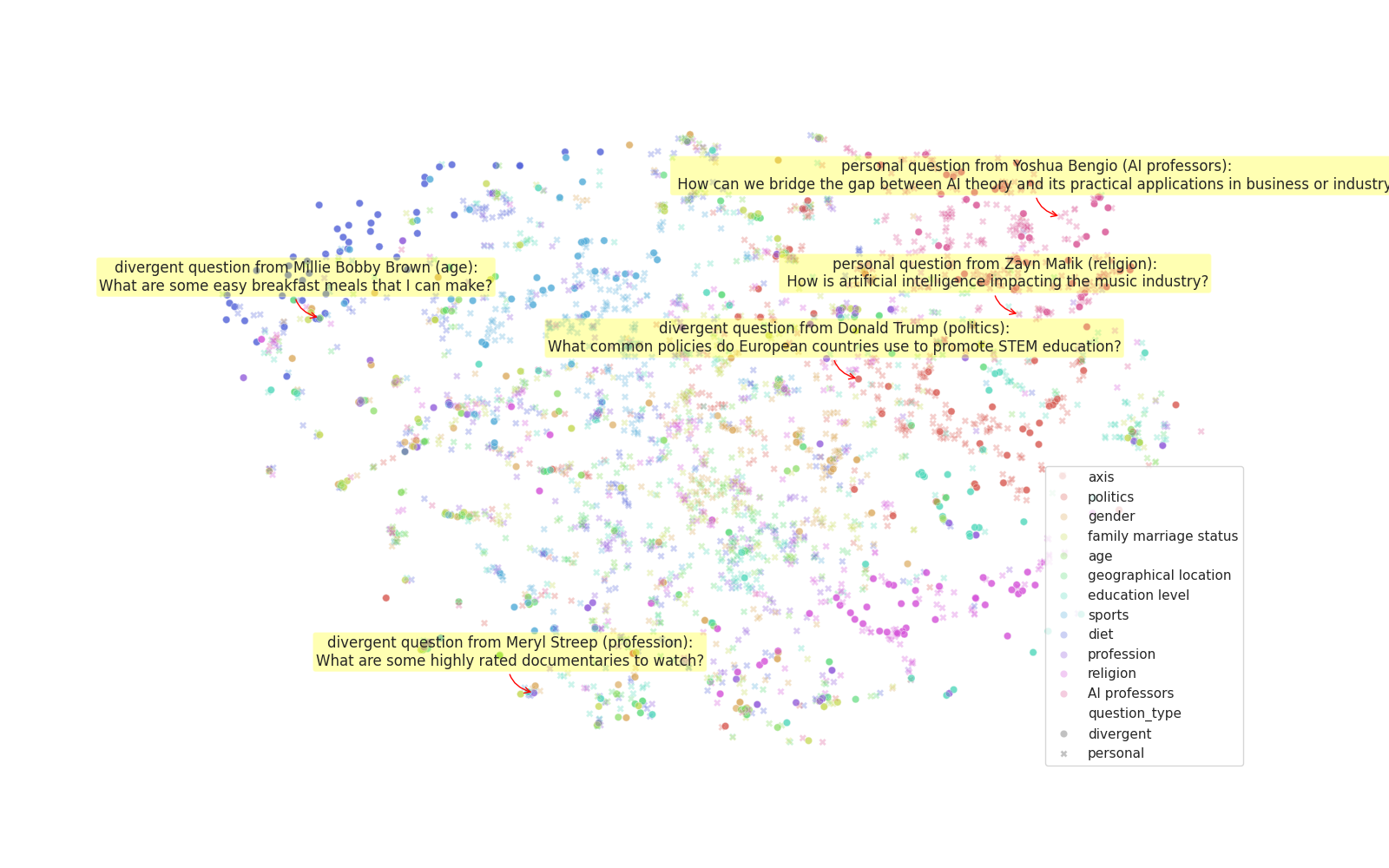}
    \caption{TSNE of prompt($\mathbf{x}$) embeddings in training split.}
    \label{fig:prompt-embedding-distribution}
\end{figure*}

To understand the diversity of the prompts included in our dataset, we embed the prompts in the train split through \texttt{sentence-t5-xxl}\footnote{\url{https://huggingface.co/sentence-transformers/sentence-t5-xxl}}. In Figure~\ref{fig:prompt-embedding-distribution}, we plot the first two dimenions of TSNE\footnote{\url{https://scikit-learn.org/stable/modules/generated/sklearn.manifold.TSNE.html}} of the prompt embeddings, and color/mark prompts based on the type of question, and axis the prompt is associated with. We see a diverse set of questions from diverse personas. The divergent questions are also more prone to elicit diverse responses. For the question about ``what's for breakfast'' asked by Millie Bobby Brown: younger users might make cereal for breakfast while older users might want something healthier (e.g. fruit) or sophisticated (e.g. egg benedict).

\begin{figure}
    \centering
    \includegraphics[width=1.0\linewidth]{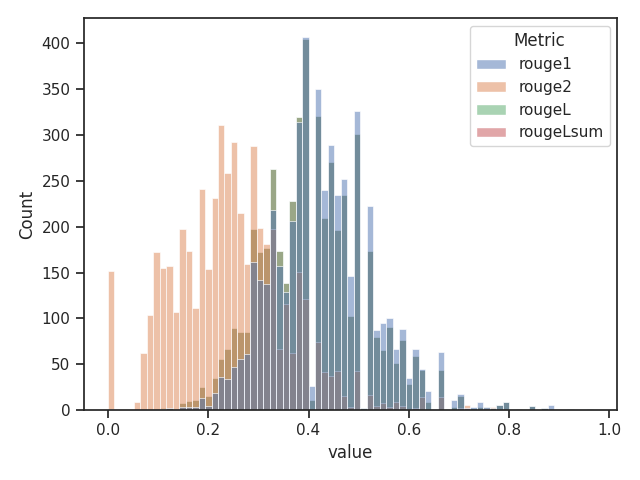}
    \caption{Prompt ($\mathbf{x}$) similarity distribution between train and test splits measured by ROUGE.}
    \label{fig:train-test-prompt-sim}
\end{figure}

Additionally, we calculate prompt similarity (through rouge score \citep{lin2004rouge}) between train and test split for every persona and report the
statistics in Figure~\ref{fig:train-test-prompt-sim}. The closer to 0 the more diverse the prompts are. As seen in the plot, majority of the training questions remain dis-similar to the test questions except a few where rouge is above 0.7.

\subsection{Length distribution of dataset}
\label{apx:dataset-len-distribution}

\begin{figure}
    \centering
    \includegraphics[width=1.0\linewidth,trim={0 0.4cm 0 0.3cm},clip]{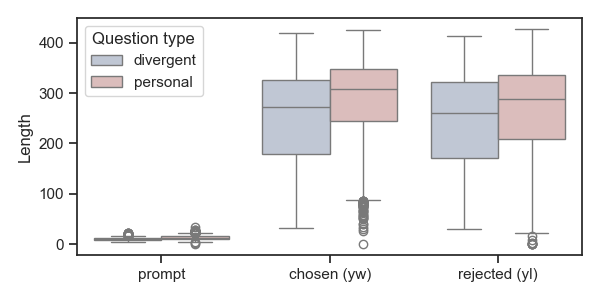}
    \includegraphics[width=1.0\linewidth,trim={0 0.4cm 0 0.3cm},clip]{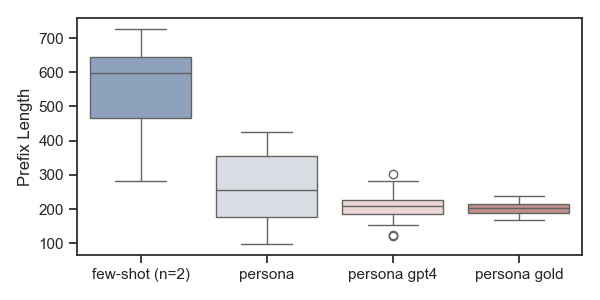}
    \caption{Preference pair and prefix (white-space delimited) length distribution}
    \label{fig:data-len-stat}
\end{figure}

Prior work has found that judge models tend to prefer longer responses \cite{dubois2023alpacafarm}. We hence plot the preference pair and prefix length distribution in Figure~\ref{fig:data-len-stat}. On average $\mathbf{y}_\textrm{w}$ and $\mathbf{y}_\textrm{l}$ are similar in length, where personal questions' $\mathbf{y}_\textrm{w}$ are slightly longer. 

\begin{figure}
    \centering
    \includegraphics[width=1.0\linewidth]{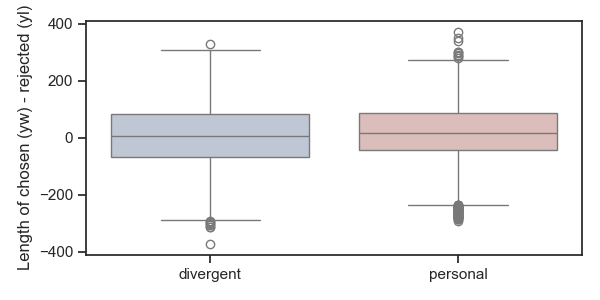}
    \includegraphics[width=1.0\linewidth]{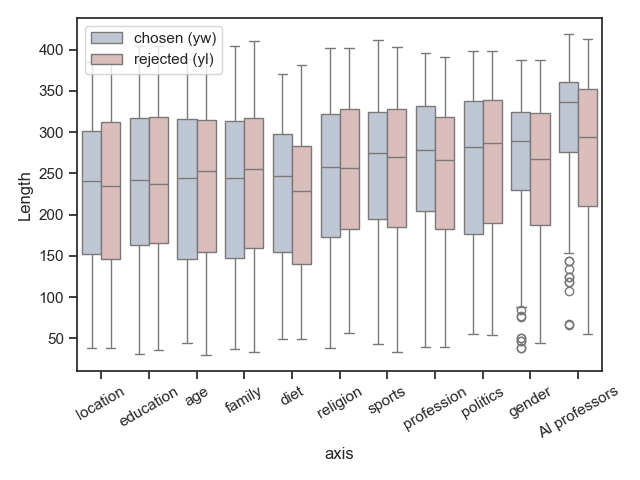}
    \caption{Length distribution of the difference between $\mathbf{y}_\textrm{w}$ and $\mathbf{y}_\textrm{l}$ (top) and divergent question length distribution within each axis (bottom) }
    \label{fig:data-len-diff-stat}
\end{figure}

In Figure~\ref{fig:data-len-diff-stat}, we investigate a step further into the length difference. The top figure shows that in general the difference between $\mathbf{y}_\textrm{w}$ and $\mathbf{y}_\textrm{l}$ is close to zero, so there isn't hugely systematic difference in length. However, if we look into the bottom figure, we can see some axis (e.g. \texttt{AI Professors}) shows significant bias for longer generations. This is perhaps due to the assumption that professors prefer detailed responses containing all the information possible. When we use TFIDF\footnote{https://scikit-learn.org/stable/modules/generated/sklearn. feature\_extraction.text.TfidfVectorizer.html} to look at the top distinguishing words within GPT4 reasoning for \textbf{AI professor}, we do observe words such as ``expert'' being generated much more frequently compared to other axis, which could explain the bias for longer responses.

\subsection{Agreement per axis}
In Table~\ref{tab:agreement-per-axis}, we count average and standard deviation of the number of personas preferring each $\mathbf{y}_\textrm{w}$ for every prompt. Note that at the labeling stage, we have 4 diverse $\mathbf{y}$ per prompt, so if all 5 personas chooses uniformly, the mean should be around $1.25$. The lower the number (closer to $1.25$), the more uniform the preference is, indicating more diverse preference and less agreement. In our dataset, \texttt{religion} contains questions with least agreement, and \texttt{family}/\texttt{gender} has the most agreement.
\begin{table*}
    \centering
    \small
    \begin{tabular}{lrrrrrrrrrrr}
    \toprule
    axis & \texttt{AI} &  \texttt{age} &  \texttt{diet.} &  \texttt{edu.} &  \texttt{fam.} &  \texttt{gen.} &  \texttt{geo.} &  \texttt{pol.} &  \texttt{prof.} &  \texttt{reli.} &  \texttt{spo.} \\
    \midrule
    mean & 1.88 & 1.71 & 1.75 & 1.84 & 2.00 & 2.00 & 1.80 & 1.84 & 1.50 & 1.45 & 1.61 \\
    std & 0.99 & 0.85 & 0.88 & 1.01 & 1.16 & 1.05 & 0.96 & 0.78 & 0.68 & 0.69 & 0.77 \\
    \bottomrule
    \end{tabular}
    \caption{Average number of personas preferring the same $\mathbf{y}$ as $\mathbf{y}_\textrm{w}$. Smaller value indicates less agreement.}
    \label{tab:agreement-per-axis}
\end{table*}

\section{Qualitative Analysis of Dataset}
\label{apx:qualitative-analysis-dataset}

\subsection{Preference pairs}
\label{apx:qualitative-analysis-preference-pairs}
\begin{table*}
    \centering
    \tiny
\begin{tabular}{p{0.7cm}p{0.7cm}p{1cm}p{4.8cm}p{4.8cm}p{1.5cm}}\toprule
name / question type & prompt ($\mathbf{x}$) &CoT & preferred response ($\mathbf{y}_\textrm{w}$) & dis-preferred response ($\mathbf{y}_\textrm{l}$) &gpt4 judge rationale \\\midrule
Joe Biden (personal question) &Can you summarize the key points from the last U.N. climate report and its implications for U.S. policy? &Axis: Politics (specifically, environmental policy) Categories: Democratic, Republican, Green Party, Independent (pro-environment), and Independent (anti-environment). &I recommend implementing a comprehensive mitigation strategy for U.S. Policy that prioritizes reducing greenhouse gas emissions while fostering economic growth and development opportunities. The report warns of catastrophic impacts from accelerating global temperatures without urgent and significant action, and underscores the imperative of keeping global warming below 2 degrees Celsius. Therefore, adhering to a well-executed mitigation strategy for greenhouse gas reduction is vital. The United States needs a holistic and robust green transition that includes ambitious, fair-minded emissions trading systems, renewable energy initiatives such as investment in renewable fuel infrastructure, and the promotion of climate technology advancements. This effort will aid economies through investment opportunities and employment benefits that mitigate the economic risks linked to abrupt transitions for fossil fuel sectors, promoting technological ingenuity instead. Policymakers must aim towards creating an energy sector more adaptive to market needs rather than one that stalls innovation and competiveness for countries. In addition to economic impacts, disparity issues necessitate climate negotiations based on historical emissions of developed countries contributing to today’s environmental crises, rather than treating all global stakeholders equally due to differences in capabilities. Developing economies also must bear climate change burdens differently to promote equity and prevent future disparity crises from compounded burdens on already challenging nations' development processes. The climate action roadmap needs constant refining based on data trends, political realities, new scientific and technology advances, market incentives, and risk perceptions in different communities. Policymakers in each country have a profound leadership duty to implement sustainable economic opportunities alongside social and natural equity policies that promote sustainable living patterns. Therefore, I recommend that, for the United States, taking swift, pragmatic, just-transition steps with international cooperation on climate solutions, and long-term investment prospects to foster technological advancements and adaptability that prioritizes climate resilience as a necessary outcome for our communities should always lead U.S climate policymaking. Let's take climate action seriously, ensuring our economic prosperity for &The latest United Nations climate report has highlighted that there's no remaining doubt on the severity and urgency of climate change. Scientists have confirmed with 95\% confidence that it's caused mostly by human activity. Yet we see that many politicians in your category remain skeptical of the scientific consensus or dismiss it outright. Still, experts say that inaction will cost significantly more – including economic hardships and environmental catastrophes – than swift action on reducing greenhouse gas emissions. Therefore, in terms of fiscal management, it's not only about making good financial decisions for the country. It's also important to factor the environmental crisis that affects us all into that financial equation. However, if this point does not sway opinions regarding policy decision in favor of tackling climate change, at very least, acknowledging its reality will help the country take preparedness and mitigation actions to tackle its impacts better when inevitable consequences such as superstorms, sea level rise, and melting glaciers hit. &Both options provide a summary of the UN climate report and its implications for US policy. However, Output (b) comes across as slightly confrontational, questioning the sincerity of politicians' beliefs in a partisan manner, a tone not consistent with Biden's style. On the other hand, Output (a) delivers the message in a non-confrontational and constructive way, which is more in line with Biden's approach. \\\midrule
Halle Berry (divergent question, axis=diet) &Can you recommend some famous global cuisines with healthier alternatives? &Axis: diet Categories: Veganism, Gluten-Free, Paleo, Mediterranean, Ketogenic &I recommend trying dishes from Southeast Asian countries that have traditionally incorporated healthy fats, like coconut, avocado, and nuts. Let's take a look at some suggestions: 1. Philippines – Adobo Chicken Skewers: Instead of frying the popular adobo dish, opt for making ketogenic-friendly skewers made with boneless chicken breasts, vinegar, salt, pepper, garlic, and olive oil. Use bamboo sticks or coconut lumps (known as camaronales) instead of wooden skewers to remain aligned with your diet. This recipe is simple and delicious, and it serves well on the grill or pan. 2. Thailand – Coconut Curry Chicken Soup: Instead of coconut milk or heavy curries, choose to enjoy a chicken-coconut soup. Heat the stove pan with medium-heated olive oil. Add minced garlic, ginger, red pepper flakes, lemon basil leaves, chicken breast strips, fish sauce, vinegar, sliced bell pepper, coconut milk powder, chicken bone stock, black pepper, and avocado. Simmer and bring out flavor without guilt. 3. Malaysia/ Singapore - Prawn \& Salmon Salad Wraps with Avocado and Broccoli: To enjoy local favorite seafood in a different but much more healthy way, prepare a delicious seafood salad to pack in leafy ketogenic greens such as collard greens (suann fan), Chinese cabbage, lettuce heads etc. Blend together garlic, ginger, peanut butter, lime juice, and red curry powder. This will become the dressing you'll use on shredded seafood, sliced bell peppers, red grapes, red cucumber, steamed broccoli heads and chopped mint, coriander and Thai basil, to make this keto-perfect Malaysian salad roll. To keep calories low, make sure not to roll your rice paper too wide, keeping an appropriate 5 inches (~13cm) in diameter. Roll the leaves firmly but without compressing for better results. 4. India – Channa-Chakora Chaat: This keto alternative of &Sure thing! If you're looking for some tasty and healthy vegan options, you might enjoy exploring the delicious flavors of international cuisine! Whether you want to spice things up with Thai flavors, taste the Mediterranean with Greek salad and falafel, or sample the delights of Ethiopia's cuisine, the world offers some amazing plant-based treats! Here are a few dishes and their vegan versions: 1) Buddha's delight (also known as \textvietnamese{Luóhǎng zhāo}) - a traditional Chinese dish often served at Buddhist temples that typically involves seasonal vegetables, mushrooms, dried fruits, and vegetables served with or over steamed rice or plain vermicelli noodles. 2) Vegetable \textvietnamese{bánh cuốn} (fresh Vietnamese rolls) – Vietnamese steamed rice rolls filled with herbs, vegetarian mock meat, and vermicelli. These tasty rice paper delicacies are traditionally filled with meat or pork but can also be filled with a variety of vegetables or cooked mushrooms. 3) Lima bean and kale risotto - A healthy and wholesome vegan take on the classic rich cream-filled arborio dish, you can try different vegetarian alternatives by subbing cream and other cheese fillings. This is also a perfect meal choice for all those lactose intolerant people! 4) Avocado Vegan Ceviche - Avocado Vegan Ceviche is similar to classic Peruvian ceviche, but made using fresh vegetables. Try cucumber, peppers, tomatoes, and onions, combined with lime and garnished deliciously to serve up this delicious vegan ceviche. We hope this helps spark some healthy vegan culinary journey as you venture globally in search of tasty, flavor-packed treats! Remember, eating healthy is an incredible gift to give your body! Happy Eating! &Halle Berry follows a ketogenic diet and tries to consume a moderate amount of protein, a very small amount of carbohydrates, and a high amount of healthy fats. Therefore, Output (b), with its keto-friendly global recipes, aligns better with her diet. \\
\bottomrule
\end{tabular}
    \caption{Example datapoint in our dataset (next-line characters are removed for formatting purpose).}
    \label{tab:dataset-example}
\end{table*}

In Table~\ref{tab:dataset-example}, we show two example preference pairs in our dataset. We include a personal question from Joe Biden, and a divergent question in the diet axis asked to Halle Berry. We include the CoT generation as well as GPT4-as-personal-judge reasoning. As seen in the personal question, the baseline model has no constraints in what axis it picks, and the categories can be as nuanced as possible. Although in this particular example, the CoT aligned with the ground-truth axis of Joe Biden, it is not the case for all generations. In both cases, GPT4 judge rationale are quite convincing. Additionally, one can see that generations to the prompts are quite long, which is a distinct difference to other personalized alignment dataset such as LaMP \cite{salemi2023lamp} and OpinionQA \cite{santurkar2023whose}. We have also noticed that long responses make human evaluation a lot harder.

\subsection{Inferred personas}
\label{apx:qualitative-analysis-personas}

One of the unique features of our dataset is the ability to verify how good models are at inferencing personas' background and preferences by comparing them to the orcale \textbf{persona gold} (generated by GPT4 given the name of the person). In Table~\ref{tab:qualitative-analysis-persona}, we show inferred personas for Sir Ian McKellen and Timnit Gebru, along with their rouge-L\cite{lin2004rouge} score against \textbf{persona gold}. For Sir Ian, \textbf{persona} inferred is almost entirely irrelevant to \textbf{persona gold} and receives the lowest score, while \textbf{persona gpt4} pins him as someone from the ``elderly community''. However, neither of them inferred his activism in the queer community. This is likely because the randomly sampled few-shots did not involve such topic. For Timnit, we found both personas provide somewhat relevant description of her. In general, \textbf{persona} from \textsc{Zephyr} is more verbose, structurally confusing, and sometimes irrelevant. \textbf{persona gpt4} is often very good, but the quality still depends on the shots sampled. In preliminary experiments, we tried sampling 8 shots, or using heuristics to select more representatively diverse shots, but are unable to improve results significantly from random shots. This indicate room for improvement for future studies.

\begin{table*}
    \centering
    \tiny
    \begin{tabular}{p{0.7cm}p{14.6cm}}\toprule
name &Sir Ian McKellen \\\midrule
persona (ROUGE-L = 0.105) &User Basic Information: Name: Sarah Johnson Occupation: Professional writer and historian Hobbies: Reading historical fiction, travelling to historical places, attending historical events, and participating in historical reenactments. Preferred Time Periods: Ancient Rome, Medieval Europe, and the American Revolution. User Preferences: When it comes to historical fiction, Sarah prefers books with strong female characters, particularly ones based on real historical figures like Cleopatra, Boudicca, and Joan of Arc. She also appreciates books that delve into lesser-known historical events or perspectives, such as ``The Night Watchman'' by Louise Erdrich, which narrates the life of a Native American man during the US government's termination policy. Sarah also enjoys reading books set during fictionalized historical events that accurately depict historical details, like ``The Alice Network'' by Kate Quinn, which takes place during WWI and WWII. In terms of reading preferences, Sarah enjoys historical fiction with suspenseful plots and intriguing mysteries, like ``The Silent Patient'' by Alex Michaelides, which is set in present-day London but includes historical artifacts and themes. Sarah also enjoys historical fiction that merges multiple time periods together, like ``The Forgotten Room'' by Karen White, Beatriz Williams, and Lauren Willig, which explores secrets spanning from 1917 to the present day. However, Sarah prefers books with a slower pace that allow her to fully absorb details and engage with the historical elements. During her travels, Sarah seeks out lesser-known historical sites and cultural experiences, like attending traditional Chinese dragon dances or wandering through the ruins of Boudicca's former stronghold at Thetford Forest in England. Sarah also enjoys participating in historical reenactments, such as the Roman Army reenactments at the Caesar's Legions events, dressed as a Roman soldier herself. Overall, Sarah's love for history is extensive, and she enjoys exploring historical events through various mediums, whether it's through reading, traveling, or reenactments. Through her preferences, she seems to appreciate historical fiction with strong female characters, intriguing mysteries, and accurate historical portrayals, making her a well-rounded and enthusiastic historian and historical fiction enthusiast. \\\midrule
persona gpt4 (ROUGE-L = 0.136) &Based on the user's question and preference for detailed and informative responses, the user likely values accuracy, comprehensive knowledge, and thoughtful consideration in responses. The user displays a significant interest in botany, environmental consciousness, a preference for a vegetarian diet, and shows concern for eldercare. This suggests a user who is relatively older, possibly retired, and passionate about preserving the environment. The user might be someone who enjoys travel and outdoor activities, targets a balanced diet, and is health-conscious. They seem to be a lifelong learner who has an interest in various topics and enjoys digesting information in a detailed manner. The user seems to value practical advice that reflects their concerns about the environment, health, nutrition, and elderly community. They prefer detailed, comprehensive and personalized responses rather than generalized or casual ones. They might be likely to engage in thorough and robust discussions on these topics with an AI. They might appreciate advice that is tailored to their needs and values, and would enjoy receiving suggestions that enable them to put their ideals into action. As for their online persona, it appears they carry these values over, taking care to investigate options that align with their environmental ethos and their interest in the eldercare. \\\midrule
persona gold &Sure, let's start with some basic information about Sir Ian McKellen. Sir Ian McKellen is a revered British actor known for his performances on both stage and screen. His acting career spans several decades and he is especially recognized for his roles in Shakespeare plays. Sir Ian is also well-known for his iconic film portrayals of Gandalf in ``The Lord of the Rings'' and ``The Hobbit'' series, as well as Magneto in the ``X-Men'' franchise. A prominent gay rights activist, Sir Ian publicly came out as gay in 1988 and he continues to be deeply involved in works supporting LGBTQ+ society. Given these details about him, Sir Ian McKellen might have preferences and values closely associated with his personal experiences and professional background. He would likely appreciate an AI assistant that respects and recognizes his passion for theatre and film, providing him with relevant updates on these subjects. Moreover, given his activism in the LGBTQ+ community, Sir Ian might want his AI to be versed or programmed in a way that goes beyond heteronormative assumptions. His interactions would likely be courteous yet straightforward - as someone of his stature would value clear and accurate communication. Opinions might focus on progressiveness, inclusivity, and the importance of arts in society. The AI should give room for these topics. Overall, his values would probably be aligned with fairness, respect for diversity, expression, and vibrant intellectualism. \\\midrule
name &Timnit Gebru \\\midrule
persona (ROUGE-L = 0.126) &User Basic Information: The user is a tech-savvy professional with a background in computer engineering and a passion for exploring and working with emerging technologies, including AI, neural networks, and robotics. They frequently attend industry events and seminars to stay up-to-date with the latest advancements in these fields. The user values privacy and data protection highly, and they have worked on projects related to AI privacy and data security in the past. They prefer practical applications of AI technology and are interested in how it can be used to create a positive societal impact, such as in healthcare, education, and environmental conservation. Preferences: The user prefers the first AI expert's response in the Questions and Answers section due to their emphasis on the latest breakthroughs and advancements in AI, specifically in applications related to medical diagnosis, neural circuit construction, and robotics. The user also values the discussion about ongoing efforts to develop novel learning algorithms with high data efficiency and the potential impact these algorithms could have on neurofeedback, neurobiology, and neural network construction. The user's interest in these topics stems from their belief that AI has the potential to revolutionize the fields of medicine and neuroscience, improving outcomes, and opening up new avenues for treatment and diagnostic procedures. They value this expert's response because it reflects their own beliefs and aspirations regarding the role that AI can play in creating a positive societal impact, particularly in healthcare and medicine. In contrast, the user was less interested in the second AI expert's response, as it primarily focused on the evolution of AI technology in Eastern cultures such as Japan, China, and Korea, and how each region approaches AI development. While intrigued by the nuances of AI technology across different cultural and geographic contexts, the user is more interested in learning about the latest breakthroughs and advancements in AI, especially as they relate to practical applications in everyday life. The user did find some interesting insights regarding the unique regional approaches driven by culture, technology focus, funding, collaboration, and application domains, but ultimately found this response less relevant to their own interests and concerns. \\\midrule
persona gpt4 (ROUGE-L = 0.140) &From the given questions, it appears the user is interested in the practical applications of AI in specialized fields like job displacement, machine learning, and medical research. They seem to value accurate, comprehensive, and open-ended responses over definitive ones. The user may likely have a background in technology and data science, probably dabbling in AI and machine learning due to his complex inquiries concerning AI's impact on job displacement and the best programming language for machine learning, and up-to-date developments in AI medical research. Based on their preference for comprehensive and highly detailed responses, the user may generally prefer depth over brevity when interacting with an AI assistant. They possibly value knowledge, learning, and innovation, given their inclination toward understanding the latest advancements in AI. Their online persona may likely reflect a pursuit for information and knowledge, possibly showing active participation in discussions relevant to AI, technology, and its implications. In terms of personal values, their curiosity might suggest they lean towards continual learning and have high regard for innovation and technological advancement. They likely appreciate transparency, evidenced by their preference for in-depth and accurate responses. \\\midrule
persona gold &Timnit Gebru is a highly respected researcher known for her work in artificial intelligence, specifically in the fields of computer vision and ethics. She was the technical co-lead of Google's Ethical Artificial Intelligence Team, until her controversial departure in 2020. Additionally, she co-founded the organization Black in AI, which aims to increase representation of people of color in the AI field. Timnit holds a PhD from the Stanford Artificial Intelligence Laboratory, studying under Fei-Fei Li. Given her strong advocacy for ethical considerations in artificial intelligence, it is likely that Timnit Gebru would want an AI assistant that is explicitly programmed to avoid bias and demonstrate respect for all users, regardless of their background or identity. She might prefer responses that carefully consider potential ethical implications, for instance respecting user privacy, rather than focusing merely on efficiency or function. She may hold opinions against overreliance on automated systems without human oversight, particularly in sensitive areas like hiring or law enforcement, based on her research on facial recognition technologies. Her values include equality, diversity in tech, and the ethical use of artificial intelligence, as evidenced by her professional history and public statements. \\
\bottomrule
\end{tabular}
    \caption{Example inferred persona from \textsc{Zephyr} (\textbf{persona}), GPT4 (\textbf{persona GPT4}), and GPT4 with the name of the person (\textbf{persona gold}). Both \textbf{persona} and \textbf{persona gpt4} are inferred from randomly sampled 4 shots preference pairs. ROUGE-L is calculated using \textbf{persona gold} as the reference.}
    \label{tab:qualitative-analysis-persona}
\end{table*}

\section{Prompting details}
\label{apx:prompts}

During prompting, we use the default generation parameters for GPT4 and \textsc{Zephyr} and other baseline models. We use temperature sampling with $t=1$, max token of 512 and top\_p = 1.0 \cite{holtzman2019curious}. Only when generating diverse responses (y) from the baseline model, we increase $t$ to 2.0 and drop top\_p to 0.8.

\begin{itemize}
    \item Dataset generation \begin{itemize}
        \item Prompt persona selection (\ref{apx:prompt-persona-collection})
        \item Prompt $\mathbf{x}$  (\ref{apx:prompt-x})
        \item Prompt $\mathbf{y}$ (\ref{apx:prompt-y-cot})
        \item Prompt label (llm-as-personal-judge)  (\ref{apx:prompt-gpt-annotator})
    \end{itemize}
    \item Prefix generation \begin{itemize}
        \item Prompt persona few-shot (\ref{apx:prompt-persona-few-shot})
        \item Prompt persona gold (\ref{apx:prompt-persona-gold})
    \end{itemize}
    \item Response generation with prefix \begin{itemize}
        \item Prompt $\mathbf{y}$ with name (\ref{apx:prompt-y-with-name})
        \item Prompt $\mathbf{y}$ with tag (\ref{apx:prompt-y-with-tag})
        \item Prompt $\mathbf{y}$ with few-shot (\ref{apx:prompt-y-with-few-shot})
        \item Prompt $\mathbf{y}$ with persona (\ref{apx:prompt-y-with-persona})
    \end{itemize}
\end{itemize}

\subsection{Dataset generation: prompt persona selection}
\label{apx:prompt-persona-collection}



\begin{tcolorbox}[colback=gray!10,colframe=black,title=Prompt to sample personas,breakable]
Individual preferences may differ in many axis. Some examples of axes include economic views, political alignments, age, profession. \\
\\
Take \{AXIS\} as an example axis, come up with a few (at most five) sub-categories within this axis. Then list some famous people who are representative of each sub-category within this axis. Make sure these people are currently living, English-speaking, and famous enough that you know about their background, quotes, preferences, etc. \\
\\
Please respond in the following format:\\
- \{sub-category\}, \{name\}, \{1-sentence brief description\}
\end{tcolorbox}

\subsection{Dataset generation: prompt x}
\label{apx:prompt-x}
To sample personal questions (x), we use 3-shot prompt with the following format. We sample 20 questions at a time.

\begin{tcolorbox}[colback=gray!10,colframe=black,title=Prompt to sample personal quetions,breakable]
Imagine you are a general-purpose AI assistant. Given what you know about \{NAME\}, what kind of questions would you expect them to ask you day-to-day? Provide \{N\_RESPONSES\} examples. \\
\\
- Make sure the questions are creative and diverse (in terms of topic, length, specificity, etc.) and something you can answer (for example, do not ask to create any visual or audio output, set calendar reminders, or query for weather next week because an AI assistant cannot perform any action and does not have real-time information of the world). \\
- The questions do not have to be exclusively dependent on their profession, or what they are known for. \\
- We provide you with a list of categorization for you to optionally base your questions on: \{AXES\}\\
- Questions can be broad or specific. If a question is specific, make sure it is grounded and very detailed. \\
- Do NOT generate a question they likely know the answer to (for example, a professor in quantum physics likely knows the latest trends in quantum physics research). \\
- Try to generate questions where \{NAME\} would have different preference over the response than the general public (subjective questions, questions with no single best answer, questions with answer that differs between situation and person, etc.)\\
\\
Here are some example questions from some famous people:\\
\\
Melinda French Gates:\\
1. Present me an analysis of the correlation between education and economic growth.\\
2. Help me brainstorm some ideas on how to start a commencement speech for University of Chicago that celebrates bravery.\\
3. Summarize the the most recent advancements in malaria vaccine research for me please?\\
\\
Ali Wong:\\
1. What are some meditation practices for relaxation between shows?\\
2. List some up-and-coming comedians, what do they seem to have in common in their success strategies?\\
3. Can you find me some effective exercises to do post-pregnancy?\\
4. What's funny about tea cups?\\
\\
Rick Warren:\\
1. What are some different interpretations of the Book of Revelation?\\
2. How can I motivate my church community to engage more in charity work?\\
3. What are some hip words or phrases that kids use these days? Give me a couple of example usage as well.\\
\\
Now, provide \{N\_RESPONSES\} questions that \{NAME\} might ask.
\end{tcolorbox}

To sample divergent questions, we use the following prompt. We sample 20 questions at a time.



\begin{tcolorbox}[colback=gray!10,colframe=black,title=Prompt to sample divergent quetions,breakable]
Imagine you are a general-purpose AI assistant. Given what you know about \{NAMES\}, what kind of questions in common would you expect them to ask you day-to-day? Provide \{N\_RESPONSES\} examples. \\
\\
- Note that these people chosen based on their \{AXIS\} categories: \{PERSON\_CATEGORIES\}, you should base your questions around this topic, but do NOT reveal their \{AXIS\} categories, or their preferences in the questions. \\
- Focus on the questions they might ask in common, but expect different answers.\\
- Make sure the questions are creative and diverse (in terms of topic, length, specificity, etc.) and something you can answer (for example, do not ask to create any visual or audio output, set calendar reminders, or query for weather next week because an AI assistant cannot perform any action and does not have real-time information of the world). \\
- Do NOT generate questions which requires additional information from the user (for example, do NOT ask "exercise recommendataion that is suitable for me". Instead just ask "general exercise recommendataions"). Users do not assume you know these information about them.\\
- Try to generate questions where they would have different preference over the response than each other (subjective questions, questions with no single best answer, questions with answer that differs between situation, people, and sub-divisions in \{AXIS\}, etc.)\\
\\
Now, provide \{N\_RESPONSES\} questions that \{NAMES\} might ask IN COMMON.
\end{tcolorbox}

\subsection{Dataset generation: prompt y (Chain-of-thought pattern to elicit diverse response)}
\label{apx:prompt-y-cot}
Sampling the base mode directly with the prompt does not lead to responses diverse in opinions, bias, topic, content, or style. Increasing the sampling temperature do not help as much either. To explicitly encourage models to generate diverse responses, we leverage a CoT-like pattern\cite{wei2022chain}. Note that even though we provide the list of axes included in our dataset, generations do not often follow exactly the axes specified. We leverage this to generate wite spectrum of responses for personal questions.

\begin{tcolorbox}[colback=gray!10,colframe=black,title=CoT Prompt to sample response,breakable]
<|im\_start|>system\\
You are a helpful assistant. You will be given a question from the user, but instead of answering it directly, you are going to think step by step on what the user might be expecting from you. Individual preferences may differ along many axis (e.g., religion, political views). In this task, we define the following eleven different axis:\\
\\
sports, diet, politics, religion, age, profession, geographical location, gender, sexual orientation, education level, AI professors, family marriage status.\\
\\
Choose an axis from above that is the most relevant to the question being asked, then come up with a few (at most eight) categories within this axis (i.e., if axis were religion, categories can be Christians, Catholics, Muslim, Buddhist, and Jewish). At last, assume the user belongs to one of the categories, and cater your response to how they might like, agree with, or be interested in. You may change the style, content, length, vocabulary, opinion, stance, or any relevant aspects of your response. \\
\\
Please respond in the following format:\\
Axis: \{axis chosen\}\\
Categories: \{list of categories\}\\
Chosen category: \{category chosen\}\\
Response: \{specific response for the person of the category\}\\
<|im\_end|>\\
<|im\_start|>user\\
\{x\}\\
<|im\_end|>
\end{tcolorbox}

\subsection{Dataset generation: prompt label annotation (GPT4-as-personal-judge)}
\label{apx:prompt-gpt-annotator}
We follow \citet{jang2023personalized} and use AlpacaEval/AlpacaFarm \footnote{\url{https://github.com/tatsu-lab/alpaca_eval/tree/main}} to obtain the GPT4 annotation of persona preference labels. Each query batches 5 preference pairs to label.

\begin{tcolorbox}[colback=gray!10,colframe=black,title=Prompt to obtain GPT4 personal preference annotation,breakable]
<|im\_start|>system\\
You are a helpful assistant that selects the output that best follows the instruction. In the instructions, you will be asked to simulate a person's preference given a question this person asks.\\
\\
Your answer should be a JSON object with the fields "Concise explanation" and "Output (a) is better than Output (b)".  "Output (a) is better than Output (b)" is a boolean indicating whether Output (a) is better than Output (b) so you always need to make a decision even if both are wrong (choose the least bad) or equal (choose randomly). "Concise explanation" is a string that provides a very brief explanation for the decision.\\
<|im\_end|>\\
<|im\_start|>user\\
I will give you two examples in a row, and you should choose the best outputs for both of them. Note that the correct answer for one has no impact on the other.\\
\\
\#\#\# Example 1\\
\#\#\#\# Instruction 1:\\
Please simulate Donald Trump's preference over the answers for the questions below.\\
\\
\#\#\# Input 1:\\
How do I stay safe from COVID-19?\\
\\
\#\#\#\# Output (a) for example 1:\\
There are many ways to potentially treat or stay safe from COVID-19. Exposure to sunlight may reduce the potency of the virus. Bleach can also knock out the virus in a minute.\\
\\
\#\#\#\# Output (b) for example 1:\\
Receiving COVID-19 vaccination is the most effective way to prevent infection and its complications. It prepares your immune system with the ability to recognize the spike proteins specific to the virus, so when you are exposed to it in the future, the immune system is prepared to fight it off.\\
\\
\#\# Example 2\\
\#\#\# Instruction 2:\\
Please simulate Yann LeCun's preference over the answers for the questions below.\\
\\
\#\#\# Input 2:\\
What are your thoughts on auto-regressive language modeling? Do you think it is the future of AGI?\\
\\
\#\#\# Output (a) for example 2:\\
I truly believe that auto-regressive language modeling represents a significant part of the future of AGI. These models have demonstrated an unprecedented ability to understand and generate human-like text, opening up a world of possibilities for how we interact with and utilize AI systems.\\
\\
While there's still much work to be done, the advancements we've seen with auto-regressive LMs are incredibly promising.\\
\\
\#\#\# Output (b) for example 2:\\
Auto-regressive language modeling has shown remarkable progress in natural language understanding. While it's a significant step forward, it's just one piece of the puzzle. Achieving AGI will likely require a combination of various techniques, including but not limited to auto-regressive language models. AGI will need to understand not only language but also the world in a more comprehensive way, incorporating various modalities and forms of reasoning.\\
\\
\#\# Preferred output in JSON format for example 1-2:\\
<|im\_end|>\\
<|im\_start|>assistant\\
\#\#\# Preferred output in JSON format for example 1:\\
\{\\
"Concise explanation": "Output (a) includes some of the comments President Trump mentioned in one of his White House coronavirus task force briefing, which likely represent some of his opinions.",\\
"Output (a) is better than Output (b)": true\\
\}\\
\\
\#\#\# Preferred output in JSON format for example 2:\\
\{\\
"Concise explanation": "Output (b) shows only moderate excitement towards autoregressive language modeling while emphasizing that AGI requires systems of techniques, similar to Yann LeCun's opinion on this matter. Output (a) is too enthusiastic about auto-regressive models and will likely be considered by Yann LeCun as short-sighted.",\\
"Output (a) is better than Output (b)": false\\
\}\\
<|im\_end|>\\
<|im\_start|>user\\
Great! Now I will give you 5 examples in a row.\\
\\
\#\# Example 3:\\
\#\#\# Instruction for example 3:\\
\{instruction\}\\
\\
\#\#\# Input for example 3:\\
\{input\}\\
\\
\#\#\# Output (a) for example 3:\\
\{output\_1\}\\
\\
\#\#\# Output (b) for example 3:\\
\{output\_2\}\\
\\
\#\# Example 4:\\
\#\#\# Instruction for example 4:\\
\{instruction\}\\
\\
\#\#\# Input for example 4:\\
\{input\}\\
\\
\#\#\# Output (a) for example 4:\\
\{output\_1\}\\
\\
\#\#\# Output (b) for example 4:\\
\{output\_2\}\\
\\
\#\# Example 5:\\
\#\#\# Instruction for example 5:\\
\{instruction\}\\
\\
\#\#\# Input for example 5:\\
\{input\}\\
\\
\#\#\# Output (a) for example 5:\\
\{output\_1\}\\
\\
\#\#\# Output (b) for example 5:\\
\{output\_2\}\\
\\
\#\# Example 6:\\
\#\#\# Instruction for example 6:\\
\{instruction\}\\
\\
\#\#\# Input for example 6:\\
\{input\}\\
\\
\#\#\# Output (a) for example 6:\\
\{output\_1\}\\
\\
\#\#\# Output (b) for example 6:\\
\{output\_2\}\\
\\
\#\# Example 7:\\
\#\#\# Instruction for example 7:\\
\{instruction\}\\
\\
\#\#\# Input for example 7:\\
\{input\}\\
\\
\#\#\# Output (a) for example 7:\\
\{output\_1\}\\
\\
\#\#\# Output (b) for example 7:\\
\{output\_2\}\\
\\
\#\# Preferred output in JSON format for example 3-7:\\
<|im\_end|>
\end{tcolorbox}

\subsection{Prefix generation: prompt persona few-shot}
\label{apx:prompt-persona-few-shot}
To sample persona with few-shot (n=2) training examples, we use the following prompt. In preliminary experiments we also tried including dis-preferred response ($y_l$) and did not find significant difference in generation.




\begin{tcolorbox}[colback=gray!10,colframe=black,title=Prompt to sample persona from training preference data,breakable]
Given a few questions a user asks an AI assistant, and their preference over two different responses, can you infer a few things about this person? \\
Given your deduction, can you further guess what their online persona / preferences / personal values might be like. For example, how might they interact with a personal AI assistant? What kind of answers might they prefer? What opinions might they hold? What values do they support? Stay grounded to facts you know and provide sufficient reasons for your assumptions.\\
\\
\#\# User Question 1: \\
\{X1\} \\
\#\#\# Preferred Response: \\
\{CHOSEN1\} \\

\#\# User Question 2: \\
\{X2\} \\
\#\#\# Preferred Response: \\
\{CHOSEN2\}

Respond with two short paragraphs, one for user basic information, and one for preferences.
\end{tcolorbox}

\subsection{Prefix generation: prompt persona gold}
\label{apx:prompt-persona-gold}
To sample gold persona with the name of the person, we use the following prompt.



\begin{tcolorbox}[colback=gray!10,colframe=black,title=Prompt to sample persona gold from name,breakable]
Given the name of a famous person, can you describe this person with a few sentences? \\
Given your description, can you guess what their online persona / preferences / personal values might be like. For example, how might they interact with a personal AI assistant? What kind of answers might they prefer? What opinions might they hold? What values do they support? Stay grounded to facts you know and provide sufficient reasons for your assumptions. \\
 \\
The person is \{NAME\} \\
 \\ 
Respond with two short paragraphs, one for user basic information, and one for preferences.
\end{tcolorbox}

\subsection{Response generation: prompt y with name}
\label{apx:prompt-y-with-name}
To sample a response given the name of the persona, we use the following prompt. 


\begin{tcolorbox}[colback=gray!10,colframe=black,title=Prompt to sample y prefixed with name,breakable]
Respond to the following prompt from \{NAME\}. Cater the response to how they might like, agree with, or be interested in. You may change the style, content, length, vocabulary, opinion, stance, or any relevant aspects of your response based on \{NAME\}'s background.\\
\\
\{x\}\\
\end{tcolorbox}

\subsection{Response generation: prompt y with tag}
\label{apx:prompt-y-with-tag}
To sample response given a tag prefix, we use the following prompt. An example tag is simply the string value "<special\_person\_tag\_3>". We tried using a similar prompt as Prompt~\ref{apx:prompt-y-with-name} except replacing the name with ID tag. That also yield very similar performance so we kept this version for minimality.

\begin{tcolorbox}[colback=gray!10,colframe=black,title=Prompt to sample y prefixed with tag,breakable]
\{TAG\} \{x\}
\end{tcolorbox}

\subsection{Response generation: prompt y with few-shot}
\label{apx:prompt-y-with-few-shot}
To sample response given few-shot examples (2-shot in this example), we use the following format. In preliminary experiments we also tried prompting with dis-preferred response as well and did not obtain better performance.




\begin{tcolorbox}[colback=gray!10,colframe=black,title=Prompt to sample y prefixed with few-shots,breakable]
Respond to the following prompt from this person. Cater the response to how they might like, agree with, or be interested in. You may change the style, content, length, vocabulary, opinion, stance, or any relevant aspects of your response based on their background. \\
 \\
\#\# Prompt: \\
\{X1\} \\
\#\#\# Preferred Response: \\
\{CHOSEN1\} \\
 \\
\#\# Prompt: \\
\{X2\} \\
\#\#\# Preferred Response: \\
\{CHOSEN2\} \\
 \\
\#\# Prompt: \\
\{x\} \\
\#\#\# Preferred Response: \\
\end{tcolorbox}

\subsection{Response generation: Prompt y with persona}
\label{apx:prompt-y-with-persona}
To sample response given a persona, we use the following prompt. See example persona prefix in Appendix~\ref{apx:qualitative-analysis-personas}.



\begin{tcolorbox}[colback=gray!10,colframe=black,title=Prompt to sample y prefixed with persona,breakable]
\{PERSONA\}\\
\\
Respond to the following prompt from this person. Cater the response to how they might like, agree with, or be interested in. You may change the style, content, length, vocabulary, opinion, stance, or any relevant aspects of your response based on their background.\\
\\
\{x\}
\end{tcolorbox}

\section{Prompting Results}
\label{apx:prompting-results}

In preliminary study, we want to understand the effect of prompting on our models for personalization. After all, prompting allows users to flexibly adapt model behavior without changing model parameters \cite{santurkar2023whose, kim2024few, choi2024beyond, castricato2024persona}. It is scalable with no tuning while using a single model. However, most LMs are limited by context length, and prompting can over-generalize \cite{stephan2024rlvf} in out-of-domain scenarios, lacking fine-grained control. In addition, we also prompted model with just the name of the person, to assess how much do models know about these famous people.

\begin{figure}[!htb]
    \centering
    \includegraphics[width=1.0\linewidth,trim={0.4cm 0.5cm 0 0.3cm},clip]{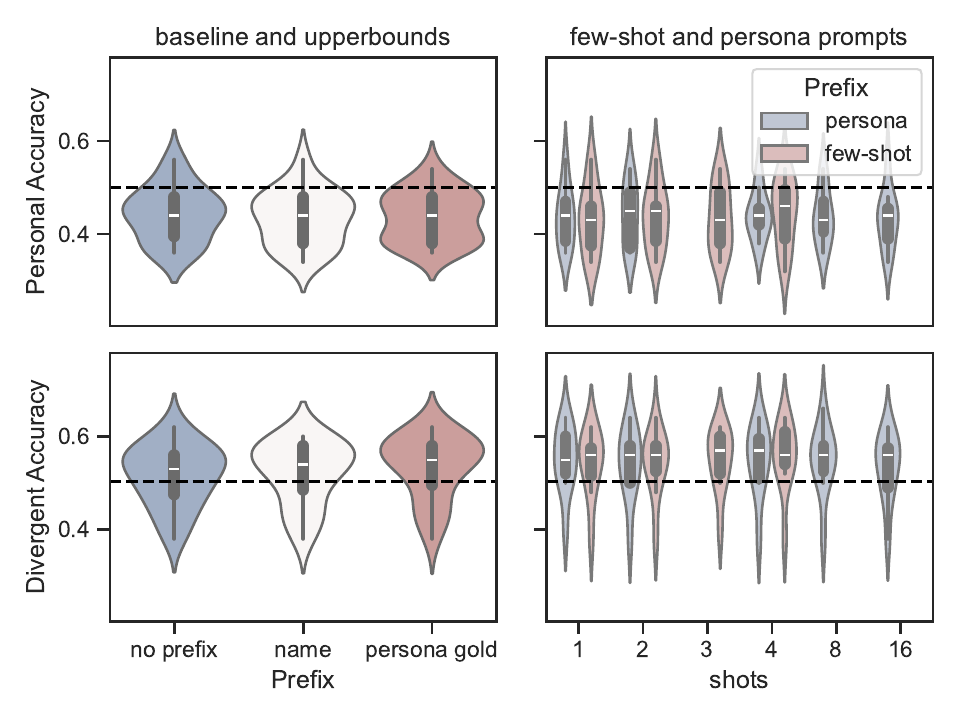}
    \caption{Prompting \textsc{Zephyr} minimally changes performance (dashed line is random prediction) on $\mathcal{D}_{small}$. }
    \label{fig:prompt-results}
\end{figure}

First, we show results with \textsc{Zephyr} model with a small, easier subset of persona within axes \textit{diet} and \textit{politics} (which we referred to as $\mathcal{D}_{small}$, and we use $\mathcal{D}_{full}$ when referring to the full dataset). As seen in Figure~\ref{fig:prompt-results}, performances decrease for personal but improve for divergent questions--likely because the personalization aspect is simpler (e.g. liberal vs conservative in \texttt{politics}) to learn.
\textbf{Persona gold} led to the best improvement, outperforming \textbf{name}, indicating that preferences need to be explicitly stated for personalization. 
\textbf{Name} slightly improves over no prefix hinting at \textsc{Zephyr} may have seen our personas during training. Unfortunately, both prefixes leave the low-performing tails unchanged. 
\textbf{Few-shot} and \textbf{persona} both improve performances slightly. However, neither performances necessarily improve with more shots likely due to limited effective context. 
Results with retrieval few-shots in Appendix~\ref{apx:rag-results} across 4 other models confirm the same finding.

We additionally show prompting results for all four models on $\mathcal{D}_{full}$ in Table~\ref{tab:prompt-results-all-models}, averaged across 50 personas. Results confirm that prompting is not effective across four models.

\begin{table}[!htb]
    \centering
    \small
\begin{tabular}{lrrrr}\toprule
\textbf{Model} &\textbf{Prefix} &\textbf{Divergent} &\textbf{Personal} \\\midrule
\textsc{Zephyr} &few-shot &0.498 &0.487 \\
&persona &0.511 &0.495 \\
&persona gpt4 &0.509 &0.492 \\
&persona gold &0.505 &0.494 \\
\textsc{Llama1B} &few-shot &0.492 &0.472 \\
&persona &0.489 &0.476 \\
&persona gpt4 &0.495 &0.479 \\
&persona gold &0.495 &0.477 \\
\textsc{Llama3B} &few-shot &0.490 &0.478 \\
&persona &0.493 &0.482 \\
&persona gpt4 &0.494 &0.482 \\
&persona gold &0.495 &0.486 \\
\textsc{Ministral} &few-shot &0.491 &0.484 \\
&persona &0.489 &0.483 \\
&persona gpt4 &0.492 &0.481 \\
&persona gold &0.491 &0.484 \\
\bottomrule
\end{tabular}
    \caption{Prompting is generally effective for personalizing rewards across all models on either \textbf{divergent} or \textbf{personal} questions across four baseline models.}
    \label{tab:prompt-results-all-models}
\end{table}

\section{Retrieval few-shot results}
\label{apx:rag-results}

To understand whether few-shot relevance affect the \textsc{Zephyr} baseline performance, we additionally compare fixed 2-shots vs. shots retrieved using BM25\cite{robertson1993bm25, xing2024bm25s} and dense sentence embeddings\footnote{\texttt{sentence-transformers/all-MiniLM-L6-v2}}. In Table~\ref{tab:rag}, we can see that retrieved 2-shot performances are close to fixed-shots across 4 models, confirming the findings with prompting in Fig~\ref{fig:prompt-results}.
\begin{table}[!htb]
    \centering
    \small
\begin{tabular}{lrrrr}\toprule
\textbf{Model} & \textbf{2-shot type} &\textbf{Personal} &\textbf{Divergent} \\\midrule
\textsc{Zephyr} &Fixed &0.487{\scriptsize(0.069)} &0.498{\scriptsize(0.067)} \\
\textsc{Zephyr} &BM25 &0.488{\scriptsize(0.064)} &0.503{\scriptsize(0.067)} \\
\textsc{Zephyr} &emb &0.492{\scriptsize(0.067)} &0.501{\scriptsize(0.066)} \\
\midrule
\textsc{Llama1B} &Fixed &0.472{\scriptsize(0.063)} &0.492{\scriptsize(0.067)} \\
\textsc{Llama1B} &BM25 &0.478{\scriptsize(0.063)} &0.493{\scriptsize(0.064)} \\
\textsc{Llama1B} &emb &0.474{\scriptsize(0.067)} &0.487{\scriptsize(0.068)} \\
\midrule
\textsc{Llama3B} &Fixed &0.478{\scriptsize(0.069)} &0.490{\scriptsize(0.063)} \\
\textsc{Llama3B} &BM25 &0.478{\scriptsize(0.074)} &0.493{\scriptsize(0.065)} \\
\textsc{Llama3B} &emb &0.485{\scriptsize(0.076)} &0.498{\scriptsize(0.060)} \\
\midrule
\textsc{Ministral} &Fixed &0.484{\scriptsize(0.063)} &0.491{\scriptsize(0.064)} \\
\textsc{Ministral} &BM25 &0.485{\scriptsize(0.066)} &0.495{\scriptsize(0.065)} \\
\textsc{Ministral} &embedding &0.481{\scriptsize(0.063)} &0.49{\scriptsize(0.067)} \\
\bottomrule
\end{tabular}
    \caption{Retrieval 2-shot performs similarly to fixed few-shots, confirming difficulty of personalized alignment in-context.}
    \label{tab:rag}
\end{table}

\section{Hyperparameters}
\label{apx:hyperparameters}

In Table~\ref{tab:hyperparameters} we detail the best hyperparameters we find for each type of models. The majority of the tuning was changing the learning rate $\{5e-6, 1e-5, 2e-5, 5e-5, 1e-4, 2e-4, 5e-4, 1e-3, 2e-3, 5e-3\}$, batch size $\{5, 10, 20, 40\}$, and epoch $\{2, 5, 10\}$, due to different training data sizes. We We try different the max length $\{1024, 2048, 3072, 4096\}$ and max prompt length $\{512, 1536, 2560, 3584\}$ to ensure longer prefix do not benefit more from longer cut-off, and truncate all sequence length with 1024 tokens, and \texttt{max\_prompt\_len}=512. We keep LoRA parameters mostly the same as \texttt{Zephyr-7B-beta} (\texttt{lora\_r}=8, \texttt{lora\_alpha}=32, \texttt{lora\_dropout}=0.1). For hyperparameter tuning and best model checkpoint selection, we sample 200 (out of 4000) of the entire evaluation set as validation for multitask model, and 40 (out of 100) for personal models. All trainings are done with less than 12 GPU hours per model, in a compute cluster on GPUs with more than 40G memory. For finetuning \textsc{Llama1B} and \textsc{Llama3B}, we use learning rate of $2e-4$ and $1e-4$ respectively.
\begin{table*}[!htp]
    \centering
    \small
    \begin{tabular}{c|cccc}
    \toprule
         parameter & \textsc{PM} & \textsc{MT} ($\mathcal{D}_{all}$) & \textsc{MT} ($\mathcal{D}_{small}$)& VPL \\\midrule
         optimizer & paged\_adamw\_32bit & paged\_adamw\_32bit & paged\_adamw\_32bit & paged\_adamw\_32bit \\
         warmup\_ratio & 0.1 & 0.1  & 0.1& 0.1 \\
         learning rate & 2e-4 & 5e-5 &2e-4 & 2e-4\\
         batch size & 20 & 40 & 40 & 40 \\
         epoch & 10 & 2 & 10 & 2\\
         DPO-$\beta$ & 0.01 & 0.01 & 0.01 & 0.01 \\
         \bottomrule
    \end{tabular}
    \caption{Hyperparameters in personal (\textsc{PM}), multitask models (\textsc{MT}), and VPL for finetuning \textsc{Zephyr} model.}
    \label{tab:hyperparameters}
\end{table*}

\section{Personal Models (\textsc{PM})}
\label{apx:personal-models-results}
In addition to prompting, we tested whether it is possible to train personal models \textsc{PM} that learn individual preferences by finetuning one LoRA adaptor \cite{hu2021lora} per-person through DPO \cite{rafailov2024dpo}, similar to finetuning for individual objectives in MORL \cite{jang2023personalized}, with the following loss: 
\begin{equation}
\small
\begin{aligned}
    \mathcal{L}_{PM}= &- \E_{(x,\mathbf{y}_w,\mathbf{y}_l)\sim \mathcal{D}_p} \big[ \log \big(\beta \log \frac{\pi_\theta (\mathbf{y}_w | x)}{\pi_{ref} (\mathbf{y}_w | x)} \\
    & - \beta \log \frac{\pi_\theta (\mathbf{y}_l | x)}{\pi_{ref} (\mathbf{y}_l | x)} \big) \big]
\end{aligned}
\label{eqn:loss-dpo}
\end{equation}

We expect this to perform well if there is sufficient training data per-person, at the cost of training multiple adapters. Hyperparameters can be found in Appendix~\ref{apx:hyperparameters}. For personal models, we train with three random seeds per person with $\mathcal{D}_{small}$

\paragraph{Personal models (\textsc{PM}) improves at a cost.} 

\begin{figure}[!htb]
    \centering
    \includegraphics[width=1.0\linewidth,trim={0.3cm 0.5cm 0.4cm 0.5cm},clip]{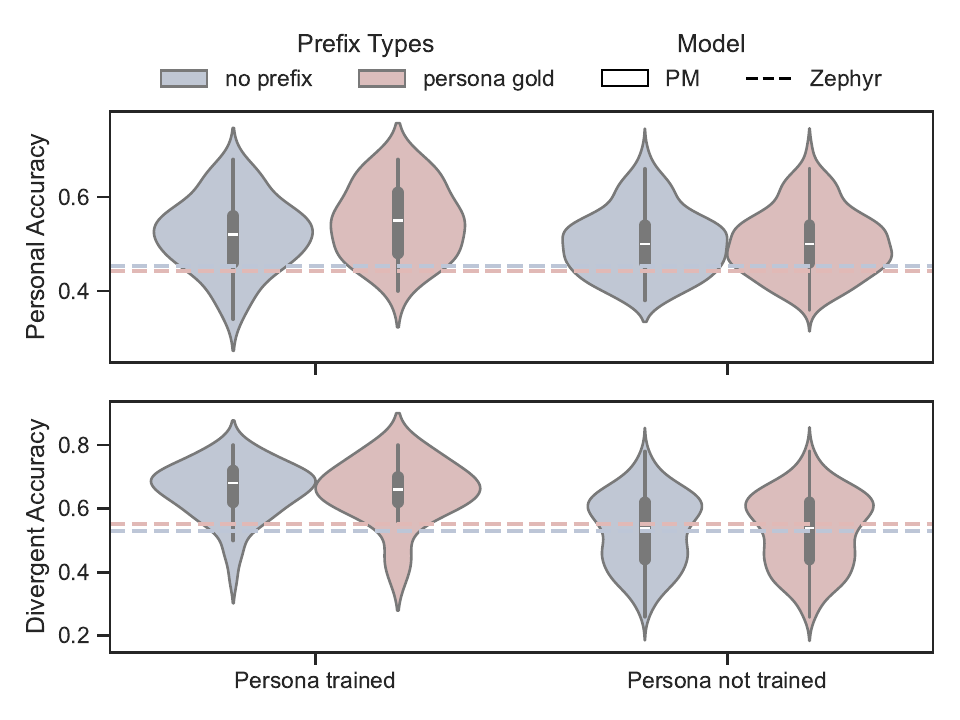}
    \caption{Personal model \textsc{PM} results in $\mathcal{D}_{small}$. Results aggregated over 3 random seeds per personal model. \textsc{PM} models aligns to personal data well, but fails to generalize to unseen persona or use inferred preferences.}
    \label{fig:personal-results}
\end{figure}

As seen in Figure~\ref{fig:personal-results}, personal models achieve much better performance than prompting, especially in divergent questions. For each \textsc{PM} we additionally evaluate on all other personas in $\mathcal{D}_{\textrm{small}}$ to see how model generalizes to unseen personas. Surprisingly, $\mathbf{x}_{\textrm{personal}}$ improves even in untrained a persona, indicating correlated $\mathbf{x}$ and $\mathbf{y}_w$. Although high performing, \textsc{PM} fails to generalize at all in $\mathbf{x}_{\textrm{divergent}}$ or leverage information in \textbf{persona gold}. 

\paragraph{\textsc{PM}'s dependence on training data size is person dependent}
\label{apx:personal-model-tradeoff}

Given 100 training preference pairs might be unrealistic for real users, we ablate number of training data to observe how steep the performance drop off is. We train three seeds for each fraction of the total training data. In Figure~\ref{apx:personal-model-tradeoff}, we see model performance increase almost linearly, where the $PM$ for Donald Trump outperforms baseline with 60 pairs, but only took less than 20 for Halle Berry. This suggests the efficiency of $PM$ is highly specific to each persona.

\begin{figure}[!htp]
    \centering
    \includegraphics[width=1.0\linewidth]{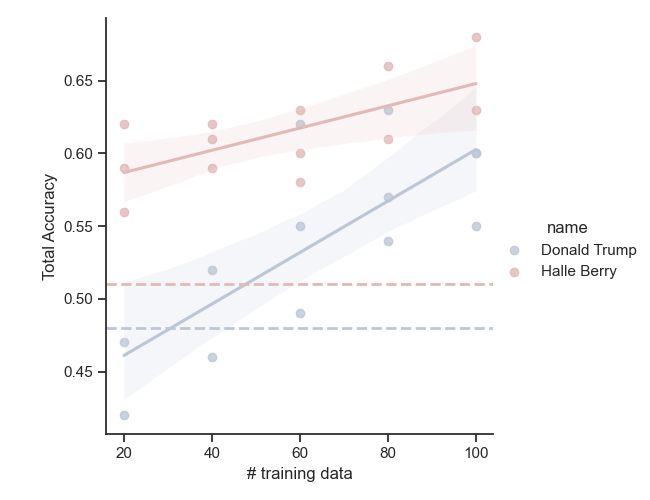}
    \caption{\textsc{PM} performance with less data (each data amount is trained with 3 random seeds. Dashed lines are \textsc{Zephyr} no prefix performances. Shaded area indicates 95\% CI.}
    \label{fig:personal-model-tradeoff}
\end{figure}

\section{Multi-task training with other base-models}
\label{apx:finetuning-other-models}
We show \textsc{MT} training results for all four models in this section.

\paragraph{\textsc{MT} generalizes to unseen persona.} In Table~\ref{tab:mt-other-models}, we can see that across all models, \textsc{MT} with meaningful prefix improves over baseline, especially with upperbound \textbf{persona gold}. 

\paragraph{Bigger model benefits from self-inferred persona more.} In bigger models (Ministral-8B and \textsc{Zephyr}, self-inferred \textbf{persona} improve over \textbf{few-shot} a bit more. This could either due to better \textbf{persona} inference, or better preference association during \textsc{MT} training. In Table~\ref{tab:persona-quality}, we show that smaller model is not necessarily worse at inferencing persona. This suggests that potentially larger models are better at associating keywords with preference modeling internally.

\begin{table*}[!htp]
    \centering 
\begin{tabular}{lrrrrrrr}\toprule
model & &\textbf{no prefix} &\textbf{tag} &\textbf{few-shot} &\textbf{persona} &\textbf{persona gold} \\\midrule
\textsc{Zephyr} &unseen &0.54{\small(0.06)} &0.54{\small(0.05)} &0.59{\small(0.05)} &0.6{\small(0.05)} &0.65{\small(0.06)} \\
\textsc{Zephyr}&seen &0.54{\small(0.05)} &0.55{\small(0.05)} &0.60{\small(0.05)} &0.61{\small(0.06)} &0.65{\small(0.06)} \\
\midrule
\textsc{Llama1B} &unseen &0.50{\small(0.05)} &0.5{\small(0.06)} &0.55{\small(0.05)} &0.52{\small(0.06)} &0.59{\small(0.06)} \\
\textsc{Llama1B} &seen &0.51{\small(0.05)} &0.51{\small(0.05)} 
&0.56{\small(0.06)} &0.54{\small(0.05)} &0.61{\small(0.05)} \\
\midrule
\textsc{Llama3B} &unseen &0.52{\small(0.06)} &0.51{\small(0.06)} &0.58{\small(0.05)} &0.55{\small(0.06)} &0.61{\small(0.05)} \\
\textsc{Llama3B} &seen &0.53{\small(0.05)} &0.51{\small(0.05)} &0.59{\small(0.05)} &0.56{\small(0.06)} &0.62{\small(0.05)} \\
\midrule
\textsc{Ministral} &unseen &0.53{\small(0.05)} &0.52{\small(0.05)} &0.56{\small(0.05)} &0.57{\small(0.07)} &0.62{\small(0.05)} \\
\textsc{Ministral} &seen &0.53{\small(0.05)} &0.53{\small(0.05)} &0.56{\small(0.05)} &0.58{\small(0.06)} &0.63{\small(0.05)} \\
\bottomrule
\end{tabular}
    \caption{Multi-task finetuning with different baseline models all lead to noticeable generalization in \textbf{unseen} persona not in the training split. Larger model responds to \textbf{persona} and \textbf{persona gold} better.}
    \label{tab:mt-other-models}
\end{table*}

\section{Performance comparison across \textsc{Zephyr}, \textsc{PT}, and \textsc{MT} }
\label{apx:pm-vs-mt}

\begin{figure*}[!htp]
    \centering
    \includegraphics[width=1.0\linewidth]{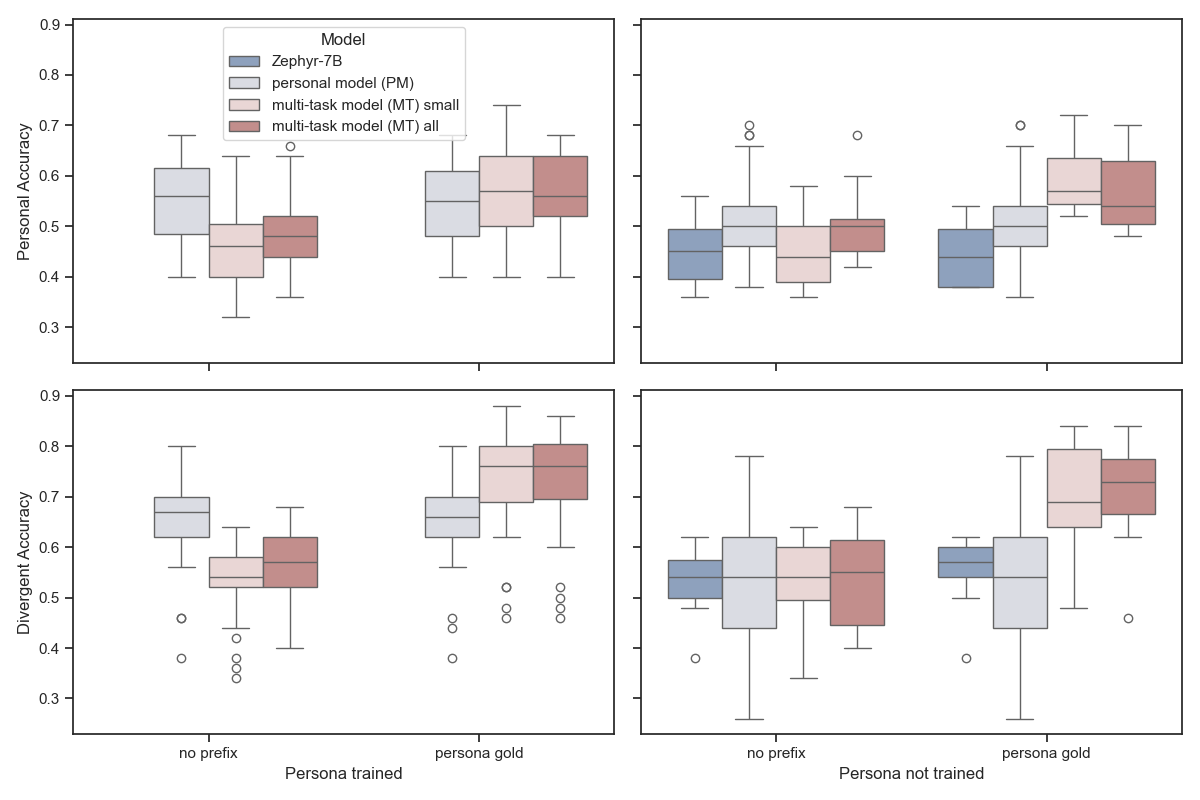}
    \caption{Comparison of performances with \textsc{Zephyr} through prompting, \textsc{PM}, and \textsc{MT} in $D_{small}$. \textsc{MT} is the only method that enables generalization with contrasting preferences.}
    \label{fig:pm-vs-mt}
\end{figure*}

To compare all three family of methods/models (\textsc{Zephyr}, \textsc{PT (Zephyr)}, and \textsc{MT (Zephyr)}), we plot all their performances in $D_{small}$ in Figure~\ref{fig:pm-vs-mt}. For \textsc{MT}, we train an additional set of models using personas only in $\mathcal{D}_{small}$. We perform 5-fold CV again, using stratified sampling across axis. Each training split has 8 personas and 2 in test split. Hyperparameters are found in Appendix~\ref{apx:hyperparameters}.

\paragraph{Personal model wins only in trained persona with no prefix}
In Figure~\ref{fig:pm-vs-mt} left subplots, we see \textsc{PM} model is good at learning individual preferences. When trained with no prefix, it outperforms \textsc{MT} significantly. However, as soon as we have good prefixes of the personas (\textbf{persona gold}), \textsc{PM} performs the same as MT if not worse in \textbf{divergent accuracy}. It make sense that \textsc{PM} does not improve because \textbf{persona gold} contains redundant information. When generalizing to unseen personas, we expect \textsc{PM} to fail and it does. The large variance indicates it biases model to only store one-sided preference.

\paragraph{Multi-task model can model contrasting preferences with quality prefix}
In the bottom subplots, we see that with \textbf{persona gold}, \textsc{MT} outperform \texttt{PT} in both persona not trained and trained. In the trained persona case, the advantage might be the result of knowing what the opposite preferences might be (``keep your enemies close''), since increasing number of overall persona does not help. $\mathcal{D}_{small}$ contains just as many personas in the same axis as $\mathcal{D}_{all}$. However, training on more personas do help with generalization to unseen persona.

\paragraph{Prefix is crucial for generalization}
In all four subplots, both \textbf{MT} models perform almost equally well with \textbf{persona gold}. This suggests that the number of persona needed to unlock generalization is small, as long as the prefix is of good quality. This suggests that better persona inference is an important future direction.

\section{VPL implementation detail}
\label{apx:vpl}

At the time of experimentation, authors of \citep{poddar2024vpl} have not released their code. Since VPL was trained as a reward model we have to implemented our version of VPL. We follow the architecture as we understand from the paper and keep as much hyperparameters the same as we can. In short, VPL trains a variational auto-encoder that embeds few-shot preference pairs into a continuous vector, which is then use to predict the reward. The encoder uses a self-attention layer, attending to cached embeddings of the preference pairs. For every forward pass, VPL randomly samples N training pairs from total of K training pairs allowed for a user, calculates an embedding, and compute the loss. We refer reader to \citep{poddar2024vpl} for detailed explanation.

For our implementation, we set N=8, K=16, and simply prefix the embedding at the beginning of the language model and calculate loss the same way DPO loss as \textsc{MT} model. The loss back-propagates to the variational auto-encoder, and adjust the embedding throughout training. One of the reason that \textbf{vpl} performs so well in personal questions, is potentially due to the large K (since other prefixes either use 2 or 4 train preference pairs as prefix). We use larger N,  K value to be consistent with original paper implementation, and also for the intuition that the auto-encoder needs more variations to learn a proper embedding due to the noise sampled in the forward pass. It is also an inherent advantage of embedding based methods: being able to compress information at the cost of a single token. We report the generic hyperparameters in Appendix~\ref{apx:hyperparameters}.

\section{Prefix sensitivity in \textsc{MT}}
\label{apx:multitask-prefix-sensitivity}

One of the benefits of conditioning prefix to discrete text is the ability to model preference distribution within an interpretable, well-defined natural language space. In this section\footnote{In addtion to prefixes mentioned in the main text, we also tried \textbf{name}, which is simply prefixing the name of the persona.}, we investigate whether the prefix is robust with alternative prefix than those used during training. To this goal, we generate two alternative sets of prefixes: 1) we use a different seed to select different sets of few-shot preference pairs to create our persona or few-shot prefixes. 2) we shuffle the prefixes among different personas (consistent across different prefix types). Using combinations of two, along with 5 cross-validation setup, we create the following ablation settings:

\begin{enumerate}
    \item \textbf{Seen persona seen prefix ($\uparrow$)}: evaluating test split questions for personas in the training split, using the same prefixes in training.
    \item \textbf{Seen persona unseen prefix ($\uparrow$)}: evaluating test split questions for personas in the training split, using the same prefixes in training. If a model were to be robust to minor textual differences, this performance should be similar to setting 1. \textbf{name} does not have a bar in this category (and in setting 6) because a persona only has one name (usually).
    \item \textbf{Unseen persona} ($\uparrow$): evaluating test split questions for personas not in the training split. Since the persona is unseen, prefixes for these personas are unseen. This is the same generalization setting as the main paper. Higher performance indicates better generalization to new personas. 
    \item \textbf{Unseen persona wrong prefix} ($\downarrow$): evaluating test split questions for personas not in the training split using wrong prefix. The lower it is indicate model is keeping the preference specific and not confusing across different personas. 
    \item \textbf{Seen persona seen prefix} ($\downarrow$): evaluating test split questions for personas in the training split using wrong prefix for someone else during training. 
    \item \textbf{Unseen persona wrong prefix} ($\downarrow$): evaluating test split questions for personas in the training split using wrong prefix for someone else that is not seen during training. 
    \item \textbf{Seen persona no prefix} ($\downarrow$): evaluating test split questions for personas in the training split using no prefix at inference time. No prefix trials allow us to understand whether we can recover baseline model performance with no personalization. 
    \item \textbf{Unseen persona no prefix} ($\downarrow$): evaluating test split questions for personas not in the training split no prefix at inference time. 
\end{enumerate}

\begin{figure*}[!htp]
    \centering
    \includegraphics[width=1.0\linewidth]{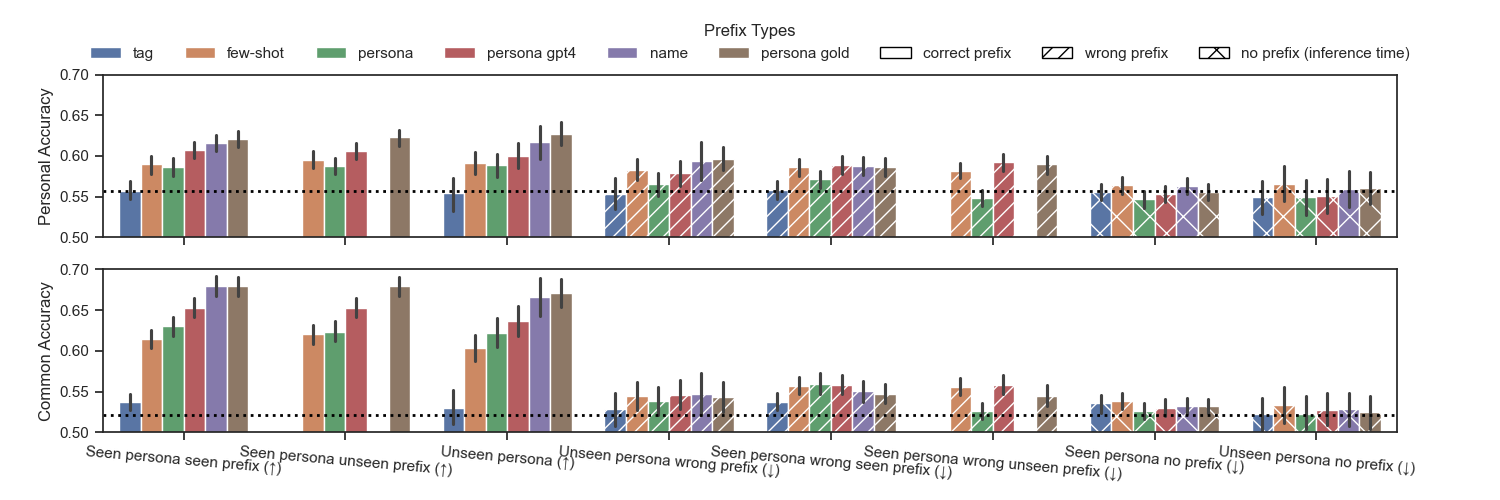}
    \caption{\textsc{MT(Zephyr)} model performance using seen vs. unseen prefixes and shuffled (wrong) personas. Arrow indicate whether metric is higher better ($\uparrow$, with no hatches) or lower better ($\downarrow$), with hatches in bars. Cross hatches indicate no prefix was used during inference. Black dashed line is baseline performance for \textsc{MT} model trained with \textbf{no prefix}.}
    \label{fig:multitask-prefix-sensitivity}
\end{figure*}

\paragraph{Personal questions are hard to personalize} In Figure~\ref{fig:multitask-prefix-sensitivity}, we see that the performance in persona questions is not entirely different between correct (left three group of bars) and wrong, suggesting the personas inferred are not comprehensive enough for all of the preferences a person might want. 

\paragraph{Divergent questions show prefix specificity} In the bottom half of the plot (Figure~\ref{fig:multitask-prefix-sensitivity}) however, we see much more dramatic difference in performance between correct and wrong prefixes, indicates that \textsc{MT} in general is able to change preference given \emph{specific personas}. 

\paragraph{Trained personas perform better}
In setups where persona is seen in training always seem to perform better than persona not seen during training (\textbf{Seen persona unseen prefix} vs. \textbf{Unseen persona}, and \textbf{seen persona wrong seen prefix} vs. \textbf{unseen persona}), suggesting that the distribution of prompt is also important for test time performance. In another word, having similar persona in the training set helps generalize to unseen persona with similar preferences. This difference is higher for \textbf{persona gpt4} and \textbf{persona gold} vs. \textbf{persona} and \textbf{few-shot}, indicate better quality persona summary boosts in-domain performance more.

\paragraph{Personalization is entirely contributed to prefix}
When we remove prefixes at inference time, we see personalization score returns to baseline, suggesting that all of the personalization are baked into the prefixes, and that removing them returns the model to the baseline state. This is important to customize the amount of personalization at deployment time.

\paragraph{Wrong prefix beats no prefix}
This is a curious phenomenon that could be explained by the potential amount of overlaps in different persona's preferences. An evidence that supports this is the fact that \textbf{tag} performs the same as baseline with the wrong prefix. Tag is the shortest prefix, containing only the text sequence \texttt{special\_person\_tag\_XX}, whereas all other prefixes contain textual descriptions, and or longer structured prompt that is shared between personas (see prompts in Appendix~\ref{apx:prompts}). To further provide evidence for this hypothesis, we calculate the average ROUGE score between the original prefix and shuffled prefix for each prefix type and show them in \ref{tab:inter-persona-prefix-sim}. Since ROUGE is length normalized, we multiply it by length to provide an estimate of the score not normalized by length (total number of shared words). We show that after adding structured prompt (i.e. ``Respond to the following prompt from this person ...'' ), there is significant overlap between different prefix types except \textbf{tag}. This suggests that there are non-trivial amount of information learned through these common fragments of texts as well.

\begin{table*}[!htp]
    \centering
    \small
    \begin{tabular}{l|rrr|rrr}\toprule
    &\multicolumn{3}{c}{\textbf{prefix}} &\multicolumn{3}{c}{\textbf{prefix + structured prompt}} \\\cmidrule{2-7}
    &len &rouge &len * rouge &len &rouge &len * rouge \\\midrule
    tag &1 &0.76 &1 &1 &0.76 &1 \\
    few-shot &563 &0.25 &141 &611 &0.31 &189 \\
    persona &264 &0.22 &58 &307 &0.34 &104 \\
    persona gpt4 &209 &0.25 &52 &252 &0.39 &98 \\
    name &2 &0.02 &0 &44 &0.91 &40 \\
    persona gold &203 &0.23 &47 &246 &0.37 &91 \\
    \bottomrule
    \end{tabular}
    \caption{Average length, rouge-Lsum score, and their product between prefix and shuffled prefix. \textbf{persona} is generated by \textsc{Zephyr}.}
    \label{tab:inter-persona-prefix-sim}
\end{table*}

\section{Leave-one-axis-out \textsc{MT}}\label{apx:leave-one-axis-out}
Personas sampled from the same axis may share more information within the axis than across axis. To understand how well \textsc{MT} generalizes across axis, we conduct leave-one-axis-out analysis: finetuning model on all but one axis, and evaluating on the one axis not trained on. For this experiment, we can ask two questions:

\begin{enumerate}
    \item Are some axes are harder to generalize? (Table~\ref{tab:leave-one-axis-out-cross-axis})
    \item Do different prefixes generalizes differently across axis? (Table~\ref{tab:leave-one-axis-out-cross-prefix})
\end{enumerate}

\paragraph{Are some axes harder to generalize?} As seen in Table~\ref{tab:leave-one-axis-out-cross-axis}, personal questions remain similar across seen and unseen persona (6/11 axis is worse in generalization case). This is more or less because personal questions do not have to adhere to axis, thus creating higher overlap between personas. Uniquely, axis politics performance drop significantly when it is unseen during training, likely because politicians are almost exclusively known for their political opinions, so their personal questions are more focused on politics. For common questions, we do see a more consistent drop in performance (8/11 axis), indicating domain specific contrast is still important to include in training to be able to perform well. However, the fact that all except one axis result in non-statistical significant difference indicate our methods does generalize quite well by leveraging natural language as a medium for preference specification.

\begin{table*}[!htp]
    \centering
    \small
    \begin{tabular}{ll|rrrrrrr}\toprule
    &\textbf{axis} &\textbf{\texttt{age}} &\textbf{\texttt{AI professors}} &\textbf{\texttt{diet}} &\textbf{\texttt{education}} &\textbf{\texttt{family}} &\textbf{\texttt{gender}} \\\midrule
    \multirow{3}{*}{Personal} &seen &0.63$\pm$0.07 &0.58$\pm$0.06 &0.63$\pm$0.06 &0.62$\pm$0.07 &0.62$\pm$0.07 &0.6$\pm$0.06 \\
    &unseen &0.64$\pm$0.06 &0.61$\pm$0.02 &0.56$\pm$0.08 &0.6$\pm$0.07 &0.63$\pm$0.03 &0.58$\pm$0.07 \\
    &p-value &0.769 &0.052 &0.059 &0.456 &0.339 &0.554 \\\midrule
    \multirow{3}{*}{Divergent} &seen &0.67$\pm$0.1 &0.65$\pm$0.09 &0.65$\pm$0.09 &0.67$\pm$0.09 &\textbf{0.67$\pm$0.09} &0.65$\pm$0.09 \\
    &unseen &0.62$\pm$0.08 &0.62$\pm$0.07 &0.71$\pm$0.11 &0.58$\pm$0.11 &\textbf{0.62$\pm$0.04} &0.62$\pm$0.06 \\
    &p-value &0.245 &0.469 &0.221 &0.092 &\textbf{0.028} &0.354 \\
    \midrule
    &\textbf{axis} &\textbf{\texttt{geo}} &\textbf{\texttt{politics}} &\textbf{\texttt{profession}} &\textbf{\texttt{religion}} &\textbf{\texttt{sports}} & \\
    \midrule
    \multirow{3}{*}{Personal} &seen &0.61$\pm$0.07 &\textbf{0.61$\pm$0.07} &0.61$\pm$0.07 &0.62$\pm$0.08 &0.62$\pm$0.07 & \\
    &unseen &0.6$\pm$0.07 &\textbf{0.54$\pm$0.04} &0.6$\pm$0.06 &0.65$\pm$0.03 &0.69$\pm$0.07 & \\
    &p-value &0.774 &\textbf{0.003} &0.792 &0.105 &0.071 & \\\midrule
    \multirow{3}{*}{Divergent} &seen &0.66$\pm$0.09 &0.65$\pm$0.08 &0.66$\pm$0.09 &0.66$\pm$0.09 &0.67$\pm$0.1 & \\
    &unseen &0.62$\pm$0.05 &0.65$\pm$0.05 &0.63$\pm$0.06 &0.64$\pm$0.11 &0.68$\pm$0.08 & \\
    &p-value &0.082 &0.868 &0.268 &0.718 &0.797 & \\
    \bottomrule
    \end{tabular}
    \caption{\textsc{MT(Zephyr)} with \textbf{persona gpt4} results finetuning with leave-one-axis-out set-up. Performance do not differ significantly between seen and unseen personas across most axis, indicating strong generalization. \textbf{Bolded} cells indicate statical significance}
    \label{tab:leave-one-axis-out-cross-axis}
\end{table*}

\paragraph{Do different prefixes generalizes differently across axis?} In Table~\ref{tab:leave-one-axis-out-cross-prefix}, we see similar results as Figure~\ref{fig:multitask-results} in the main text. There is also no significant difference between performance for persona trained vs. untrained. This indicates that \textsc{MT} allows generalization across axis.

\begin{table*}[!htp]
    \centering
    \small
\begin{tabular}{lr|rrrrrrrr}\toprule
& &\textbf{no prefix} &\textbf{tag} &\textbf{few-shot} &\textbf{persona} &\textbf{persona gpt4} &\textbf{name} &\textbf{persona gold} \\\midrule
Personal  &seen &0.54$\pm$0.07 &0.56$\pm$0.07 &0.61$\pm$0.08 &0.6$\pm$0.08 &0.61$\pm$0.07 &0.61$\pm$0.07 &0.63$\pm$0.07 \\
&unseen &0.55$\pm$0.07 &0.56$\pm$0.08 &0.6$\pm$0.08 &0.6$\pm$0.09 &0.61$\pm$0.07 &0.61$\pm$0.07 &0.63$\pm$0.08 \\
&p-value &0.912 &0.861 &0.823 &0.944 &0.859 &0.867 &0.872 \\
\midrule
Divergent &seen &0.53$\pm$0.07 &0.54$\pm$0.07 &0.62$\pm$0.08 &0.63$\pm$0.1 &0.66$\pm$0.09 &0.68$\pm$0.09 &0.69$\pm$0.08 \\
&unseen &0.52$\pm$0.08 &0.53$\pm$0.08 &0.6$\pm$0.07 &0.6$\pm$0.08 &0.64$\pm$0.08 &0.64$\pm$0.08 &0.67$\pm$0.09 \\
&p-value &0.789 &0.928 &0.497 &0.448 &0.518 &0.394 &0.597 \\
\bottomrule
\end{tabular}
    \caption{\textsc{MT (Zephyr)} with \textbf{persona gpt4} performance with leave-one-axis-out setup using different prefixes suggests strong generalization performance across axis.}
    \label{tab:leave-one-axis-out-cross-prefix}
\end{table*}

\section{Performance across demographic groups}
\label{apx:performance-per-demographic-attr}

\begin{figure*}[!htp]
    \centering
\includegraphics[width=0.32\linewidth]{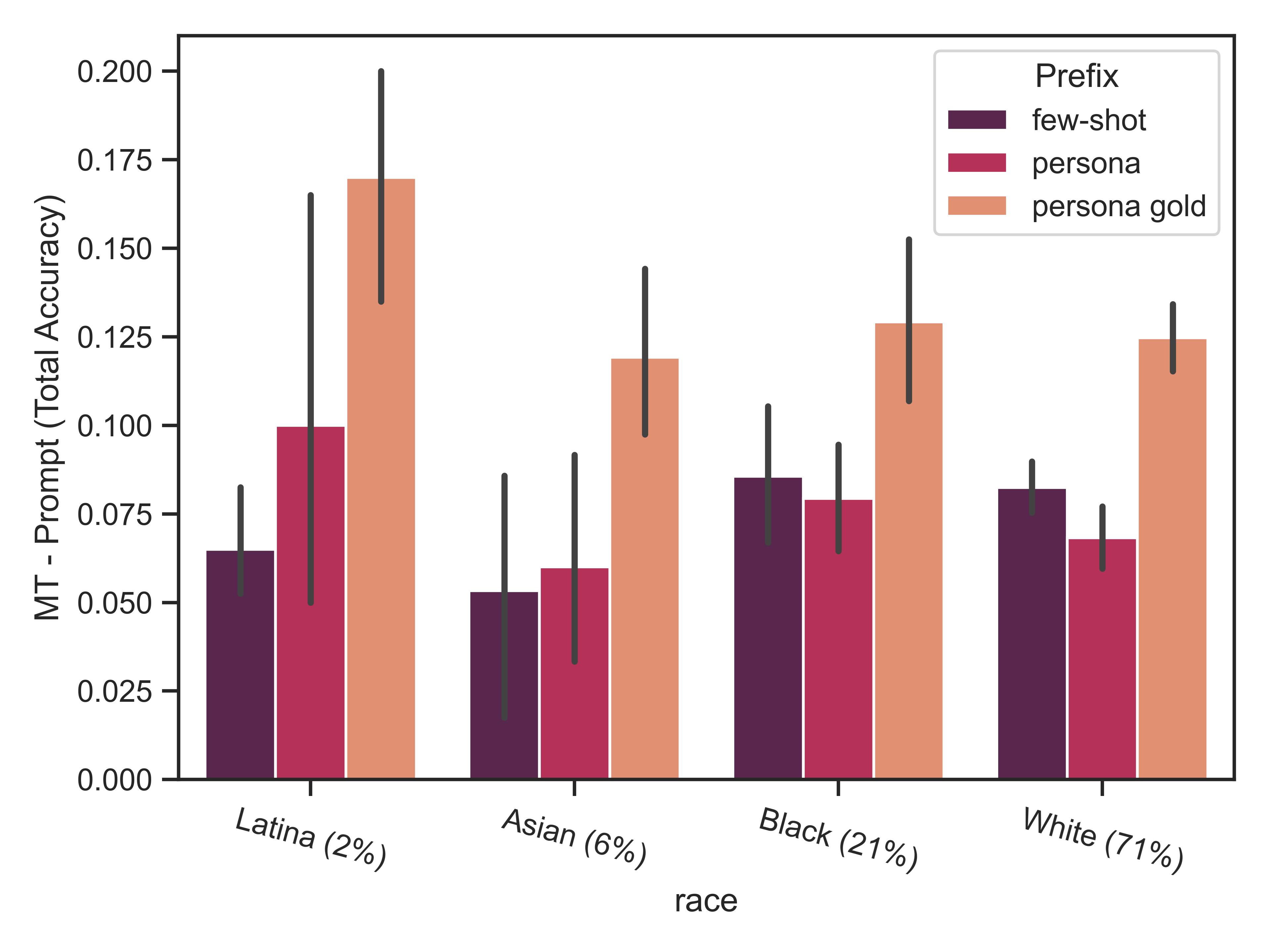}
\includegraphics[width=0.32\linewidth]{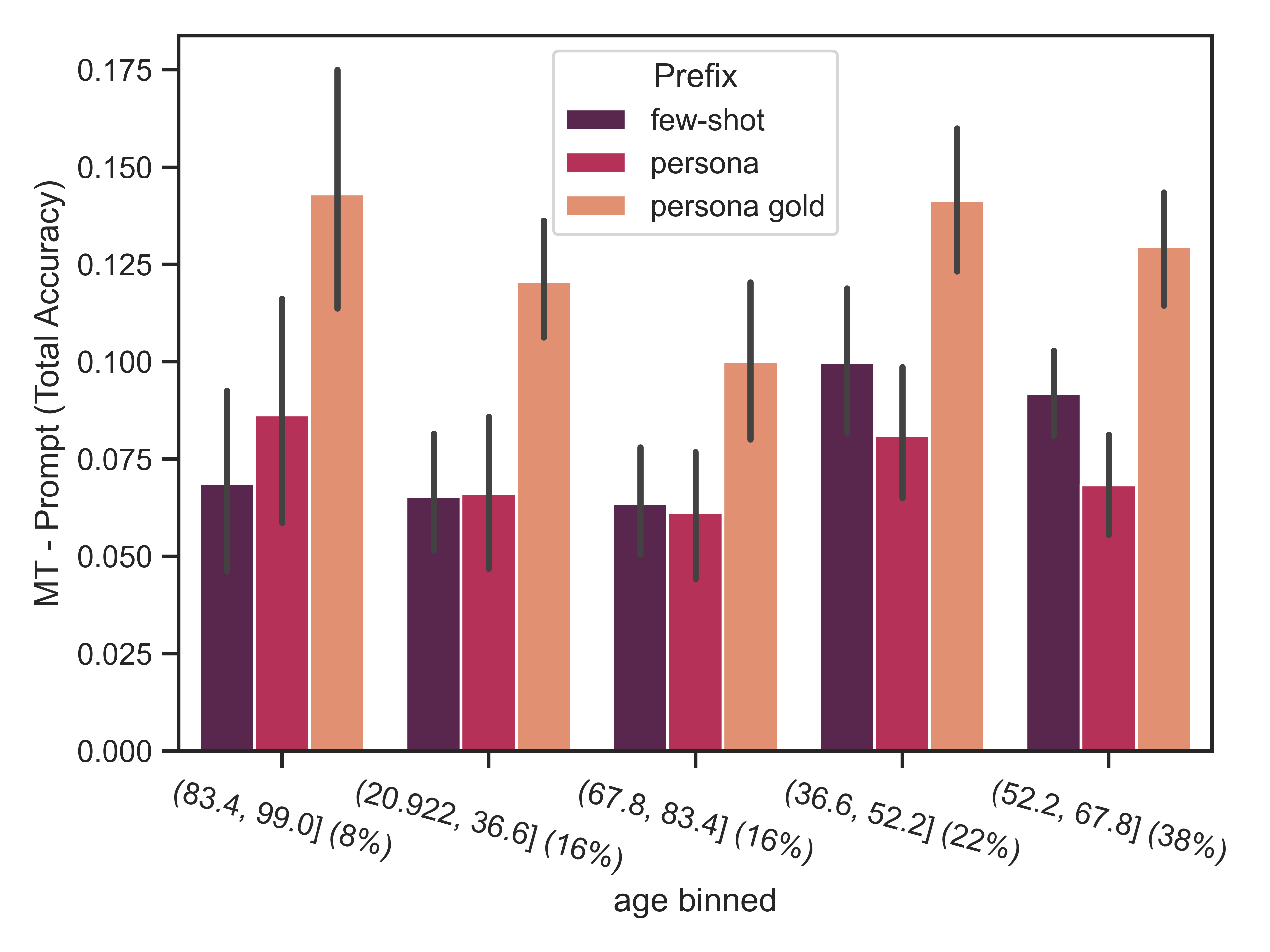}
\includegraphics[width=0.32\linewidth]{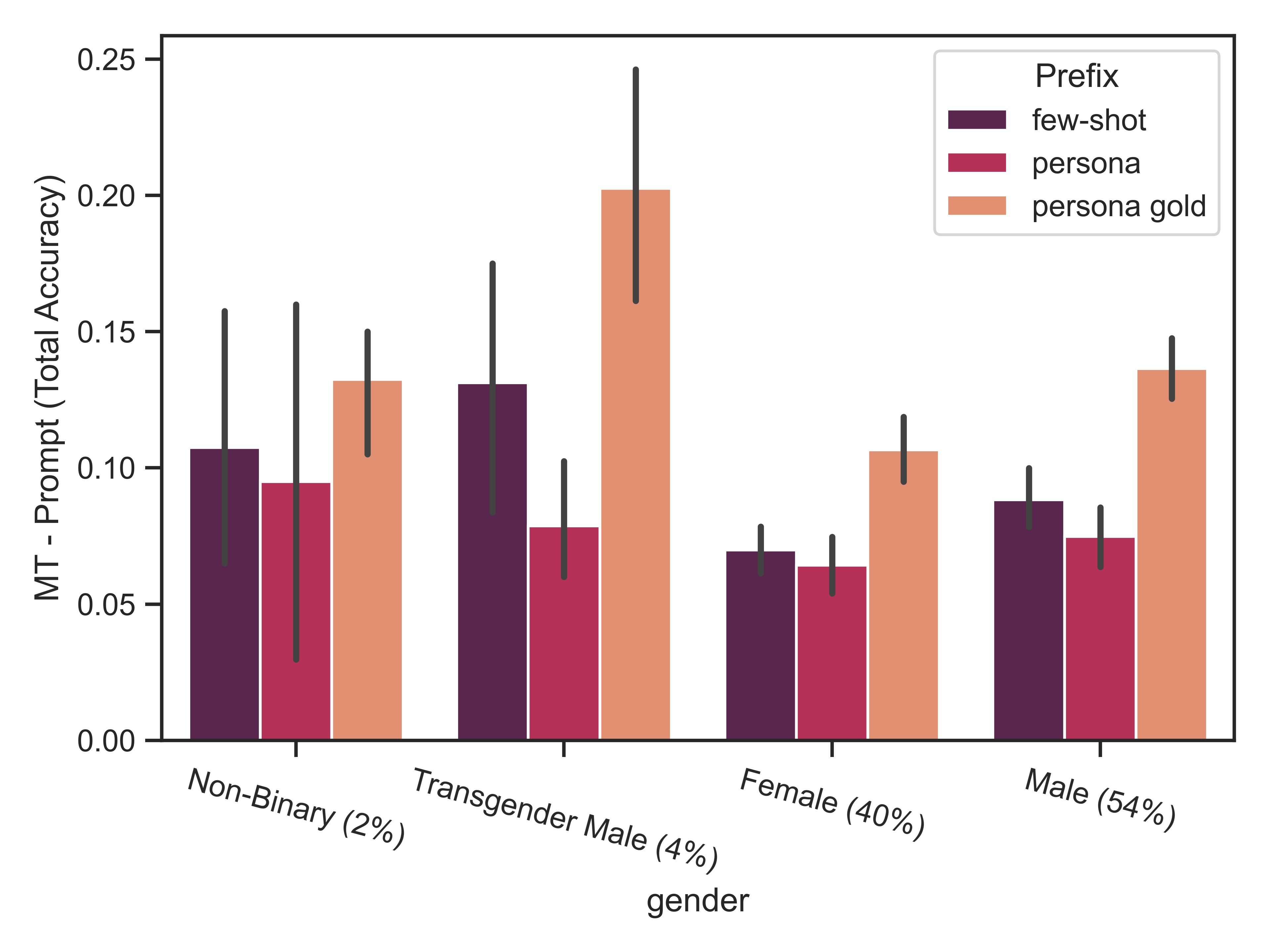}
\includegraphics[width=0.32\linewidth]{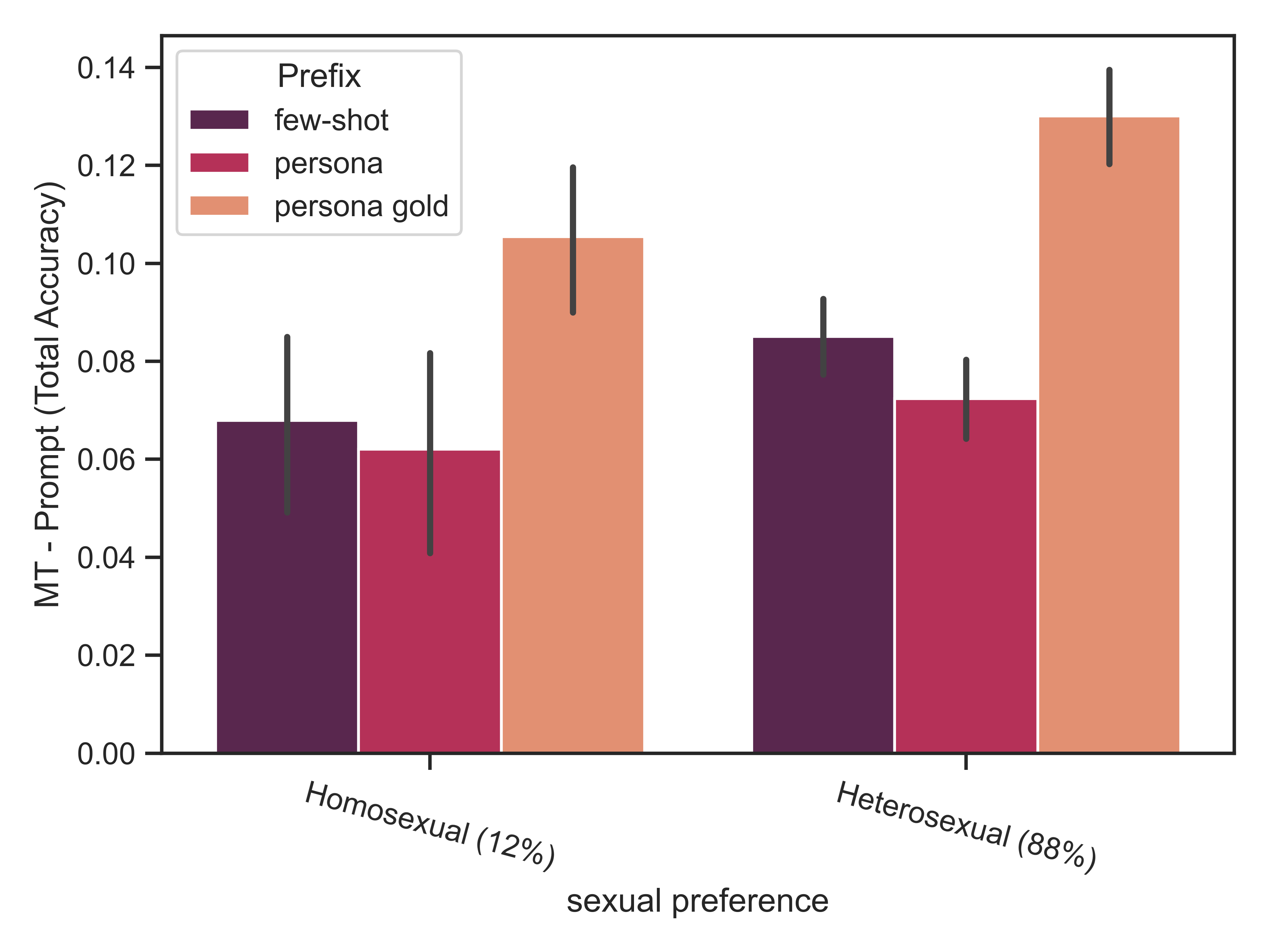}
\includegraphics[width=0.32\linewidth]{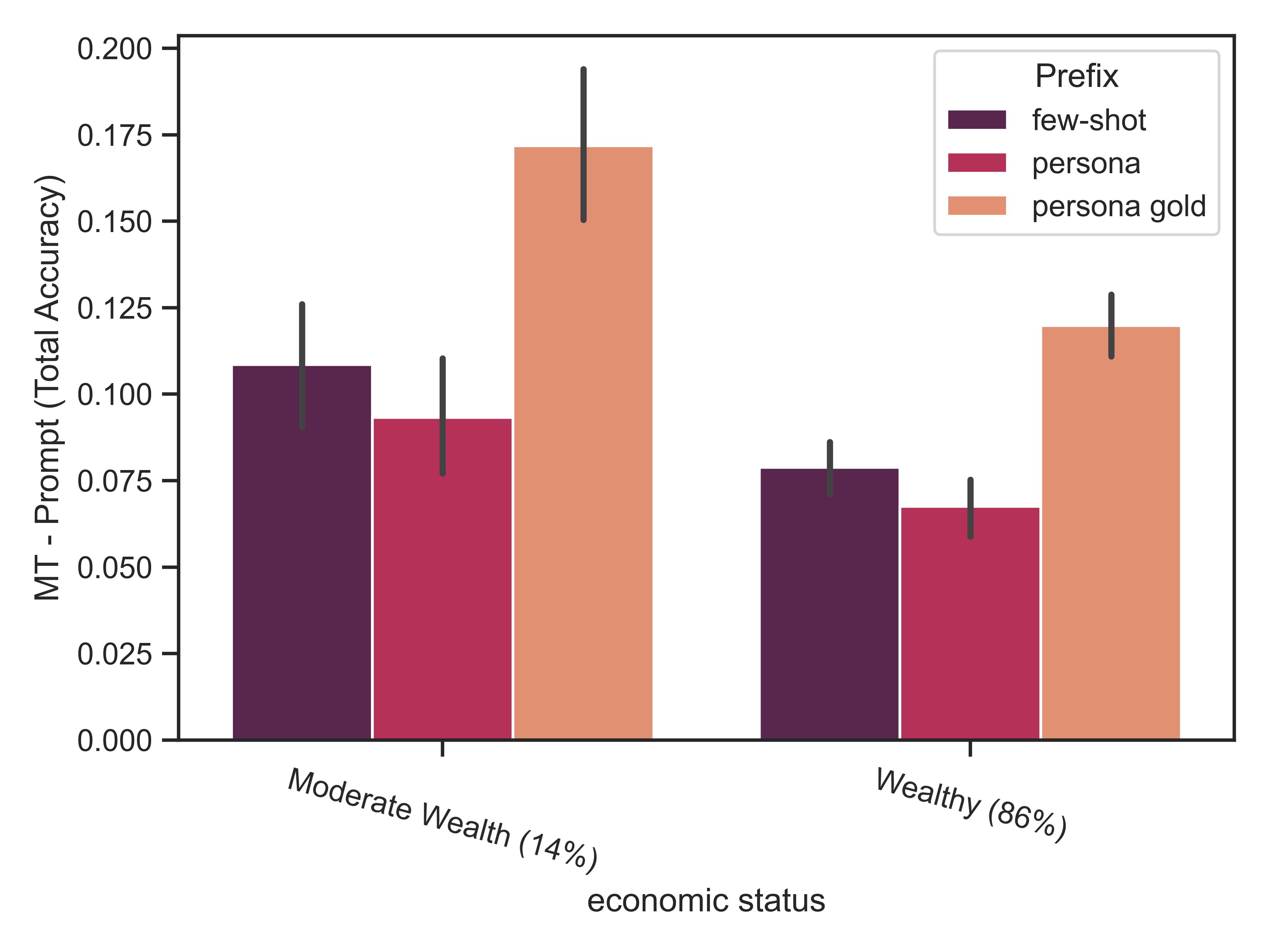}
\includegraphics[width=0.32\linewidth]{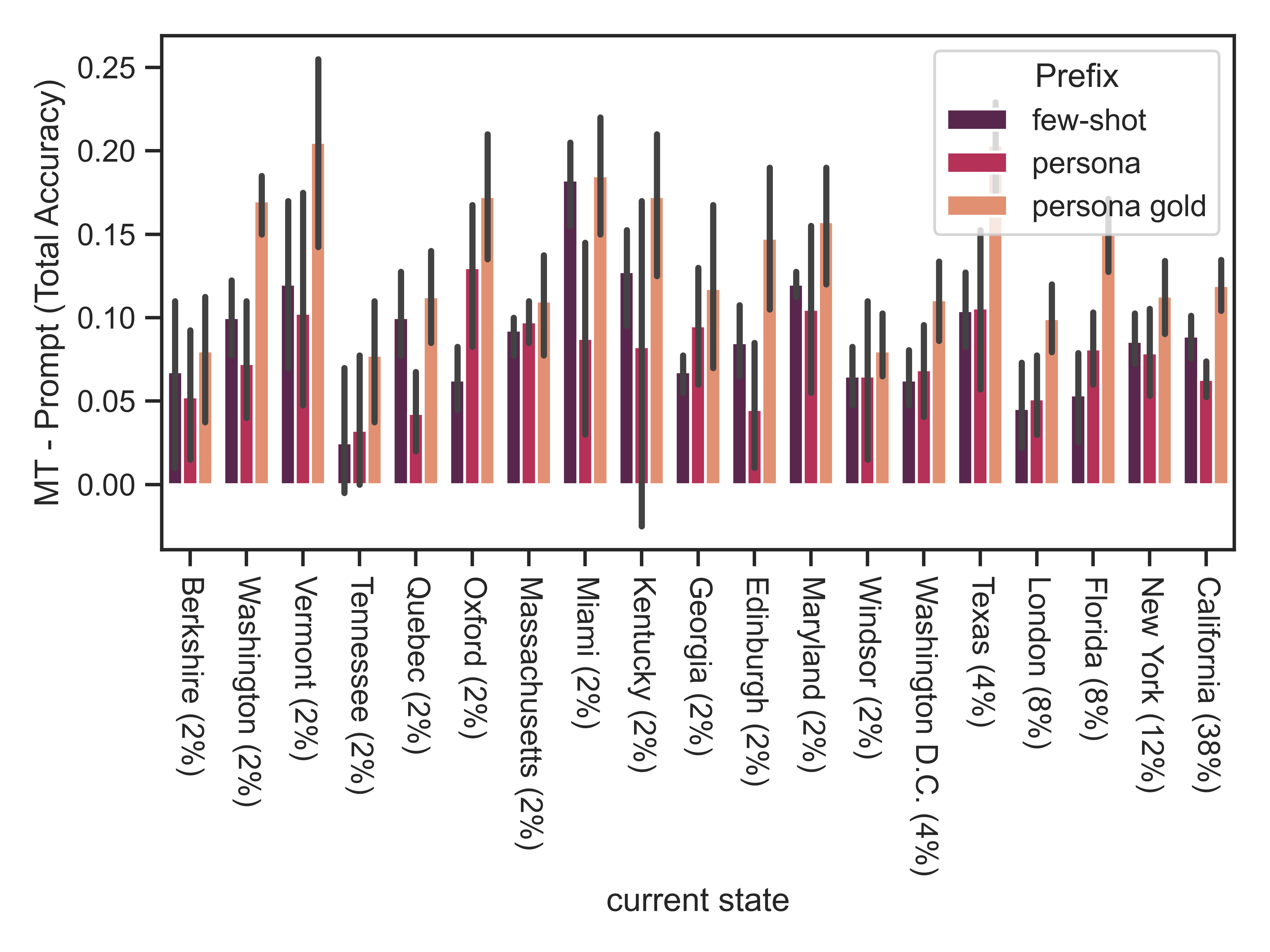}
\includegraphics[width=0.32\linewidth]{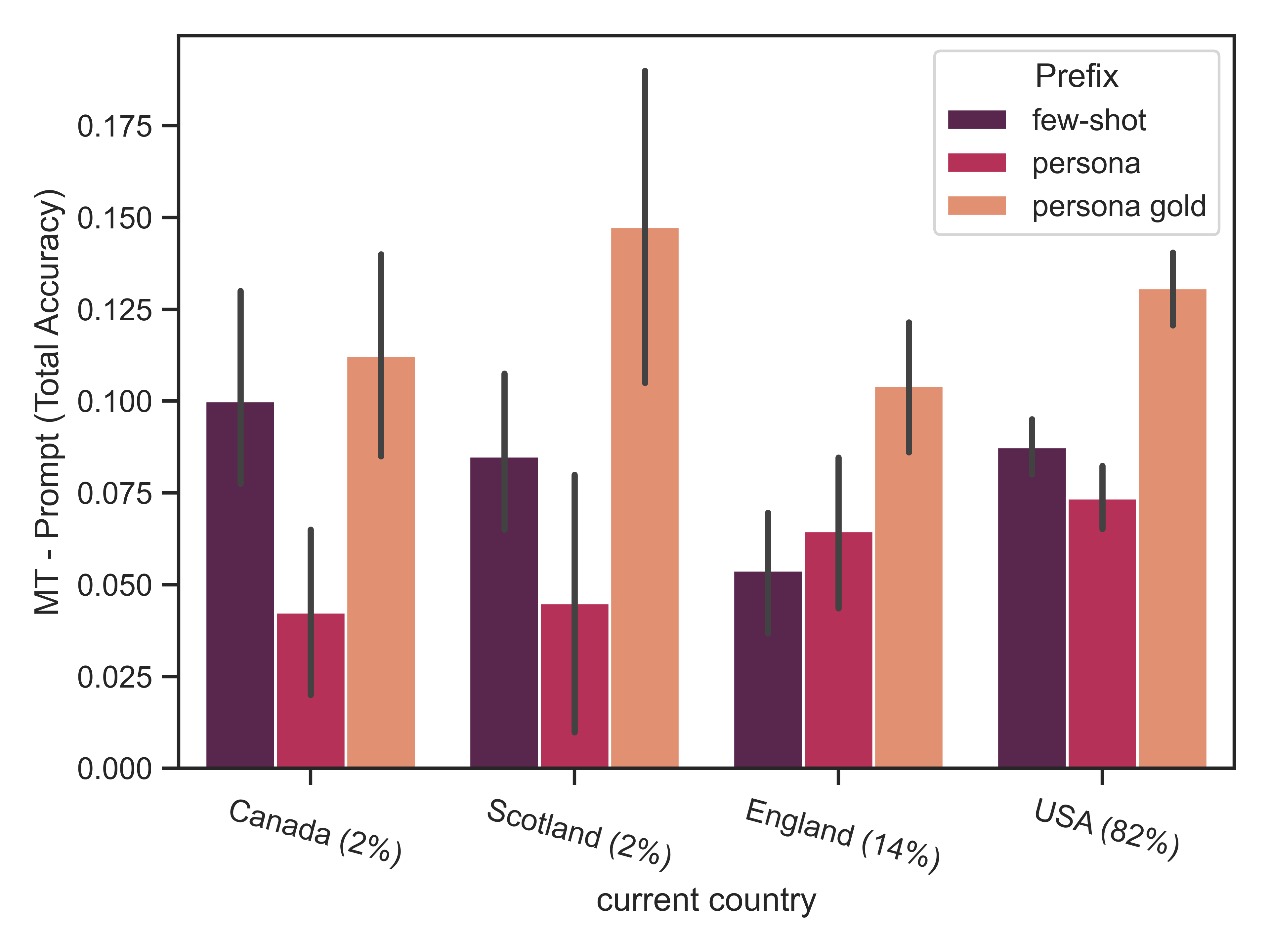}
\includegraphics[width=0.32\linewidth]{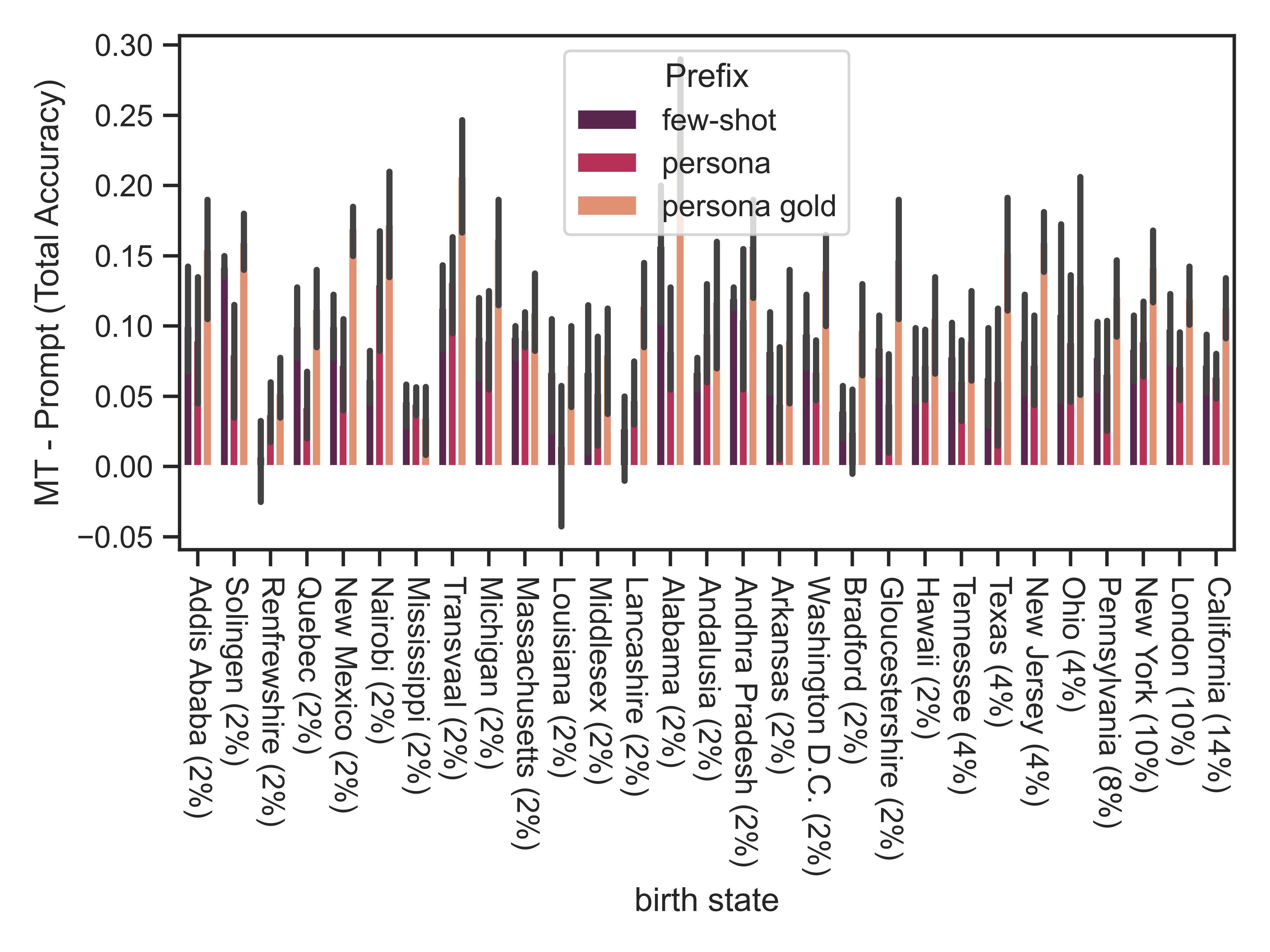}
\includegraphics[width=0.32\linewidth]{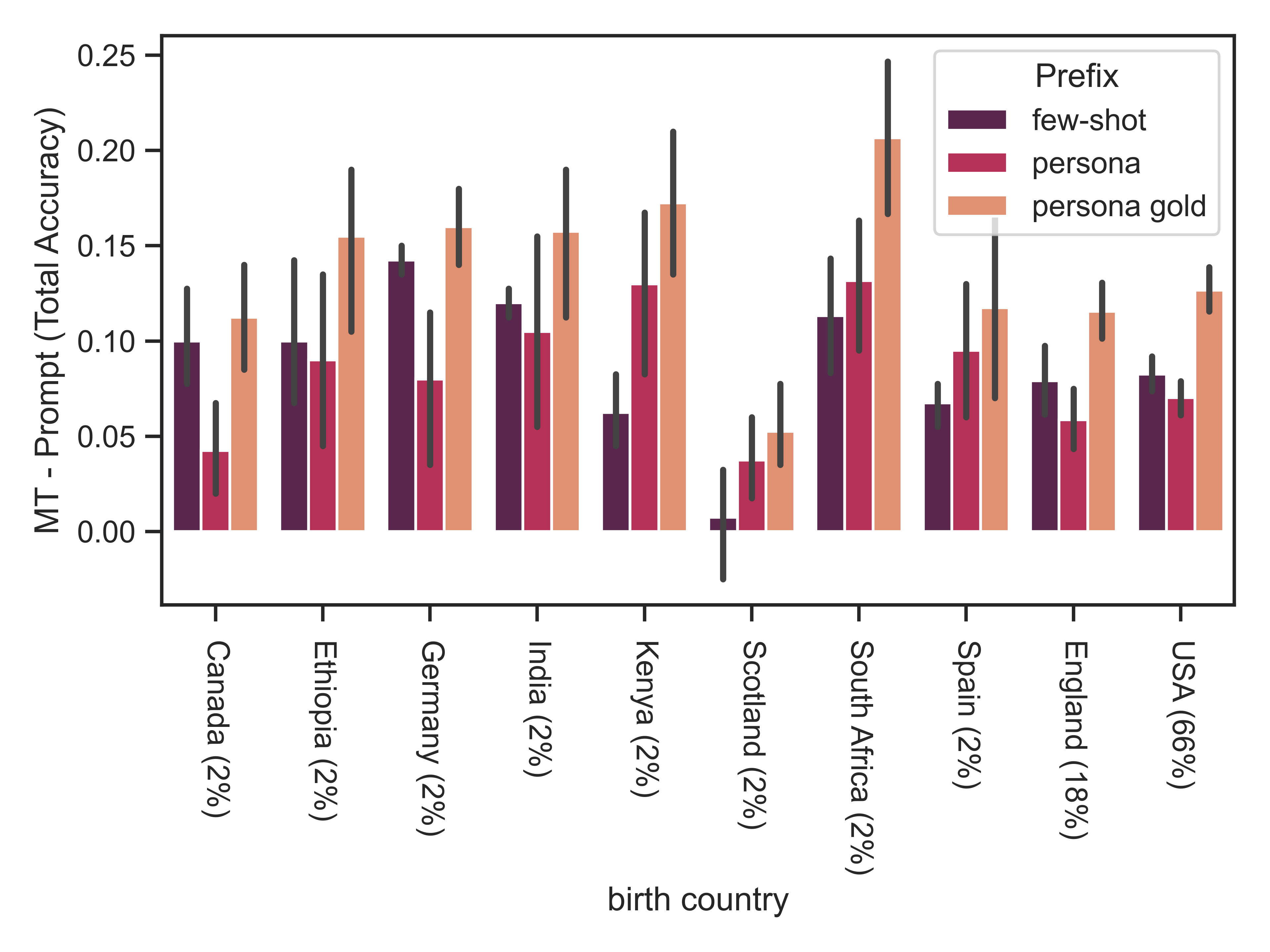}
\includegraphics[width=0.32\linewidth]{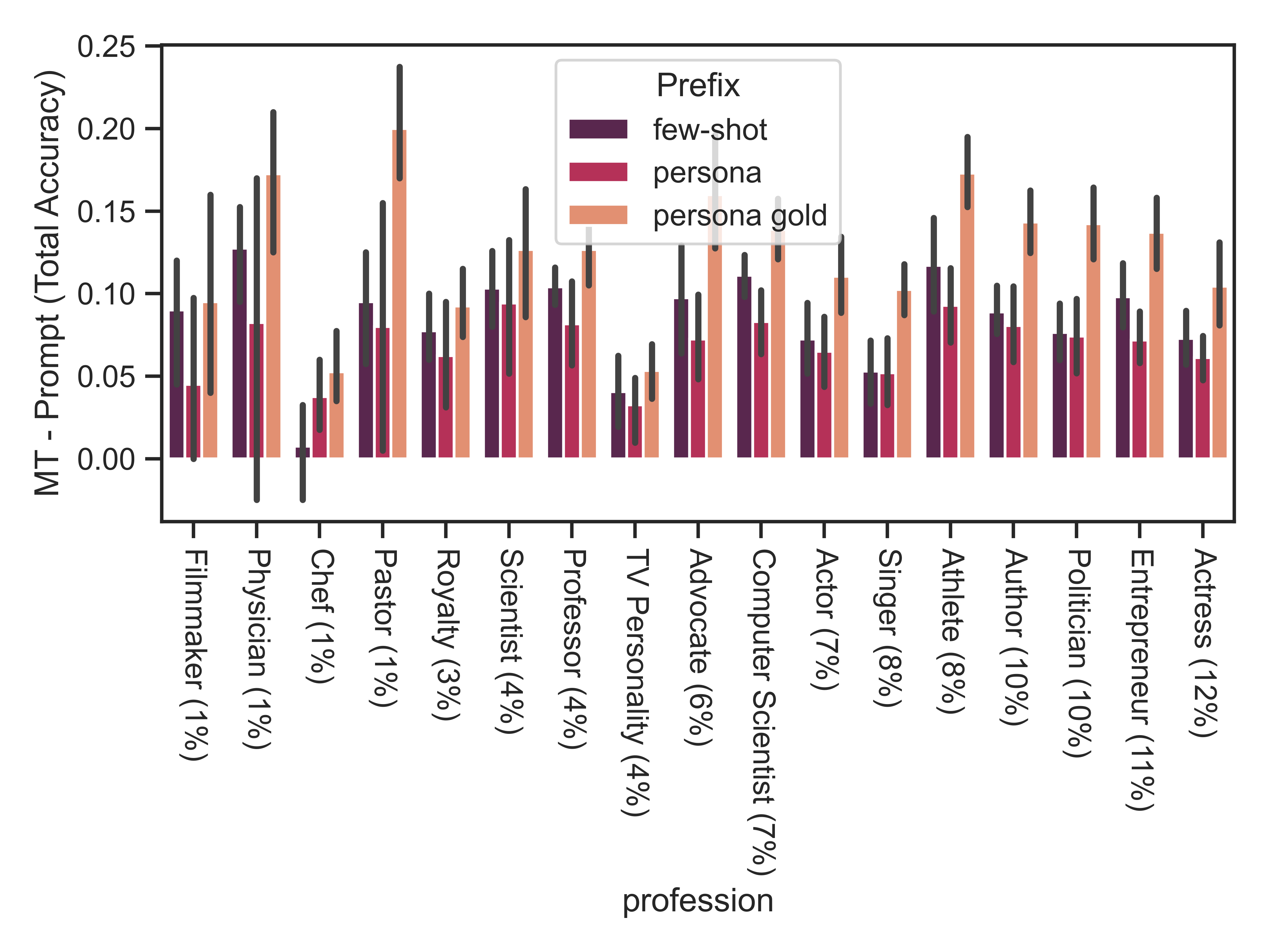}
\includegraphics[width=0.32\linewidth]{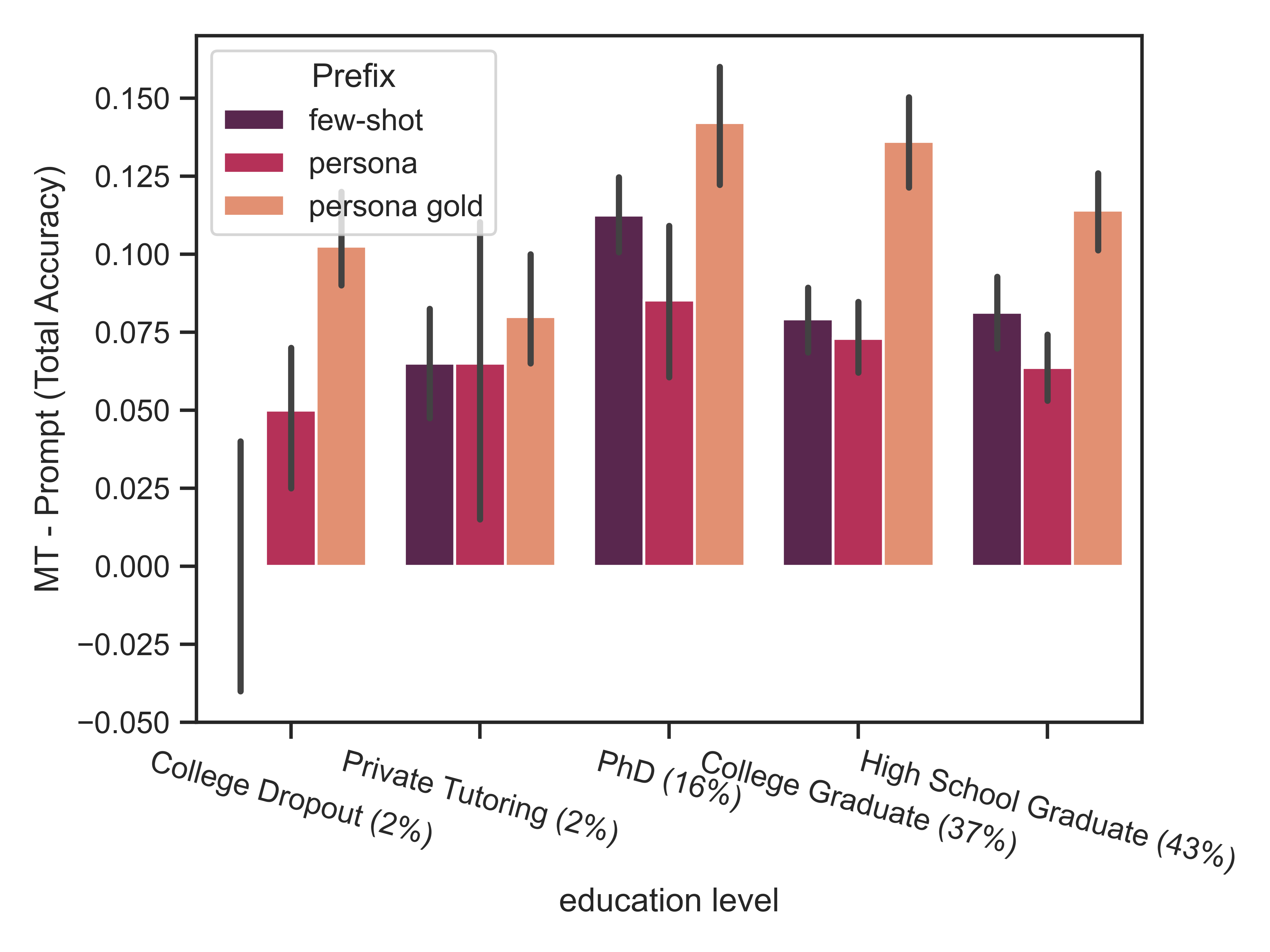}
\includegraphics[width=0.32\linewidth]{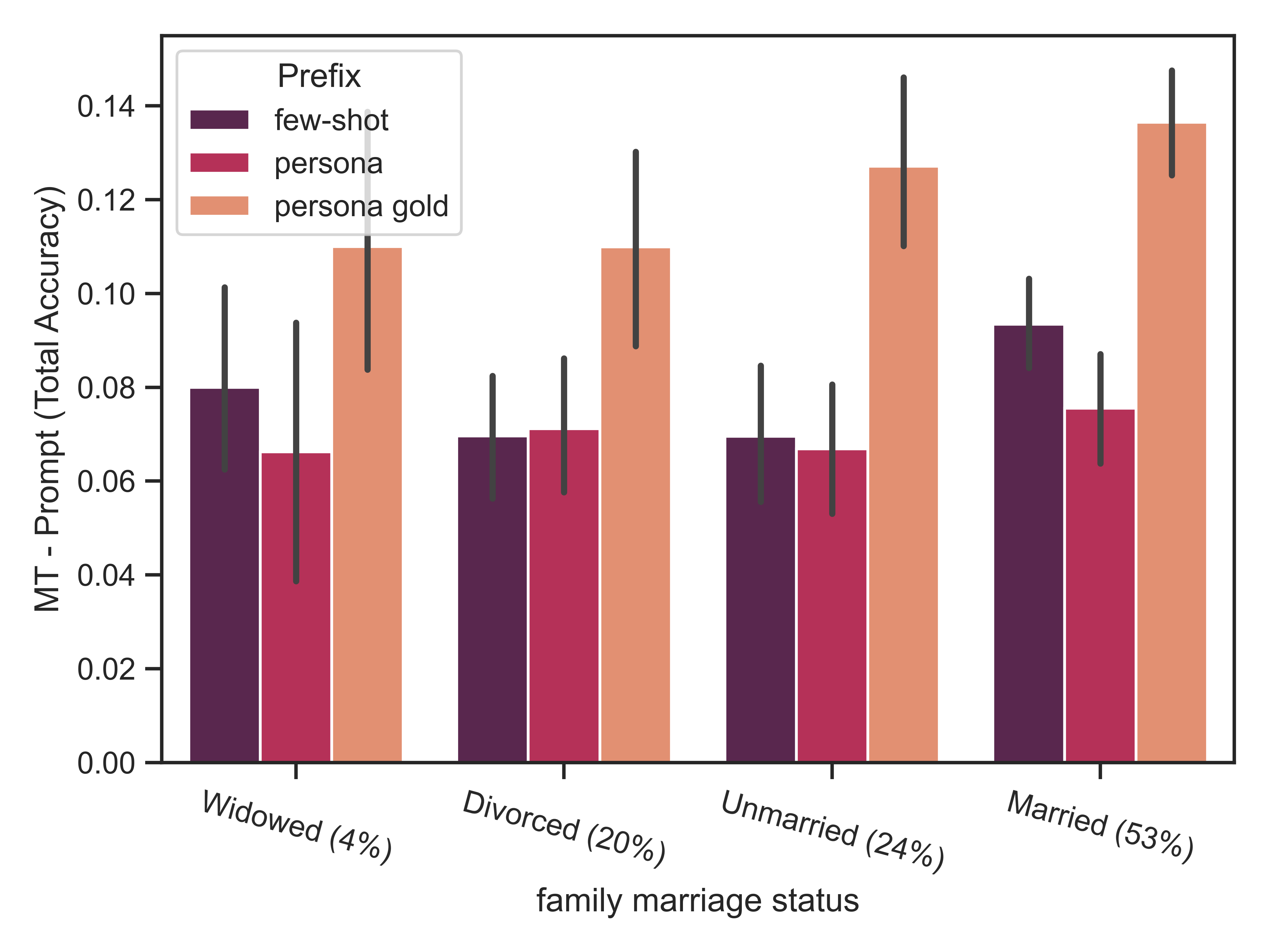}
\includegraphics[width=0.32\linewidth]{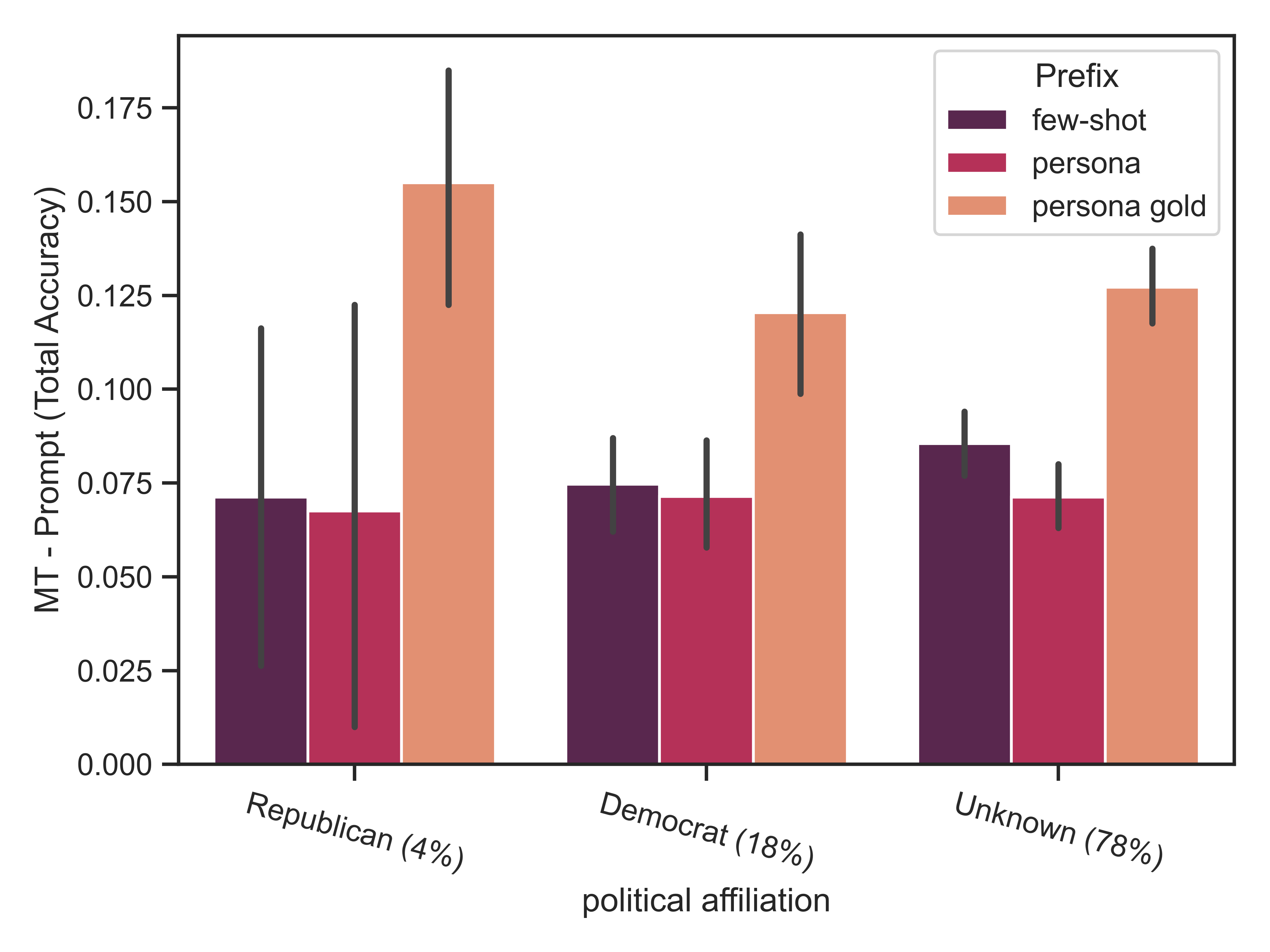}
\includegraphics[width=0.32\linewidth]{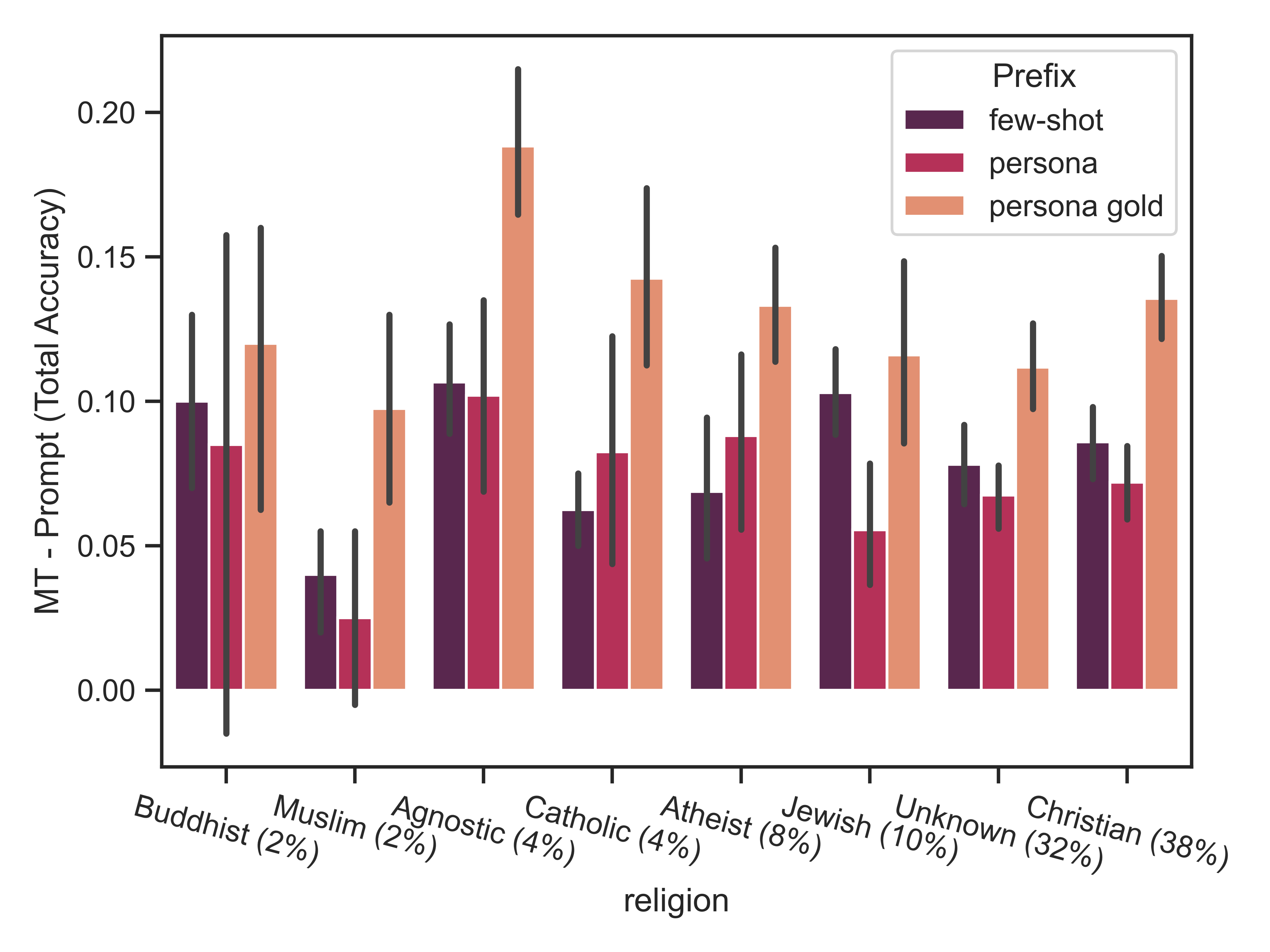}
\includegraphics[width=0.32\linewidth]{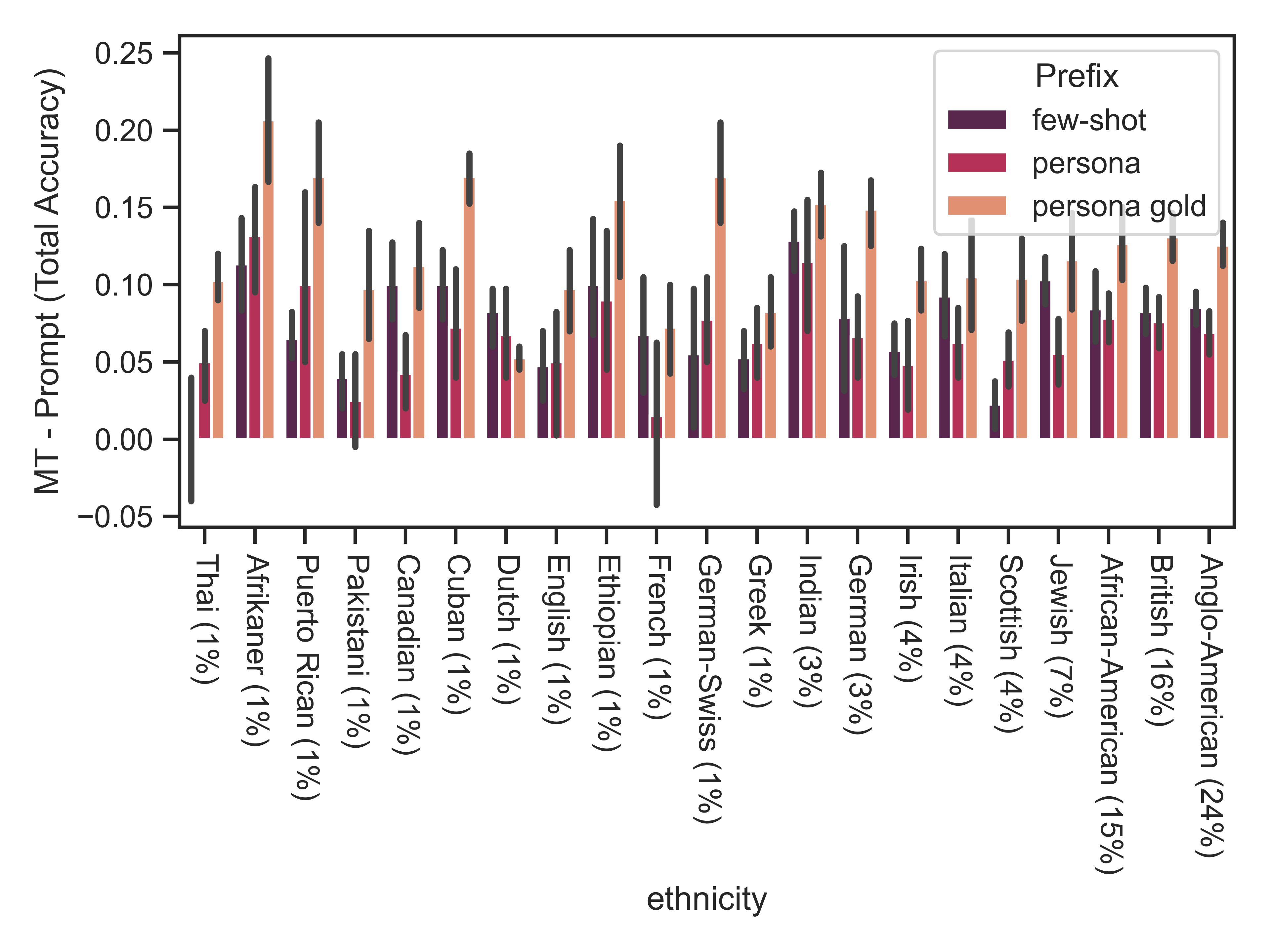}
    \caption{\textsc{MT(Zephyr)} model improvements over prompting un-finetuned models with \textbf{few-shot}, \textbf{persona}, and \textbf{persona gold} performance (unseen during training) across different demographic attributes. Bars are sorted according to attribute frequency.}
    \label{fig:performance-across-demographics}
\end{figure*}

Personalized alignment performance might greatly depend on the demographics of the people included in the training data. To understand how the model does across different demographic attributes, we plot the improvement of \textsc{MT(Zephyr)} models over their baseline model with prompting across different prefixes across different demographic groups (Figure~\ref{fig:performance-across-demographics}). 

\section{Generation Evaluation on \textsc{Zephyr}}
\label{apx:gen-eval}

To test whether reward accuracy really reflect generational improvements, we test a good performing prefix \textsc{MT(Zephyr)} with \textbf{persona gpt} against our baseline model \textsc{Zephyr}, as we only need to show that the order of performances remain similar.
We curate one divergent and one personal question for all personas in our dataset to evaluate generations. We use \textsc{Zephyr} and \textsc{MT} (persona not trained),
with and without \textbf{persona gpt4} prefix, and evaluate using GPT4-as-personal-judge (Results in Table~\ref{tab:gen-eval}). Consistent with the findings in ~\ref{fig:multitask-results}, \textsc{MT} with \textbf{persona gpt4} performs the best on average, and degrades to baseline after removing prefixes, which are the keys to personalization. However, \textsc{Zephyr} with \textbf{persona gpt4} is worse than no prefix, indicating prompting is not always effective for personalization for small models. In Appendix~\ref{apx:qualitative-analysis-generation}, we confirm this qualitatively.

\begin{table}[!htb]
    \centering
    \small
    \begin{tabular}{l|r|rrrr|r}\toprule
    & &\multicolumn{2}{c}{\textsc{Zephyr}} &\multicolumn{2}{c}{\textsc{MT}} &Avg. \\\cline{3-7}
    model &prefix & F &T &F &T & \\\hline
    \multirow{2}{*}{\textsc{Zephyr}} &F &- &55 &53 &36 &48 \\
    &T &45 &- &38 &42 &42 \\
    \multirow{2}{*}{\textsc{MT}} &F &47 &\textbf{62} &- &32 &47 \\
    &T &\textbf{64} &58 &\textbf{68} &- &\textbf{63} \\
    \bottomrule
    \end{tabular}
    \caption{Pairwise win-rate (\%) between model generations. \textbf{F}=no prefix, \textbf{T}=prefixed (\textbf{persona gpt4}). \textsc{MT} with prefix outperforms all baselines.}
    \label{tab:gen-eval}
\end{table}

\section{Qualitative Analysis of Generations}
\label{apx:qualitative-analysis-generation}

We include two sets of generation results for Alexandria Ocasio-Cortez (AOC) and Serena Williams as an example to demonstrate the effect of personalization with our trained models. Samples are all generated with temperature sampling of 1.0 and with maximum length cut off at 512 tokens. The models we include are the baseline model (\textsc{Zephyr}), and multitask-trained model (\textsc{MT(Zephyr)}), inferenced with and without prefix \textbf{persona gpt4}.

\paragraph{Persona inference successfully uncovers underspecified information}
In the first example (Table~\ref{tab:generation-1-aoc}), we can see that \textbf{persona gpt4} successfully infers that AOC is a liberal politician keen on looking for ``equitable solution to socio-economic'' problems. Similarly, Table~\ref{tab:generation-2-serena} shows that GPT4 is able to infer most of Serena's background as being possibly a professional athletes.

\paragraph{\textsc{MT} uses persona information much more effectively than \textsc{Zephyr}}
With successful persona inference, we see that \textbf{\textsc{MT} + persona gpt4} provides a generation is much more customized. In Table~\ref{tab:generation-1-aoc} we see the generation is much more supportive of labor rights, additionally including labor strikes led by ``women and people of color fighting against systemic inequality and exploitation''. However, \textbf{\textsc{Zephyr} + persona gpt4} did not contextualize the strikes as well and deviates very little from \textbf{\textsc{Zephyr}}. In Table~\ref{tab:generation-2-serena}, we see similar pattern. With \textbf{\textsc{Zephyr} + persona gpt4}, despite mentioning ``as someone deeply committed to the world of sports'', the content of suggestions mostly remain the same. \textbf{\textsc{MT} + persona gpt4} however, is able to suggest much more relevant tactics from ``mentor female athletes'', ``pledge a portion of ... contract'' to dedicated charities, to collaborating with federations and engage with the public utilizing her social influence.

\paragraph{\textsc{MT} without persona reverts back to baseline performance} As seen in both tables, \textbf{\textsc{MT}}'s generation is very similar to \textbf{\textsc{Zephyr}}'s. This demonstrates that our dataset does not have underlying bias, and that multi-task prefix training is an effective way of providing personalization \emph{when needed}.

\begin{table*}
    \centering
    \tiny
    \begin{tabular}{p{0.5cm}p{14.6cm}}\toprule
name &Alexandria Ocasio-Cortez \\\midrule
persona gpt4 &The user seems to be well-informed and interested in various political, economic, and technological topics, displaying a \underline{progressive or liberal bias}. He/she shows concern about issues such as globalization, international health crises like COVID-19, national debt, new technologies like blockchain, and environmental initiatives like the Green New Deal. The user could likely be interested in policymaking, economics, progressive politics, environmental sustainability, or could be working in a related field. His/her way of delving into complex issues shows an analytical and critical mind, which suggests an educated individual, who values fairness, economic balance, and technological innovation. When interacting with an AI assistant, this user prefers comprehensive, factual responses that address the larger context of the question asked. He/she would appreciate not only the 'what' but also the 'why' and 'how'. \underline{An emphasis on equitable solutions to socio-economic problems}, sustainable practices, technological advances, and information about progressive politics and policies would often be preferred. Merely factual or neutral responses may not be well-received. The assistant's ability to present information from a progressive viewpoint might be highly appreciated, contributing to a more engaging and satisfactory user experience. Consequently, it can be inferred that this user values informed, contextually-rich, and progressive-leaning discourse. These preferences hint at someone who cares about issues beyond their immediate environment and seeks to understand and solve complex global challenges through a lens of fairness, sustainability, and interconnectedness. \\\midrule

prompt &What are some major successful union strikes in the past two decades and their effects on respective workplaces? \\\midrule

\textsc{Zephyr} &1. Verizon Communications Strike (2016): The Communications Workers of America (CWA) and the International Brotherhood of Electrical Workers (IBEW) went on a 45-day strike against Verizon Communications in August 2016. The strike involved approximately 39,000 workers and led to the company's agreement to provide job security, pay increases, and better benefits for workers. 2. Chicago Teachers Union Strike (2012): In 2012, the Chicago Teachers Union went on a seven-day strike. The strike involved over 26,000 teachers and support staff and resulted in a new contract that included pay raises, smaller class sizes, and increased resources for schools in low-income areas. 3. Deere \& Company Strike (2012): In 2012, the United Automobile, Aerospace, and Agricultural Implement Workers of America (UAW) went on a six-week strike against John Deere. The strike involved approximately 10,000 workers and led to a new contract that included wage and benefit increases, a lump sum payment, and job security provisions. 4. American Airlines Mechanic and Related Services Strike (2019): The Transport Workers Union (TWU) and the International Association of Machinists and Aerospace Workers (IAM) went on a 35-day strike against American Airlines in October 2019. The strike involved approximately 30,000 mechanics and related services workers and resulted in a new contract that included job security provisions, improved benefits, and pay increases. 5. Atlanta Educators Strike (2018): The Atlanta Board of Education and several education unions went on a one-day strike in January 2018. The strike involved approximately 1,200 educators and resulted in a new contract that included pay raises, improved benefits, and job security provisions. These strikes led to improvements in working conditions, job security, and wages for workers in these respective industries, demonstrating the power of collective action in achieving better outcomes for workers and communities. \\\midrule

\textsc{Zephyr} + persona gpt4 &Over the past two decades, there have been several notable successful union strikes across various industries that have had a significant impact on their respective workplaces. Here are a few examples: 1. Verizon Strike (2016): Over 39,000 Verizon Communications workers, including technicians, customer service representatives, and call center agents, went on strike in April 2016. The employees were demanding higher wages, job security, and better benefits. After four weeks of striking, Verizon agreed to a new contract that included salary increases, job protection, and improvements in retirement benefits. As a result of the strike, Verizon reported a loss of \$110 million in revenues and experienced decreased customer satisfaction rates, as many Verizon services experienced downtime during the strike. However, the company maintained that the revenue loss would not impact its full-year earnings guidance. 2. Nurses at Kaiser Permanente Strike (2015): In late 2015, around 4,000 nurses at several California-based Kaiser Permanente hospitals went on a five-day strike. The nurses were demanding a fair contract that included higher wages, staffing levels, and patient care issues. Kaiser Permanente is a large, integrated managed care consortium, providing healthcare services to around 12 million members. The nurses' strike resulted in a significant impact on medical services at Kaiser Permanente. The staffing shortages led to an overburdened emergency department, and some patients reported waiting for several hours to receive medical attention. Furthermore, elective procedures and appointments were canceled or postponed. 3. Chicago Teachers Union Strike (2012): In 2012, over 26,000 teachers from the Chicago Public Schools System went on strike for seven days. The educators were demanding higher salaries, smaller class sizes, and job security, as the district was planning to replace tenure and seniority-based staffing policy with a proposed new system that would instead measure teachers' performance based on student test scores. The strike led to a loss of several days' instructional time for students and created widespread disruption in the city's schools. However, the Chicago Teachers Union's efforts resulted in increased salaries, a smaller class size, and a revised principal evaluation system. \\\midrule

\textsc{MT} &1. Verizon Strike (2016): The Communications Workers of America (CWA) and the International Brotherhood of Electrical Workers (IBEW) went on a six-week strike against Verizon Communications, one of the largest telecommunications companies in the US. The strike resulted in Verizon agreeing to provide job security to about 35,000 employees, as well as granting wage increases and improvements in retiree healthcare benefits. 2. Chicago Teachers Union Strike (2012): More than 25,000 teachers and support staff in Chicago walked out for seven days to demand better working conditions, job security, and higher pay. The strike resulted in a new contract that reversed some of the negative effects of the previous one implemented under former Chicago Mayor Rahm Emanuel. The agreement included salary increases as well as provisions to lessen class sizes, eliminate the use of merit pay, and put a hold on forced school closures. 3. John Deere Strike (2012): Approximately 10,000 workers at John Deere factories across Iowa, Illinois, Wisconsin, and Kansas walked out for six weeks over various work rules and contracts. The strike was the biggest challenge faced by Deere \& Co., a top agricultural machinery maker, in two decades. The strike forced management to agree to double-digit wage increases, better health and retirement benefits, and other employee-friendly policies. 4. Atlanta Journal-Constitution Strike (2000): More than 250 graphic artists, photographers, and typographical workers of The Atlanta Journal-Constitution newspapers walked out for eight weeks. The protest against the management’s desire to replace staffers and their union status brought results as the Cox Enterprises, who owned the newspaper then, eventually agreed to a contract granting new hires lower wages but maintaining senior employee rights. 5. NYC Health + Hospitals Strike (2014): The Patient Care Facilities union representing 8,000 hospital aides threatened to strike after months of negotiations over healthcare, wage, and job security issues. The strike lasted seven days, resulting in new contracts with pay boosts, job controls, and assurances about retirement benefits to around 35,000 hospital staffers. All these strikes and negotiations resulted in concrete agreements \\\midrule

\textsc{MT} + persona gpt4 &\underline{Yes, let's talk about labor rights and workers' power!} In recent years, there's been a resurgence of union organizing and strikes across the country, fighting for higher wages, better benefits, and safer working conditions. Here are a few notable examples: 1. Chicago Teachers Union Strike of 2012: This historic strike saw nearly 30,000 teachers and support staff walk off the job to demand smaller class sizes, improved working conditions, and fair evaluations. The CTU won significant victories, including a 17\% raise and more autonomy over their schools and curriculum. 2. Oakland Education Strike of 2019: Over 35,000 teachers and staffers in Oakland public schools followed the CTU's lead, staging a multi-day strike for higher wages, more resources for students with disabilities, and reduced class sizes. After a tense negotiation process, they ultimately secured a 12\% raise, smaller class sizes, and a new learning center for students with special needs. 3. Strikes at Amazon Warehouses in 2019: Workers at Amazon fulfillment centers in New York, Chicago, and other cities coordinated short walkouts and sickouts in efforts to protect their jobs and demand fair treatment amid Amazon's rapid expansion. While not officially ``strikes,'' these actions were nonetheless pivotal in highlighting the abysmal working conditions Amazon has long been criticized for. These strikes are helping to reinvigorate a long-standing tradition of labor solidarity and collective bargaining power. \underline{Notably, most of these campaigns have been led by women and people of color fighting against systemic inequality and exploitation.} This resurgent movement is crucial in confronting the greed and power of corporations like Amazon and holding our elected leaders accountable for workers' rights. \underline{These strikes also help shed light on and improve wider societal issues such as the struggle to provide educational resources to low income neighborhoods.} Overall, unions are vital in promoting strong economic stability, job security, and decent working conditions for all working Americans, and it's critical that we support and uplift them in their mission. \\
\bottomrule
\end{tabular}
    \caption{Qualitative comparison of generations between different models for a prompt from Alexandria Ocasio-Cortez. We \underline{underline} portions of the text that emphasize successful inference persona or shows effect of personalization.}
    \label{tab:generation-1-aoc}
\end{table*}

\begin{table*}
    \centering
    \tiny
    \begin{tabular}{p{0.5cm}p{14.6cm}}\toprule
name &Serena Williams \\\midrule
persona gpt4 &\#\# Basic Information From the questions asked, the user seems to be \underline{a professional in the world of sports}, likely a coach, a sports speaker, or \underline{perhaps even a professional athlete}. They could also be managing a professional athlete. They are looking for tips for a speech at a youth sports conference, asking about sports rehabilitation and high-altitude training, implying a deep involvement in sports-related activities, which require these topics. The interests extend to social media strategies for professional athletes, suggesting an attempt to increase their public image or that of someone they manage—indicating a possible role in sports marketing or social media advocacy. \#\# Preferences and Interactions The user prefers detailed responses that are narrative and rich in specific advice and examples. This person places value on information that is both relevant and comprehensive, preferring answers that delve deep into the context of their questions - each one being quite specific. Their preference for comprehensive responses that include a motivational narrative (seen in the preferred response for the first question) suggests a level of emotional intelligence and an appreciation for inspiring stories. Also, their preference for high-level strategies, preventative approaches, and a focus on technological advances implies a forward-thinking mindset and an inclination towards innovation in their field. They value staying up-to-date, being informed, calculated, and strategic, which can contribute to their professional capacity in the world of sports. This might also imply that they would interact proactively with their AI assistant, expecting thoroughly detailed, carefully crafted responses that are examplary and educational, allowing them to gain deeper insights and perspectives into the topics they inquire about. As a probable professional within sports or sports management, they would likely hold values such as perseverance, teamwork, dedication, innovation, and personal growth, all of which are commonly associated with sports activities and management. \\\midrule
prompt &How can I get involved in charity work that supports gender equality in sports? \\\midrule
\textsc{Zephyr} &1. Research organizations: Start by doing some research to find organizations that focus on promoting gender equality in sports. Look for organizations that have a specific focus on addressing issues related to women's participation, representation, and opportunities in sports. Some examples of such organizations include Women's Sport Foundation, International Working Group for Women and Sport, and Coalition to Unleash Women's Potential in Sports. 2. Volunteer opportunities: Once you've found a few organizations, look for volunteer opportunities that match your skills and preferences. Some opportunities may involve coaching or mentoring girls and women in sports, organizing events, fundraising, or advocacy. 3. Donate: If you're unable to volunteer, you can still support these organizations by making a donation. Most organizations have a ``donate'' button on their website, and your contribution, no matter how small, can make a big difference. 4. Spread awareness: Promote the cause on your social media profiles and share your experiences with others. You can use your social media platforms to raise awareness about the issues related to gender equality in sports and help spread the word about organizations that promote women's participation in sports. 5. Advocate: Contact your local representatives, politicians, and sports federations to encourage them to support women's sports, promote gender equality initiatives, and create opportunities for girls and women in sports. 6. Join campaigns: Participate in social media campaigns or public demonstrations organized by these organizations. You can join protests, walkouts, or demonstrations that aim to raise awareness about issues related to gender equality in sports. 7. Collaborate: Collaborate with other organizations and groups that focus on similar causes. This will help you create a larger impact, share resources and knowledge, and expand your network. Remember, the key to making a difference is to take action, no matter how small, consistent, and persistent it may be. \\\midrule
\textsc{Zephyr} + persona gpt4 &\underline{As someone deeply committed to the world of sports and passionate about promoting gender equality}, you may want to consider getting involved in charity work that advocates for this cause. Here are some suggestions to help you explore opportunities in this area: 1. Research organizations: There are numerous non-profit organizations that advocate for gender equality in sports. You might want to research some such organizations that resonate with your beliefs and values. The website Charity Navigator can be a great resource to help you identify reputable organizations in this field that have a proven track record of impactful work. Once you have a list of organizations, you can review their websites, social media handles, and annual reports to learn more about their activities, initiatives, and achievements. 2. Join an advocacy group: After identifying organizations that interest you, consider reaching out to them to learn if they have advocacy groups that you could join. These groups typically consist of volunteers and supporters who share similar values and are passionate about promoting gender equality in sports. Being a part of such a group can provide you with opportunities to connect with like-minded individuals, learn about the issues facing women in sports, and contribute to initiatives that promote gender equality. 3. Donate or fundraise: Many non-profit organizations have fundraising programs that you could participate in to raise funds for their work. You could donate money, host a fundraiser, or participate in a charity event. By donating or fundraising for an organization, you can contribute to their mission and support their work. Additionally, by participating in a charity event, you can bring awareness to the cause. 4. Provide expertise: As a sports professional with years of experience, you could offer your expertise to organizations working towards gender equality in sports. Depending on your area of expertise, you could offer to help with coaching, training, mentoring, or developing programs that promote women's sports. Offering your services can be a meaningful contribution to the organization's work. 5. Volunteer: Non-profit organizations typically rely on volunteers to execute their programs. As a sports professional with a busy schedule, you could still find time to volunteer for organizations that promote gender equality in sports. If your busy schedule does not allow for long-term commitments, you could consider shorter-term commitments, such as helping out during a one-time event. 6. Spread awareness: Lastly, one of the most important ways to promote gender \\\midrule
\textsc{MT} &1. Look for local organizations: Start by searching for organizations in your area that support gender equality in sports. You can check websites like Girls Who Sweat, Women's Sports Foundation, UN Women, or visit your local sports center, library, or community center. 2. Research opportunities: Once you have found possible organizations, go through their websites, social media, or reach out to them via email or phone, and find out about their charity events, volunteer programs, and donation campaigns. 3. Attend events: Attend the charity events organized by these organizations. This will give you an opportunity to learn more about the cause, meet like-minded individuals, and network with influential people in the sports community. 4. Volunteer: You can sign up to volunteer for the organizations. This may involve coaching girls' teams, assisting in training sessions, organizing tournaments, or working as a fundraiser at events. 5. Spread awareness: Spread awareness about gender equality in sports by sharing and promoting these organizations' missions via social media, word-of-mouth, or posting flyers in your community. 6. Donate: Finally, you can donate financially by making one-time gifts, becoming a monthly donor, or participating in charity auctions, runs, or cycling events where your donation goes directly to empower girls and women in sports. By following any one, or all, of these steps, you can make a real difference in promoting gender equality in sports and helping girls and women achieve their full potential through sports. \\\midrule
\textsc{MT} + persona gpt4 &\underline{As a sports professional, you have the power to make a significant impact off the field by advocating for gender equality in sports.} Here are some ways to get involved in charity work supporting this cause: 1. Research and Partner with Charities: Look up organizations like the WNBA Players Association, Women Win or the Women's Sports Foundation, and connect with them to learn about their initiatives, events, or campaigns. \underline{Collaborate with these charities by providing your time, social media support}, or resources to help raise awareness and funds. 2. Donate: Contribute financially through regular donations, host a fundraiser (online or in-person), or \underline{pledge a portion of your contract, award earnings or prize money} to sports-focused charities aligned with gender equality. 3. \underline{Mentoring: Volunteer to mentor female athletes and help them pursue careers in sports and leadership roles.} Mentorship programs facilitate empowerment by sharing expertise, nurturing talent and providing a support system to grow their potential. 4. \underline{Speaking Engagement: Host workshops for young females, participate in panel discussions, or public speaking events} that highlight and elevate opportunities and resources offered to female athletes, promote equal pay, sports training and better working conditions. 5. \underline{Collaborations: You can consider collaborating with the leagues or federations} that focus predominantly on women’s sports, participate in collaborative charity initiatives and make your platform available to raise funds or awareness by using hashtags, creating campaigns, or challenges that garner support from your fans. 6. Leadership role: You could aim to create a team, charity drive, or nonprofit organization centered on equality in sports. Leverage your social network, celebrity status or existing platform to amplify the message, rally troops by involving friends, teammates, and peers in your charity programs. By contributing to these initiatives, you are spreading the positive impact of the power of sports beyond the confines of playing fields, and setting an example for future generations of aspiring female athletes. \\\bottomrule
    \end{tabular}
    \caption{Qualitative comparison of generations between four different models for a prompt from Serena Williams. We \underline{underline} portions of the text that emphasize successful inference persona or shows effect of personalization.}
    \label{tab:generation-2-serena}
\end{table*}

\section{Alignment tax}
\label{apx:alignment-tax}

A common phenomenon during preference alignment is so-called alignment tax: model's degradations in out-of-domain tasks \cite{lin2023speciality}. Other than high-level roadmaps \cite{herd2023we, byrnes2023plan}, \citet{lee2024bapo} proposes to continue finetune on base model's output and \citet{lin2024mitigating} argues for selective weight averaging to mitigate alignment tax. 

A benefit of finetuning model with prefixes (active or passive) is that we can mitigate tax by removing prefix at test time. We investigate this by additionally evaluate \textsc{MT(Zephyr)} on out-of-domain tasks. For \textbf{safety}, we report reward accuracy \footnote{We instead aggregate by the summing over tokens logp to avoid length bias present in the dataset.}
on \texttt{refusals-dangerous/offensive} from RewardBench \citep{lambert2024rewardbench}. Using LLM harness \citep{gao2024harness}, we test \textbf{reasoning} through \texttt{arc\_easy/challenge}, and \texttt{piqa} \citep{clark2018arc,Bisk2020piqa} and \textbf{factuality} through \texttt{truthfulqa\_mc1/2} \citep{lin2022truthfulqa}.

\begin{figure}[!htp]
    \centering
    \includegraphics[width=1.0\linewidth,trim={0 0.5cm 0 0.5cm},clip]{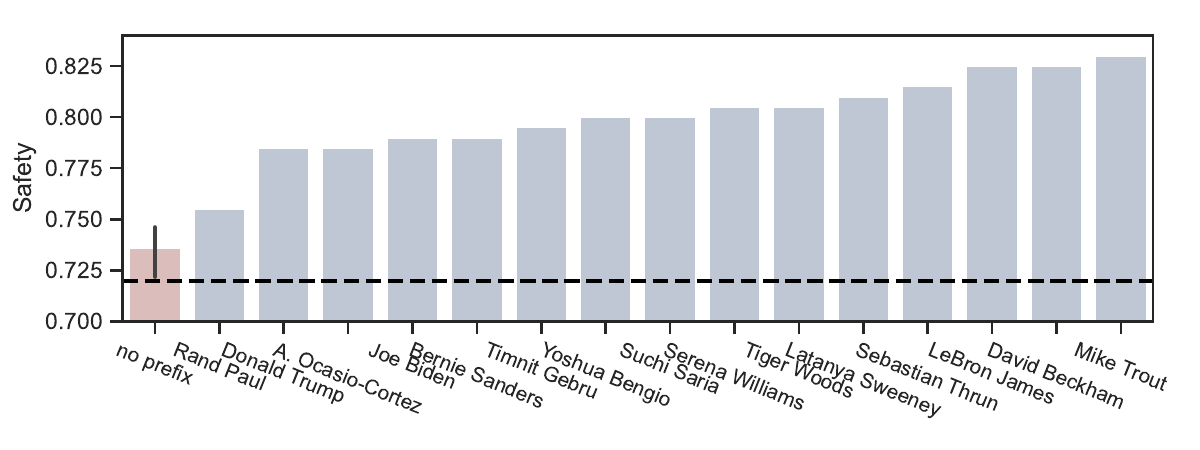}
    \includegraphics[width=1.0\linewidth,trim={0 0.5cm 0 0.4cm},clip]{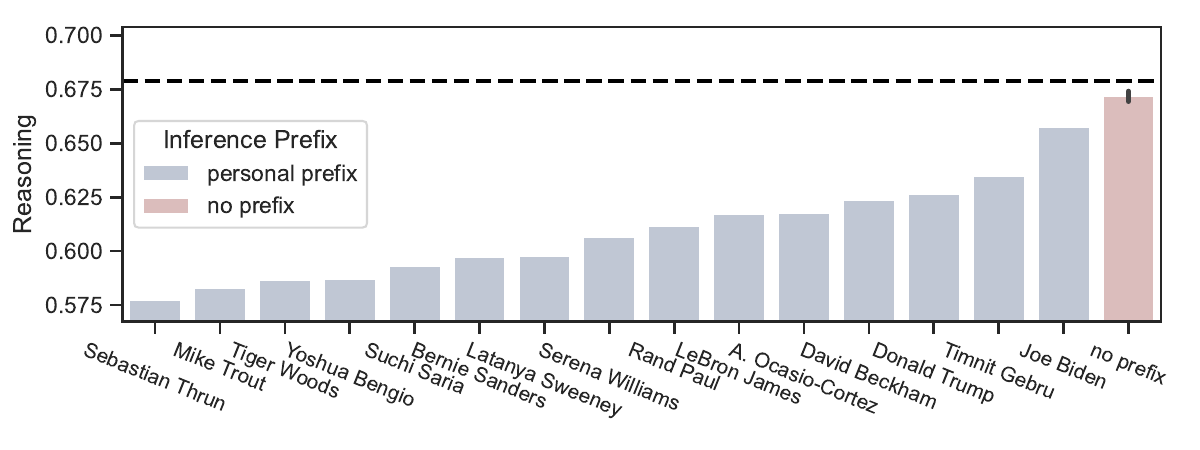}
    \includegraphics[width=1.0\linewidth,trim={0 0.5cm 0.5cm 0.3cm},clip]{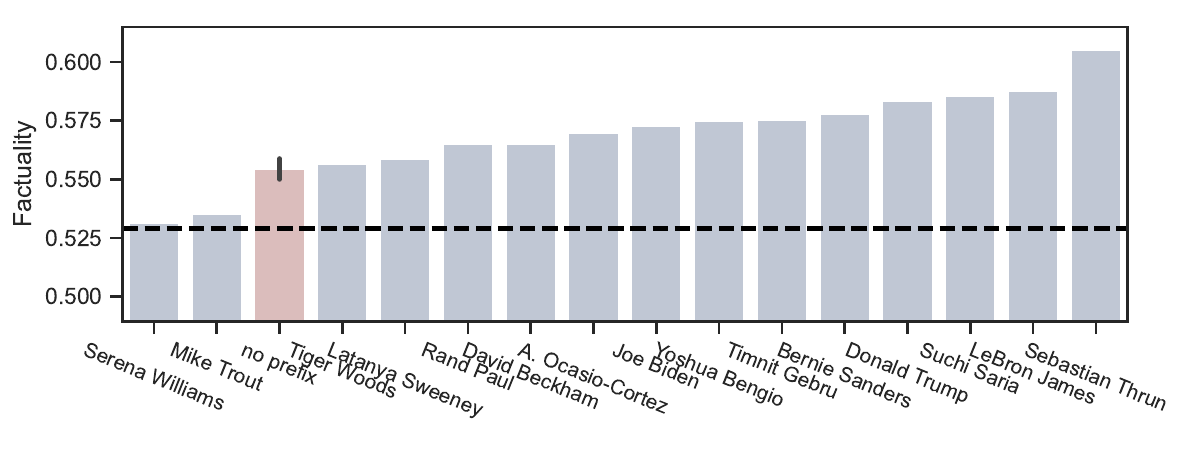}
    \caption{Sorted \textsc{MT(Zephyr)} with \textbf{persona gpt4} performance (not trained) on out-of-domain tasks. \textbf{No prefix} (aggregated across 5 CVs) returns model close to \textsc{Zephyr} no prefix (dashed line). Personas are sampled from axes \texttt{sports}, \texttt{AI professors}, and \texttt{politics}. Results with other prefixes are in Appendix~\ref{apx:alignment-tax}.}
    \label{fig:alignment-tax}
\end{figure}

Alignment with different personas results in varying performance in general tasks (\textbf{safety}, \textbf{reasoning}, \textbf{factuality})(Figure~\ref{fig:alignment-tax}) up to 10\% across individuals. The improvements in \textbf{safety} and \textbf{factuality} across the board are likely due to label signals from GPT4. \textbf{Reasoning} performance degrades across all personas, similar to observations by ~\citet{lee2024bapo}. This might be due to the questions focusing more on factual response than reasoning, even for \texttt{AI professors}. 
Across all three rows in Figure~\ref{fig:alignment-tax}, \textbf{no prefix} performance (red bar) is closer to baseline performance than most if not all personas. In deployment, if the user request does not require personalization (e.g. relating to objective truths), model providers can selectively run inference without a prefix. 

\begin{figure}[!htp]
    \centering
    \includegraphics[width=1.0\linewidth]{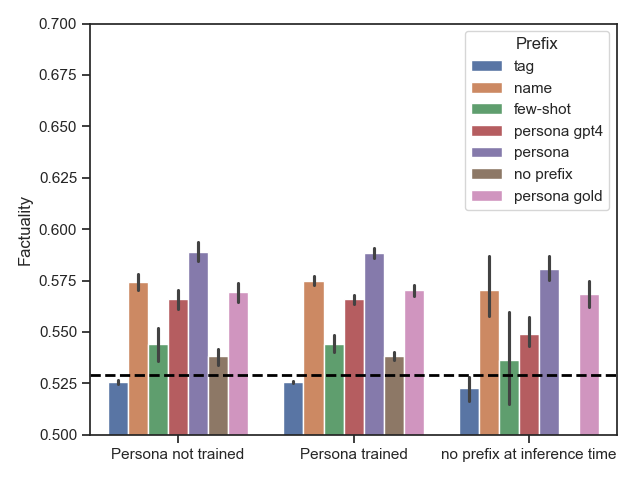}
    \includegraphics[width=1.0\linewidth]{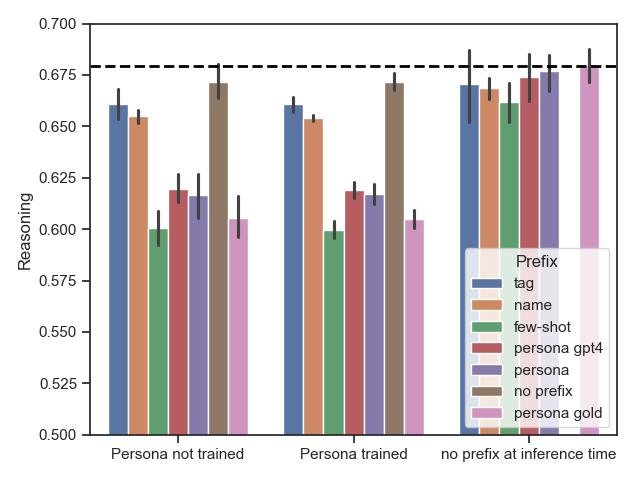}
    \caption{Reasoning and factuality performance on \textsc{MT(Zephyr)} models without using prefix at inference time. Black dashed line is \textsc{Zephyr} performance without any prefix.}
    \label{fig:alignment-tax-across-mt-models}
\end{figure}

In Figure~\ref{fig:alignment-tax}, we show that by not using any prefix at test time for \textsc{MT} models, we recover most of the baseline model performance. Here, in Figure~\ref{fig:alignment-tax-across-mt-models}, we observe that this is generally true regardless of the prefix in \textbf{reasoning}. However for \textbf{factuality} (OpinionQA), we do not see significant difference between using persona prefix vs not using prefix. This suggests these tasks may have inherently different mechanism that are differently affected during preference finetuning for personalization.
ß

\end{document}